\documentclass{article}

\PassOptionsToPackage{numbers, compress}{natbib}

\usepackage[preprint]{neurips_2021}

\usepackage[utf8]{inputenc} 
\usepackage[T1]{fontenc}    
\usepackage{hyperref}       
\usepackage{url}            
\usepackage{booktabs}       
\usepackage{amsfonts}       
\usepackage{nicefrac}       
\usepackage{microtype}      
\usepackage{xcolor}         
\usepackage{amsmath}
\usepackage{amssymb}
\usepackage{algorithm}
\usepackage{algorithmic}
\usepackage{subcaption}
\usepackage{color}
\usepackage{graphicx}
\newcommand{\argmax}{\mathop{\rm arg~max}\limits}

\title{Few-shot Learning for \\ Unsupervised Feature Selection}

\author{%
  Atsutoshi Kumagai\\
  NTT Computer and Data Science Laboratories\\
  \texttt{atsutoshi.kumagai.ht@hco.ntt.co.jp} \\
  \And
  Tomoharu Iwata \\
  NTT Communication Science Laboratories \\
  \texttt{tomoharu.iwata.gy@hco.ntt.co.jp} \\
  \AND
  Yasuhiro Fujiwara \\
  NTT Communication Science Laboratories \\
   \texttt{yasuhiro.fujiwara.kh@hco.ntt.co.jp} \\
}

\begin{document}

\maketitle

\begin{abstract}
We propose a few-shot learning method for unsupervised feature selection, 
which is a task to select a subset of relevant features in unlabeled data. 
Existing methods usually require many instances for feature selection.
However, sufficient instances are often unavailable in practice.
The proposed method can select a subset of relevant features in a target task given a few unlabeled target instances
by training with unlabeled instances in multiple source tasks.
Our model consists of a feature selector and decoder.
The feature selector outputs a subset of relevant features taking a few unlabeled instances as input
such that the decoder can reconstruct the original features of unseen instances from the selected ones.
The feature selector uses the Concrete random variables to select features via gradient descent.
To encode task-specific properties from a few unlabeled instances to the model,
the Concrete random variables and decoder are modeled using permutation-invariant neural networks that take a few unlabeled instances as input.
Our model is trained by minimizing the expected test reconstruction error given a few unlabeled instances that is calculated with datasets in source tasks.
We experimentally demonstrate that the proposed method outperforms existing feature selection methods.
\end{abstract}

\section{Introduction}
Feature selection is an important problem in machine learning that
aims to reduce dimensionality of data by identifying the subset of relevant features
\citep{chandrashekar2014survey}.
By extracting a small subset of features, 
we can more easily analyze/interpret the characteristics of datasets and 
accelerate the learning processes of subsequent tasks such as clustering and classification.
In addition, when the full set of features is expensive or difficult to collect,
feature selection can eliminate the cost of collecting irrelevant or redundant features.
Thanks to these beneficial properties,
feature selection methods have been widely used in various applications such as
biomarker discovery
\citep{xing2001feature},
document categorization
\citep{forman2008bns},
disease diagnosis
\citep{akay2009support}, 
and drug development
\citep{liu2004comparative}.

Many feature selection methods have been proposed.
Supervised methods use labeled instances for feature selection such as Least Absolute Shrinkage and Selection Operator (Lasso)~\citep{tibshirani1996regression} 
and kernel approaches
\citep{masaeli2010transformation,yamada2014high,yamada2016post}.
Although these methods are effective, labeled instances are expensive or impossible to collect
since labels need to be manually annotated by domain experts.
On the other hand, unsupervised methods require only unlabeled instances; 
therefore, they can be used in a wider range of situations than
supervised methods
\citep{he2006laplacian,zhao2007spectral,yang2011l2,balin2019concrete}.
Existing unsupervised feature selection methods usually require a relatively large amount of instances. 
However, sufficient instances might be unavailable in real-world applications.
For example, when we want to analyze data generated from industrial assets,
sufficient instances are difficult to collect from newly introduced assets.
When we want to analyze data obtained from persons for personalization
\citep{mobasher2007data}, 
sufficient instances are also difficult to collect from new persons.
\textcolor{black}{In healthcare, when medical test results are used as features, the full set of features is expensive to collect, and thus, collecting sufficient instances in a
hospital can be difficult
\cite{cesa2011efficient}.}
When there are no sufficient data, existing methods perform worse since overfitting easily occurs
\citep{liu2017deep,singh2020fsnet,yamada2014high}.
Thus, feature selection with small instances is a critical problem \citep{liu2017deep,jain1997feature}.

\begin{figure}[t]
\begin{minipage}{0.48\hsize}
      \centering
      \includegraphics[width=5.5cm]{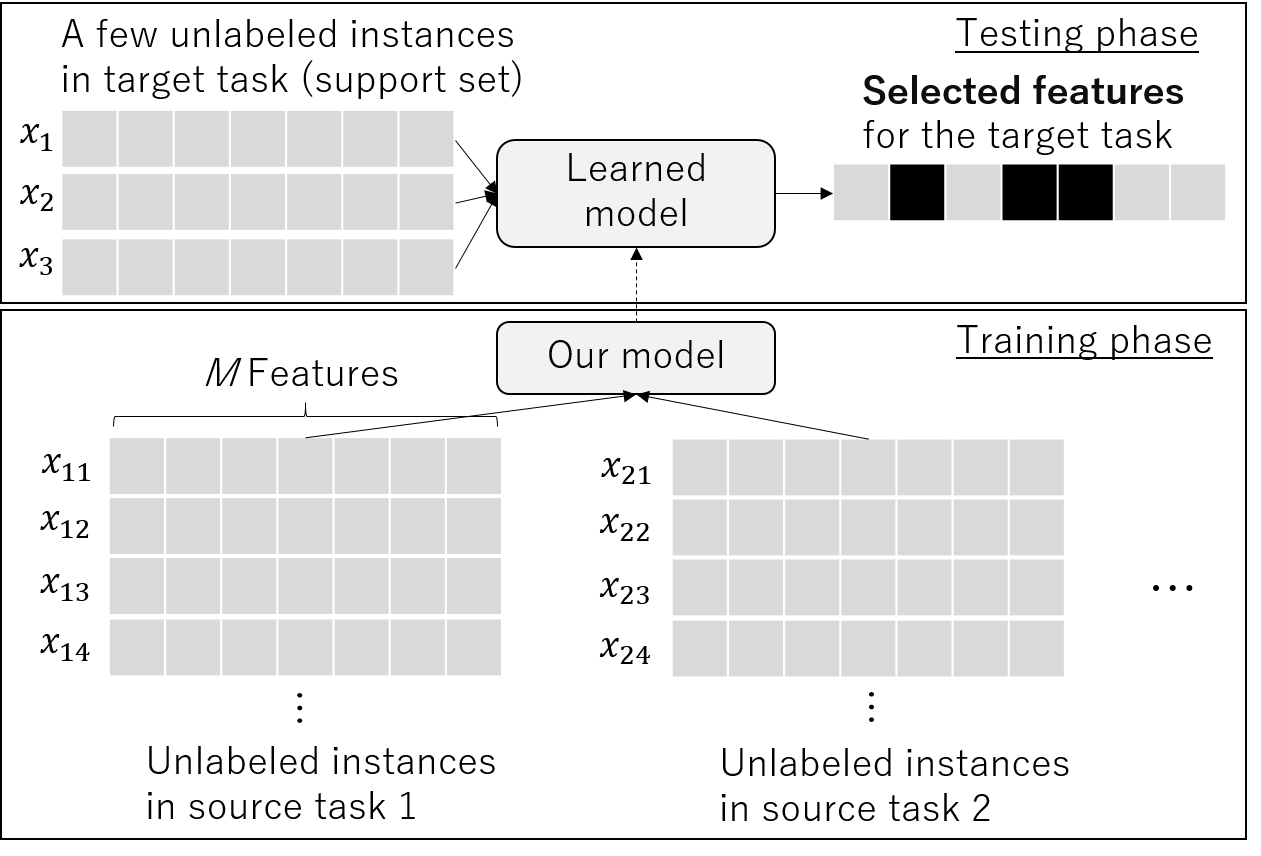}
      \caption{Our problem formulation. In the training phase, our model is learned with unlabeled instances in multiple source tasks.
In the testing phase, the learned model selects the subset of relevant features in a target task given the target support set.
We assume that all the source and target tasks have the same feature sizes $M$.}
      \label{pauc}
\end{minipage}
~~
\begin{minipage}{0.48\hsize}
      \centering
      \includegraphics[width=6.3cm]{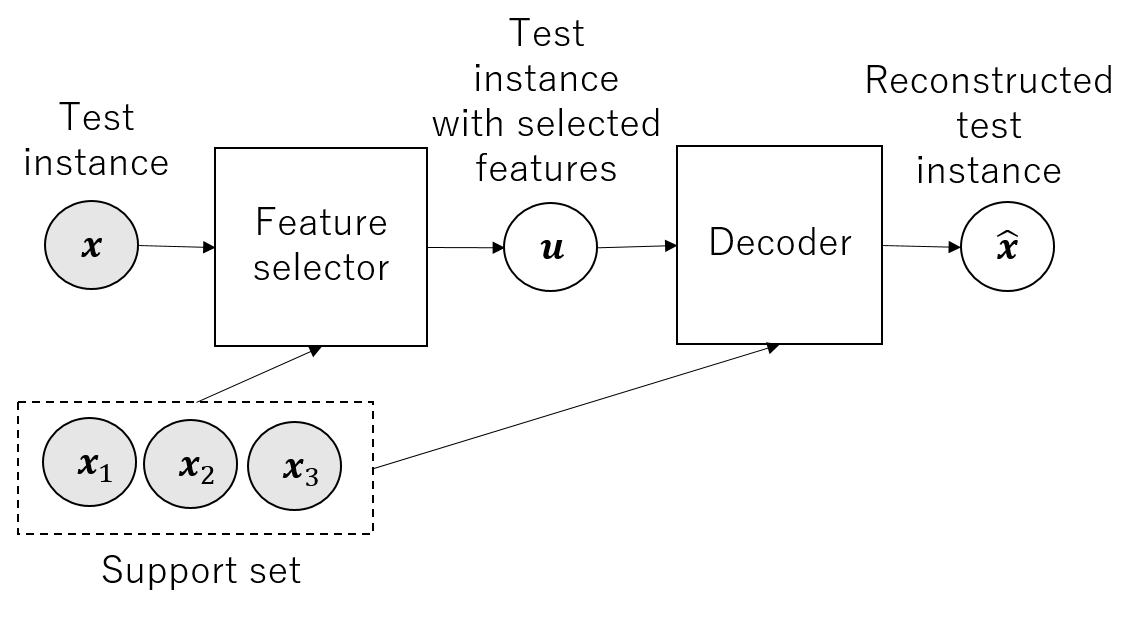}
      \caption{Overview of our model. The feature selector outputs the subset of relevant features 
      from a support set. The decoder outputs the reconstructed original features of test instances from the selected features considering the task property.}
      \label{pauc2}
\end{minipage}
\end{figure}

In this paper, we propose a few-shot learning method for unsupervised feature selection that enables us to select
the subset of relevant features in a task of interest, called a target task, given a few unlabeled instances.
To alleviate the lack of instances,
the proposed method trains the model using unlabeled instances in multiple related tasks, called source tasks.
In the above examples, other industrial assets/persons/hospitals can be regarded as related tasks.
When target and source tasks are related, we can transfer useful knowledge from the source tasks to the target task
\citep{pan2009survey}.
Figure~\ref{pauc} shows our problem formulation.

Our model consists of two components: a feature selector and decoder, 
which are based on neural networks that enable accurate feature selection with their high expressive power.
The feature selector outputs the subset of features taking a few unlabeled instances in a task, called a support set, as input.
The decoder reconstructs the original features of testing instances in the same task, called a query set, from the selected features.
The reconstruction from the selected features means that the unselected features can be approximated by the nonlinear transformation of selected features. 
Therefore, by selecting features that minimize the reconstruction errors, 
our model can automatically eliminate redundant features.
Figure \ref{pauc2} shows our model.

However, selecting a subset of features in neural networks is difficult because its operation is not differentiable.
To handle this problem, we use a continuous relaxation of the discrete random variables, the Concrete random variables
\citep{maddison2016concrete,jang2016categorical},
in the feature selector. 
\textcolor{black}{With our model, the Concrete random variables and decoder are modeled 
using permutation-invariant neural networks that take the support set as input.
By this modeling, our model can reflect task-specific properties from the support set in the parameters of Concrete random variables, 
which enables to select a suitable subset of features for each task given the support set.}

Our model is trained by minimizing the expected test reconstruction error of a query set given a support set
that is calculated using unlabeled instances in source tasks.
Since our model is explicitly trained so as to perform few-shot feature selection that works well in testing instances
in multiple tasks, 
we can expect it also to perform well on the target task.
Since all neural network parameters of our model are shared across all tasks, 
the learned model can be applied to unseen target tasks without re-training.
Although we focus on the unsupervised approach in this paper, our framework can be easily modified for the supervised one with suitable loss functions such as the cross-entropy loss. 

Our main contributions are summarized as follows:
1) To the best of knowledge, our work is the first attempt at few-shot learning for unsupervised feature selection.
2) We propose a reconstruction-based feature selection method that enables us to perform feature selection from a few unlabeled instances.
3) We empirically show that the proposed method performs well in few-shot feature selection problems.

\section{Related Work}
Feature selection methods have been widely studied.
Many existing methods use label information to select relevant features
\citep{ding2005minimum,peng2005feature,tibshirani1996regression,masaeli2010transformation,yamada2014high,yamada2016post,singh2020fsnet,yamada2020feature}.
These methods select the subset of features that can explain the labels well.
Although they are effective, labels are often expensive or impossible to prepare in real-world applications.
Unsupervised feature selection methods do not require labels~\citep{he2006laplacian,zhao2007spectral,cai2010unsupervised,yang2011l2}.
Some unsupervised methods use strong prior knowledge about a dataset (the number of clusters)~\citep{cai2010unsupervised,yang2011l2}
although the proposed method does not require such knowledge.
Recently, it has been reported that reconstruction-based feature selection methods perform better than traditional unsupervised approaches
because the data reconstruction error is a good criterion to define the relevance of features 
\citep{li2017reconstruction,wang2017feature,balin2019concrete,lemhadri2021lassonet}.
In addition, they can provide a way to naturally perform model selection on the basis of validation reconstruction errors, which is important for practical use,
although unsupervised methods typically have difficulty performing the model selection. 
Among them, concrete autoencoder (CAE)
\citep{balin2019concrete}
performs particularly well by using Concrete random variables in the autoencoder (AE) framework.
However, it requires many instances for training and thus cannot perform well with a few instances~\citep{singh2020fsnet}.
To make the best use of its capability with a few instances,
we extend the CAE's model by incorporating support set information and use an episodic training framework~\citep{vinyals2016matching}, 
which enables us to perform few-shot feature selection by transferring knowledge in source tasks.
Singh et al. \cite{singh2020fsnet} proposed a supervised extension of CAE, which uses diet networks~\citep{romero2016diet} 
to deal with high-dimensional data.
Unlike the proposed method, this method requires labels and cannot use any data in source tasks.

Some transfer learning methods for feature selection have been proposed
\citep{uguroglu2011feature,gautheron2018feature,bi2009transferred,helleputte2009feature,zhao2018msplit}.
These methods typically handle only two tasks (source and target tasks) and cannot handle more than two tasks.
In addition, these methods assume many unlabeled instances and/or small labeled instances in the target task.
In contrast, the proposed method handles multiple tasks and uses a few unlabeled target instances.
Although multi-task feature selection methods can handle more than two tasks~\citep{obozinski2006multi,zhou2010exclusive,argyriou2008convex},
they usually assume labeled instances in each task. 
Besides, transfer and multi-task learning methods require training for each (target) task.
The proposed method can perform feature selection for each target task without re-training, which enables fast adaptation to new tasks. 

Few-shot learning has recently attracted attention, which aims to adapt to new tasks rapidly and effectively with a few instances
\citep{snell2017prototypical,finn2017model,finn2018probabilistic,yoon2018bayesian,santoro2016meta,garnelo2018conditional,vinyals2016matching,rajeswaran2019meta,yin2019meta,iwata2020meta}.
Most existing studies have proposed methods for few-shot classification tasks, which aim to obtain classifiers to recognize unseen classes during training, or reinforcement learning tasks
\citep{snell2017prototypical,finn2017model,garnelo2018conditional,vinyals2016matching}.
Some methods aim to learn a subspace or subset in the latent feature space for few-shot classification tasks
\citep{dvornik2020selecting2,li2019finding,lichtenstein2020tafssl}.
These methods require labels and cannot perform feature selection from original features.
In contrast, the proposed method does not require labels and can perform feature selection from original features.

\section{Preliminary}
We briefly review Concrete or Gumbel-softmax distribution
\citep{maddison2016concrete,jang2016categorical},
which is used in the proposed method.
Let ${\bf z}$ be a categorical variable with class probabilities $( \frac{\alpha_1}{\sum_m \alpha_m}, \dots, \frac{\alpha_M}{\sum_m \alpha_m} )$ where
$\alpha _m \in \mathbb{R} _{>0} $. 
We can assume that states with $0$ probability are excluded
\citep{maddison2016concrete}.
We assume that ${\bf z}$ is represented as an $M$-dimensional one-hot vector, where the $m$-th element of ${\bf z}$, $z_m$, is one if the $m$-th class is selected and $z_m=0$ otherwise.
The Gumbel-Max trick provides an efficient way to draw an instance ${\bf z}$ from 
the categorical distribution as ${\bf z} = {\rm one hot} \left(\argmax _m  [ {\rm log } \ \alpha_m + g_m ] \right)$,
where $g_m = - {\rm log} ( - {\rm log} (u_m) )$, $ u_m \sim {\rm Uniform} (0,1)$, and ${\rm one hot} (m)$ returns a one-hot vector
where the $m$-th element is one.
The instance ${\bf z}$ is not differentiable with respect to $\alpha_m$ 
since the ${\rm one hot} (\argmax [\cdot])$ operator is not differentiable.
To deal with this problem, Maddison et al.
\cite{maddison2016concrete} 
proposed a differentiable relaxation of the discrete variable with softmax transformations:
$z_m  = \frac{{\rm exp} (({\rm log } \ \alpha_m + g_m)/\tau)}{\sum _j {\rm exp} (({\rm log } \ \alpha_j + g_j)/\tau) }, \ \ m=1, \dots, M,$
where $\tau > 0$ is a temperature parameter.
This approximate variable is called a Concrete random variable.
Obviously, Concrete random variables are differentiable with respect to $\alpha_m$.
When $\tau$ is large, the Concrete random variable becomes a uniform vector.
When $\tau$ approaches zero, the Concrete random variable becomes a one-hot vector and 
follows the categorical distribution with class probabilities $( \frac{\alpha_1}{\sum_m \alpha_m}, \dots, \frac{\alpha_M}{\sum_m \alpha_m} )$
\citep{maddison2016concrete,jang2016categorical}.
The probability distribution for Concrete random variables is called the Concrete distribution or Gumbel-softmax distribution. 

\section{Proposed Method}
\subsection{Problem Formulation}
Let $X_d$ be a set of unlabeled instances in the $d$-th task, and ${\bf x} _n \in X_d $ be
the $M$-dimensional feature vector of the $n$-th instance in the $d$-th task.
We assume that each task has the same feature vector size $M$, but
each distribution can differ, which is the standard assumption used in transfer learning studies~\citep{pan2009survey}.
Suppose that unlabeled instances in $D$ source tasks $ X = \{ X _d \} _{d=1}^{D} $ are given at the training phase.
\textcolor{black}{At the testing phase, we are given a few unlabeled instances in a target task $ X _{d'} $.
Here, the target task is not contained in the source tasks, i.e., $d' \notin \{ 1,\dots,D \}$.
Our goal is to select at most $K$ features ($K \le M$) that are appropriate for the target task from $ X _{d'} $.}

\subsection{Model}
We explain our model that selects the subset of relevant features in a task given a few unlabeled instances in the task.
Our model consists of two main components: a feature selector and decoder as described in Figure \ref{pauc2}.
The feature selector outputs the subset of relevant features from a few unlabeled instances in a task, called a support set.
The decoder is used to reconstruct the original inputs of testing instances, called a query set, from the selected features in the same task.
The reconstruction means that all features can be approximated by the nonlinear transformation of selected features. 
Thus, by minimizing the reconstruction error, our model can select an informative feature subset.

We first explain the feature selector.
Given a support set ${\cal S}$ from a task, 
the $k$-th feature $u^{(k)} \in \mathbb{R}$ is obtained by the feature selector as follows:
\begin{align}
\label{k_node}
u^{(k)} 
=  {\bf x} \cdot {\bf z} ^{(k)} ({\cal S}), \ \ k=1, \dots, K,
\end{align}
where $\cdot$ is the inner product, ${\bf x}$ is a query instance in the same task, and ${\bf z} ^{(k)} ({\cal S}) = (z^{(k)}_1 ({\cal S}),\dots,z^{(k)}_M ({\cal S}))$ is a 
Concrete random variable with parameters ${\boldsymbol \alpha} ^{(k)} ({\cal S}) =(\alpha^{(k)}_1 ({\cal S}) ,\dots,\alpha^{(k)}_M ({\cal S})) \in \mathbb{R} ^{M}_{>0}$:
\begin{align}
\label{gambel_softmax_our}
z^{(k)}_m ({\cal S}) = & \frac{{\rm exp} (({\rm log } \ \alpha ^{(k)}_m ({\cal S}) + g_m^{(k)})/\tau)}{\sum _j {\rm exp} 
(( {\rm log } \ \alpha ^{(k)}_j ({\cal S}) + g_j^{(k)})/\tau) }, \ \ \   k=1,\dots,K, \ m=1,\dots,M.
\end{align}

Eq. \eqref{k_node} means that $u^{(k)} $ is a linear combination of input features.
As the temperature $\tau$ approaches zero, ${\bf z} ^{(k)} ({\cal S})$ becomes a one-hot vector and 
therefore $u^{(k)}$ exactly outputs one of the input features.
As a result, the feature selector can select at most $K$ features from the input features when ${\tau}$ is small. 
The stochasticity of Concrete random variables enables informative feature combinations to be efficiently explored,
which is analyzed in the supplemental material.

With our model, the parameters of Concrete random variable ${\boldsymbol \alpha} ^{(k)} ({\cal S})$ depend on the support set ${\cal S}$.
Therefore, our model encodes the characteristics of the support set (task) to the parameters
so that it can perform task-specific feature selection that is suitable for each task.
We model this function by the permutation-invariant feed-forward neural networks 
\citep{NIPS2017_6931}:
\begin{align}
\label{perm_net}
{\rm log} \ {\boldsymbol \alpha} ^{(k)} ({\cal S}) \!=\! g_{\phi _2} \left( \left[ \sum_{{\bf x} \in {\cal S} } f_{\phi _1} ({\bf x} ), {\boldsymbol \pi} ^{(k)} \right] \right), \ k=1, \!\dots\!, K,
\end{align}
where $f_{\phi_1}$ and $g_{\phi_2}$ are any feed-forward neural networks with parameters $\phi_1$ and $\phi_2$, respectively,
$[\cdot , \cdot]$ is the concatenation, 
${\boldsymbol \pi} ^{(k)} \in \mathbb{R} ^{T} $ is learnable parameters for the $k$-th selected feature, and
${\rm log} \ {\boldsymbol \alpha} ^{(k)} ({\cal S})$ means that ${\rm log}$ is applied for each component of ${\boldsymbol \alpha} ^{(k)} ({\cal S})$.
Due to summation, this neural network is permutation-invariant to the order of instances in the support set
and thus is well-defined as a function for set inputs
\citep{NIPS2017_6931}.
The parameters ${\boldsymbol \pi} ^{(k)}$ are introduced to select different features in the feature selector.
That is, 
each ${\boldsymbol \pi} ^{(k)}$ would take different values to minimize the reconstruction error, which leads to vary the class probability of each Concrete 
distribution, and thus, output different features.
\textcolor{black}{Note that in the absence of ${\boldsymbol \pi} ^{(k)}$, this neural network outputs the same values for all $k$ from ${\cal S}$.
It means that the same feature is selected for all $k$, which is not desirable.}

The decoder outputs the reconstructed original features ${\bf {\hat x}} \in \mathbb{R}^{M} $ from the selected features
${\bf u}  = (u^{(1)}, \dots, u^{(K)}) \in \mathbb{R}^{K} $.
Specifically, the decoder is defined as follows:
\begin{align}
\label{recon_model}
{\bf {\hat x}}   = h_{\theta}  \left( \left[ {\bf u} ({\bf x}; {\cal S}), {\bf r} ({\cal S}) \right] \right), \ 
{\bf r} ({\cal S}) = g_{\psi _2} \left( \sum_{{\bf x} \in {\cal S} } f_{\psi _1} ({\bf x}) \right),
\end{align}
where $h_{\theta}$, $f_{\psi_1}$, and $g_{\psi_2}$ are any feed-forward neural networks with parameters $\theta$, $\psi_1$, and $\psi_2$, respectively.
\textcolor{black}{In Eq. \eqref{recon_model}, we explicitly described that ${\bf u}$ depends on both ${\bf x}$ and ${\cal S}$ for clarity.}
${\bf r} ({\cal S})$ is also permutation-invariant to the order of instances in ${\cal S}$.
Since different feature subsets can be selected in each task,
different decoders are required for each task to reconstruct any instances in each task.
By using ${\bf r} ({\cal S})$, we can model task-specific decoders.
\begin{algorithm}[t]
\caption{Training procedure of our model.}
\label{arg}
\begin{algorithmic}[1]
\REQUIRE Datasets in source tasks $X$, support set size $N_{{\rm S}}$,  
query set size $N_{{\rm Q}}$, maximum number of iterations $I$, initial temp. $T_0$, final temp. $T_1$
\ENSURE Parameters of our model $\Theta$
\STATE{Initialize iteration number: $i=0$}
\REPEAT
\STATE{Set temperature: $\tau =  T_0 (T_1/T _0)^{i/I}$ }
\STATE{Sample task $d$ from $\{ 1,\dots,D\}$ }
\STATE{Sample support set ${\cal S} $ with size $N_{{\rm S}}$ from $X _d$}
\STATE{Sample query set ${\cal Q} $ with size $N_{{\rm Q}}$ from $X _d \setminus {\cal S}$}
\STATE{Update parameters with the gradients of the reconstruction error on the query set  ${\cal Q} $}
\STATE{$i=i+1$}
\UNTIL{End condition is satisfied;}
\end{algorithmic}
\end{algorithm}

\subsection{Training}
We estimate parameters of our model $\Theta = (\theta, \phi_1, \phi_2, \psi_1, \psi_2, {\boldsymbol \pi^{(1)}},\dots, {\boldsymbol \pi^{(K)}})$ by minimizing the expected reconstruction error on a query set ${\cal Q}$ given a support set ${\cal S}$
using an episodic training framework~\citep{snell2017prototypical,vinyals2016matching},
where support and query sets are randomly generated from
source datasets $X$:
\begin{align}
\label{objective}
\underset{\Theta}{{\rm min}} \ \mathbb{E} _{d \sim \{1,\dots,D \}} \biggl[ \mathbb{E} _{({\cal S},{\cal Q}) \sim X _d} \biggl[ \biggr. \biggr. \left. \left. \frac{1}{N_{\rm Q}}  \sum_{ {\bf x} \in {\cal Q}} \Vert {\bf x} - h_{\theta} ( \left[ {\bf u} ({\bf x} ; {\cal S}), {\bf r} ({\cal S}) \right] ) \Vert^2 \right] \right],
\end{align}
where $\mathbb{E}$ is the expectation, $\Vert \cdot \Vert^2$ is the squared Euclidean norm, ${\bf x} $ is an instance in the query set, and
$N_{{\rm Q} }$ is the number of instances in the query set.
The pseudocode for our training procedure is illustrated in Algorithm~\ref{arg}.
This training enables our model to learn how to select relevant features from a few instances since
it simulates feature selection from a few instances in multiple source tasks so as to work well for unseen instances.
Intuitively, since all model parameters $\Theta$ are shared across all tasks, 
they can be learned with all instances in all source tasks that enables knowledge to be shared between the tasks.
We can learn how to select features in each task from instances in the task.

We use the annealing schedule for the temperature parameter $\tau$. 
Although small $\tau$ selects individual features in each Concrete random variable, 
it cannot explore various combinations of features and converges to a poor local minimum
\citep{balin2019concrete}.
Thus, we decay the value of $\tau$ at each iteration from large value $T_0$ to small value $T_1$ as in CAE~\citep{balin2019concrete},
by which our model can effectively explore combinations of features in the initial phase and converge to informative individual features.
The effect of this annealing is analyzed in detail in the supplemental material.
Although we used the reconstruction error as the loss function for simplicity, we can use any other loss function.
For example, our framework can be straightforwardly modified for supervised settings by using the cross-entropy loss. 

\subsection{Feature Selection}
Given the target support set $X_{d'}$ at the testing phase,
we can estimate the parameters of Concrete random variables as
${\rm log} \ {\boldsymbol \alpha} ^{(k)} (X_{d'}), \ k=1,\dots,K$.
To select the subset of features from the parameters,
 we use the discrete ${\argmax}$ operator instead of Concrete random variables because 
differentiability is unnecessary for testing. 
Specifically, the $k$-th feature index is selected by ${\argmax _m} \ {\rm log} \ \alpha _m ^{(k)} (X_{d'})$.

\section{Experiments}
In this section, we demonstrate the effectiveness of the proposed method for few-shot feature selection.
We evaluated the selected features in terms of both reconstruction and clustering qualities.
For clustering, each feature selection method is first performed to select features, 
and the K-means clustering is performed on the basis of the selected features, which is the standard procedure in 
unsupervised feature selection studies~\citep{li2017reconstruction,yang2011l2,zhao2007spectral,belkin2002laplacian}.
Since K-means is a simple method, we can evaluate the selected features with less risk of overfitting.
We used scikit-learn implementation for the K-means, which automatically alleviates effects of bad initial centroid seeds by repeating multiple runs.
All experiments were conducted on a Linux server with an Intel Xeon CPU and a NVIDIA GeForce GTX 1080 GPU.

\subsection{Data}
We used four real-world datasets: MNIST-r, Isolet, Amazon, and VLCS, which have been widely used in previous studies
\citep{ghifary2015domain,gong2013connecting,motiian2017unified,balin2019concrete}.
MNIST-r is derived from MNIST by rotating the images~\citep{ghifary2015domain}.
This dataset has six tasks (six rotation angles), and its feature dimension is $256$.
Isolet consists of letters spoken by 150 speakers, and speakers are grouped into five groups (tasks) by speaking similarity. 
Each instance is represented  as a 617-dimensional vector.
Amazon consists of product reviews in four tasks (four product categories), where the feature dimension is 400.
VLCS consists of real images on four tasks (four data sources), where each image is represented by $4096$ dimensional features
\citep{ghifary2015domain,motiian2017unified}.
The numbers of classes of MNIST-r, Isolet, Amazon, and VLCS are 10, 26, 2, and 5, respectively.
The details of the datasets are provided in the supplemental material.

\subsection{Comparison Methods}
We compared the proposed method with seven commonly used unsupervised feature selection methods:
CAE-S, CAE-T, CAE-ST, LS-T, LS-ST, SPEC-T, and SPEC-ST.
CAE~\citep{balin2019concrete}
is a recently proposed reconstruction-based feature selection method
that has been reported to perform better than existing methods in various datasets
\citep{balin2019concrete}.
CAE-T and CAE-S were trained with target instances and source instances, respectively.
CAE-ST was trained with target instances by fine-tuning after pre-training with source instances.
Unlike the proposed method, CAE uses task-invariant Concrete random variables and decoder.
LS is the Laplacian Score~\citep{he2006laplacian},
which is a widely used unsupervised feature selection method
that selects features on the basis of Laplacian Eigenmaps~\citep{belkin2002laplacian}.
LS-T and LS-ST were trained with target instances and both target and source instances, respectively.
SPEC~\citep{zhao2007spectral} 
is the general framework of spectral unsupervised feature selection.
SPEC-T and SPEC-ST were trained with target and both target and source instances, respectively.
The proposed method (Ours) and CAE are neural network-based feature selection methods.
We implemented both methods using PyTorch
\citep{paszke2017automatic}.
\textcolor{black}{LS and SPEC are not neural-network based methods, which are suitable for feature selection with small instances due to their simple models.}
We used scikit-feature
\citep{li2018feature}
implementation for both methods. 

\subsection{Settings}
We used mean squared reconstruction error (MSRE) for evaluating the reconstruction ability.
For clustering, we used two metrics for evaluating each method: Adjusted Rand Index (ARI) and Normalized Mutual Information (NMI).
Both metrics are widely used to evaluate clustering results.
In both metrics, values close to one indicate high quality clustering.
For the K-means, we set the number of clusters to that of classes for each dataset, which
is the standard procedure in previous studies
\citep{he2006laplacian,han2018autoencoder,li2017reconstruction}.
The proposed method and CAE used the same architecture of neural networks for the base AEs for fair comparison.
For each dataset, we randomly selected one task for the target task and the rest for the source (training/validation) tasks.
We randomly created 20 different target, training, and validation sets and evaluated the average test MSREs, ARIs, and NMIs over 20 sets for each pair of the number of
target support instances and number of selected features.
We used the number of target support instances $N_{\rm S}$ within $\{2,4,6\}$ and selected features $K$ within $\{10,20,30,40,50\}$.
The setup details such as network architectures and hyperparameters are described in the supplemental material.

\subsection{Results}
\label{all_results}
\paragraph{Reconstruction Ability}
We evaluated the test MSRE for the reconstruction-based methods: the proposed method and CAE.
Table \ref{table:recon} shows the averages and standard deviations of test MSREs over different numbers of target support instances $N_{\rm S}$ and selected features $K$, where $N_{\rm S}$ and $K$ were evaluated within  $\{2,4,6\}$ and $\{10,20,30,40,50\}$, respectively.
In Table \ref{table:recon}, we did not include LS and SPEC because both methods do not have components for reconstructions.
In Tables \ref{table:recon}--\ref{table:nmi}, boldface denotes the best and comparable methods according to the paired t-test and the significance level of 5 \%.
The proposed method showed the best test MSREs for all datasets.
This is because our model is trained so that it selects features from a few instances that can reconstruct original features of unseen instances. 
CAE-T performed poorly for all datasets because there were too few target support instances to train the neural network.
Interestingly, CAE-ST performed worse than CAE-S even though it used target instances for fine-tuning.
This is because there were too few target support instances to fine-tune, and thus fine-tuning destroyed the pre-trained model parameters.
Figure \ref{fig_recon} shows test MSREs with $N_{\rm S}=2$ when varying the number of selected features $K$ for each dataset.
We found that the proposed method consistently achieved the best test MSREs in each case.
Note that it seems that the proposed method and CAE-S yielded similar results due to the poor results of CAE-T in MNIST-r and Isolet of Figure \ref{fig_recon}, 
the proposed method was statistically better than CAE-S as described in Table \ref{table:recon}.
The proposed method also performed well when $N_{\rm S}=4$ and $6$, which is described in the supplemental material..
\begin{table*}[t]
\caption{Averages and standard deviations of test MSREs over different numbers of target support instances and selected features.
}
\label{table:recon}
\centering
\scalebox{0.85}{
\begin{tabular}{lrrrr}
\hline
Data & \multicolumn{1}{c}{Ours} &  \multicolumn{1}{c}{CAE-S} & \multicolumn{1}{c}{CAE-T} & \multicolumn{1}{c}{CAE-ST} \\
\hline
MNIST-r & \textbf{4.258(2.252)} & 4.459(2.242) & 15.477(4.453) & 8.029(3.826) \\
Isolet & \textbf{39.441(6.179)} & 42.941(5.272) & 161.411(17.395) & 95.196(15.243)  \\
Amazon & \textbf{0.056(0.004)} & 0.065(0.004) & 0.079(0.011) & 0.143(0.137)  \\
VLCS & \textbf{0.851(0.057)} & 0.914(0.023) & 0.963(0.057) & 1.137(0.175)  \\
\hline
\end{tabular}
}
\end{table*}
\begin{figure*}[t]
  \begin{minipage}{0.245\hsize}
      \centering
      \includegraphics[width=3.85cm]{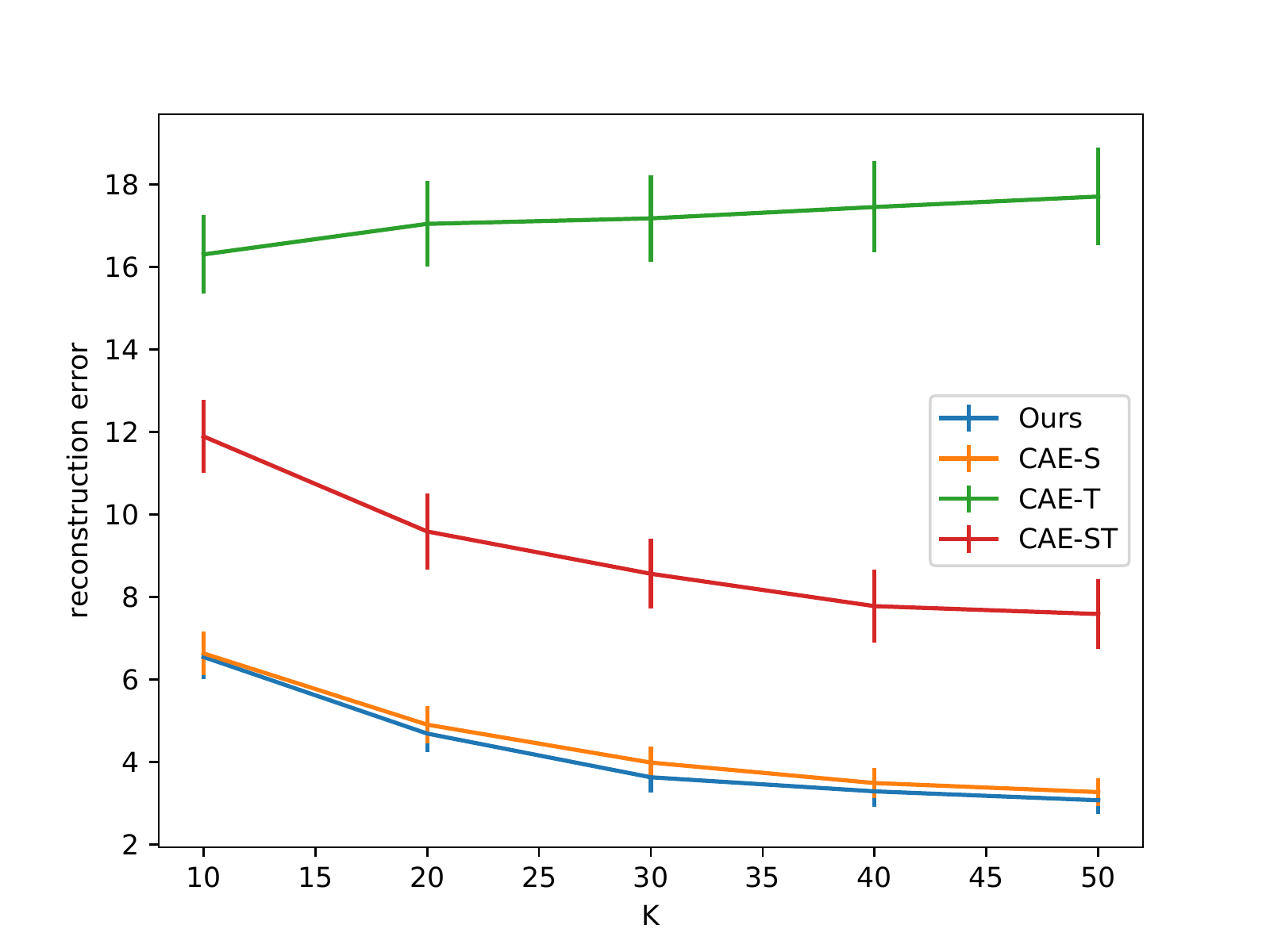}
      \subcaption{MNIST-r}
  \end{minipage}
  \begin{minipage}{0.245\hsize}
      \centering
      \includegraphics[width=3.85cm]{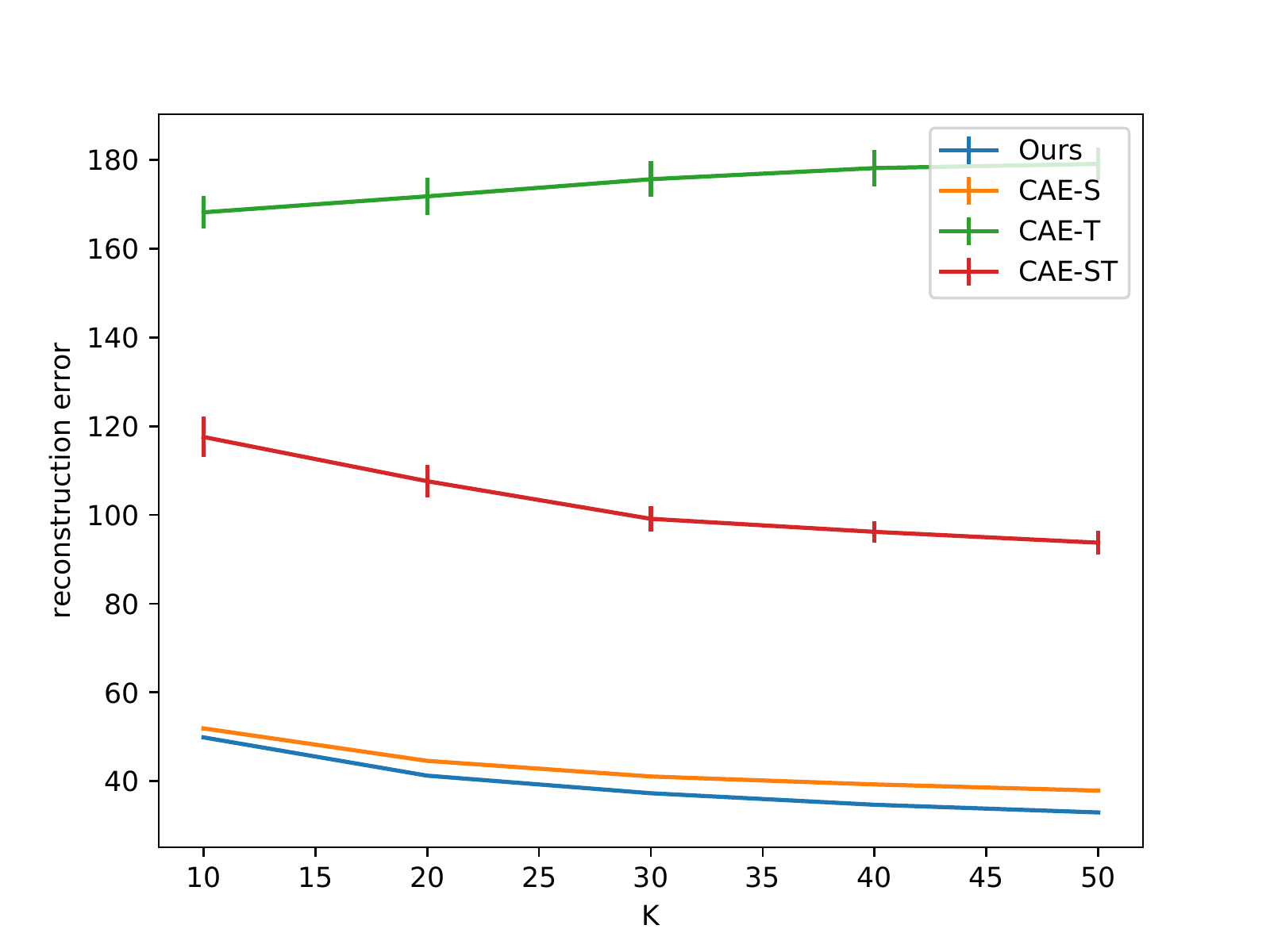}
      \subcaption{Isolet}
  \end{minipage}
  \begin{minipage}{0.245\hsize}
      \centering
      \includegraphics[width=3.85cm]{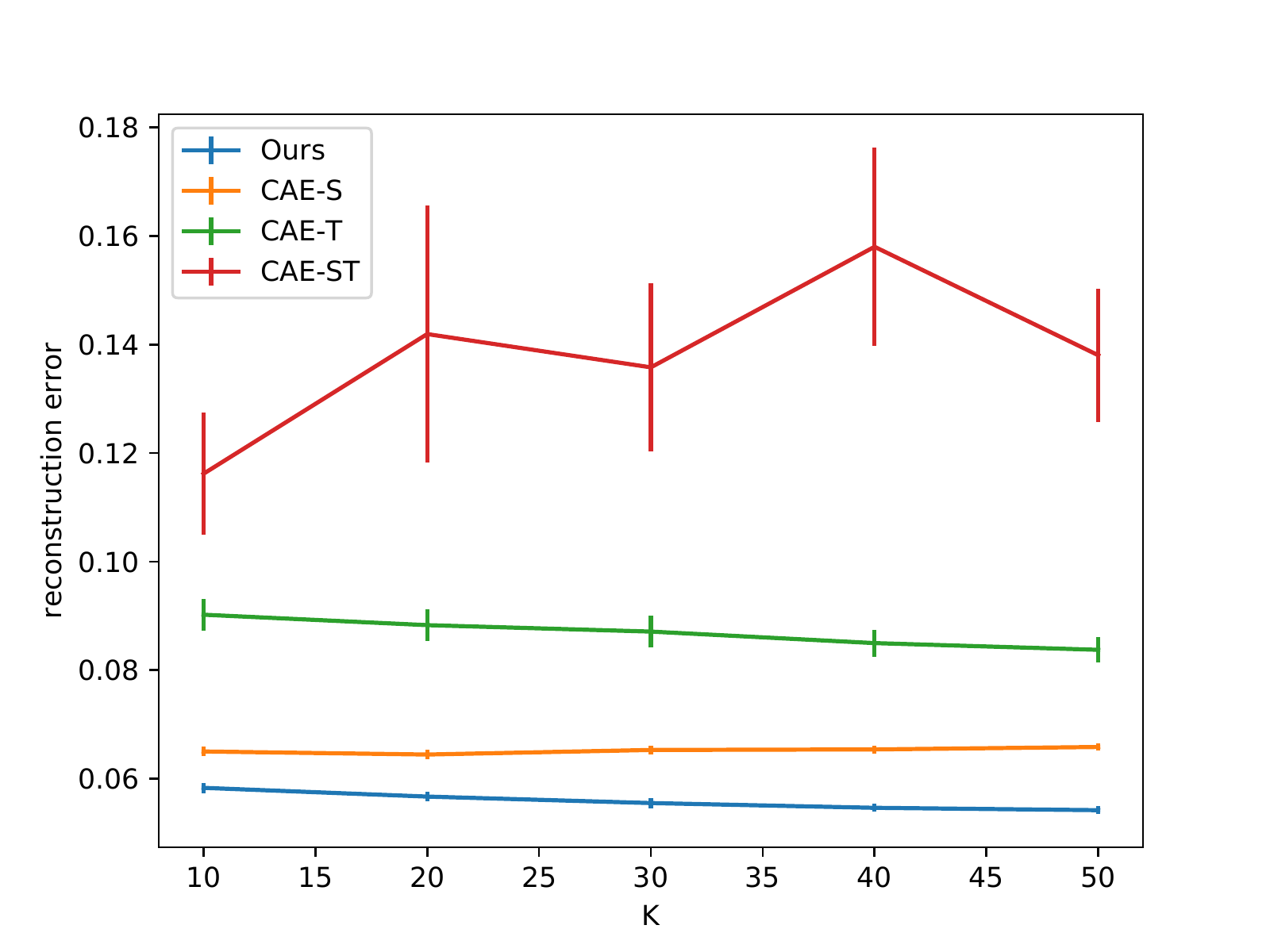}
      \subcaption{Amazon}
  \end{minipage}
   \begin{minipage}{0.245\hsize}
      \centering
      \includegraphics[width=3.85cm]{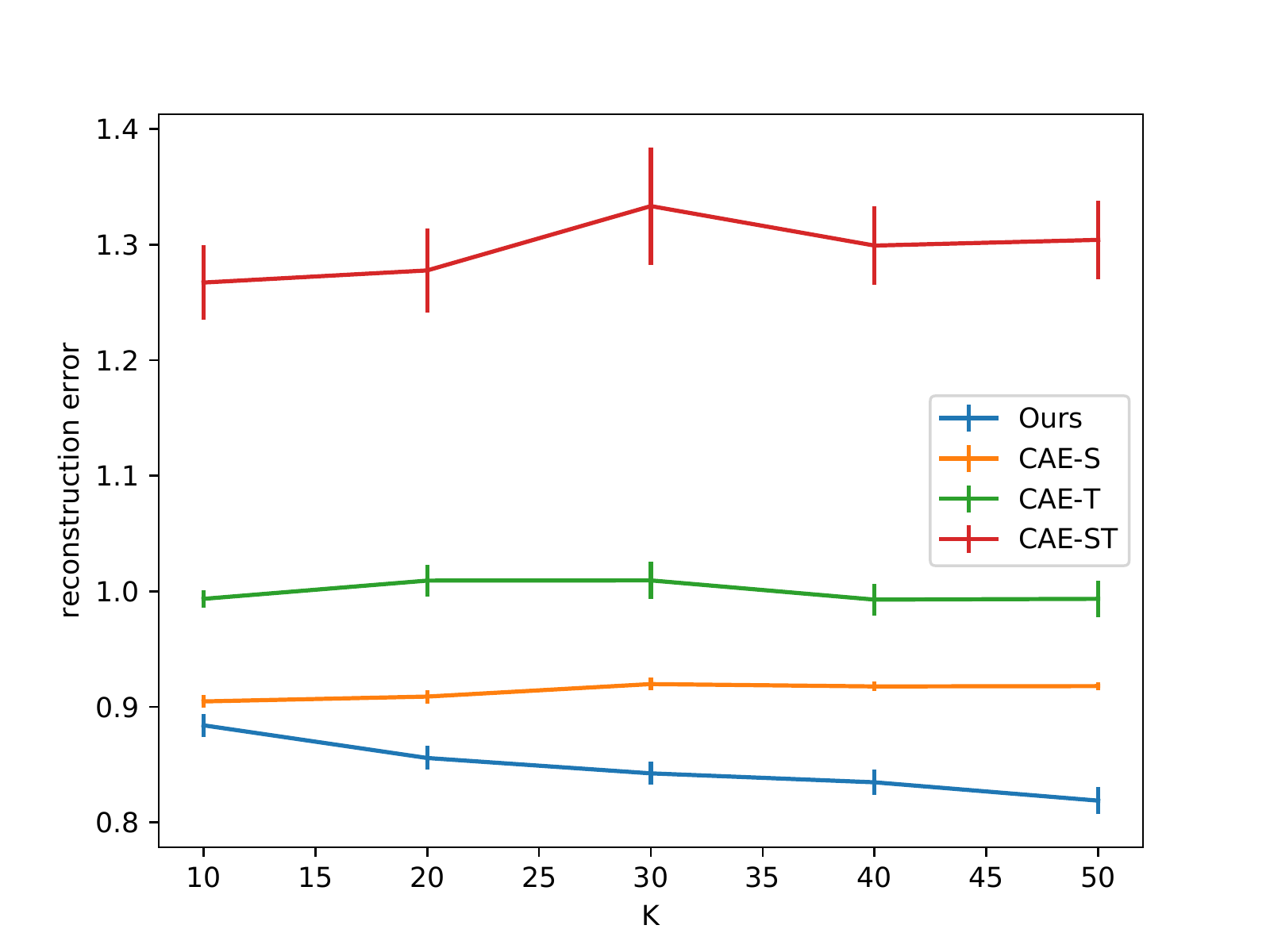}
      \subcaption{VLCS}
  \end{minipage}
  \caption{Average and standard errors of test MSREs with 2 target support instances when changing $K$.}
  \label{fig_recon}
\end{figure*}
\begin{figure*}[t]
  \begin{minipage}{0.19\hsize}
      \centering
      \includegraphics[width=2.5cm]{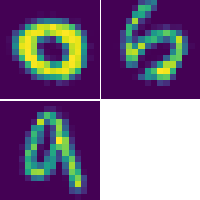}
      \subcaption{Original}
  \end{minipage}
  \begin{minipage}{0.19\hsize}
      \centering
      \includegraphics[width=2.5cm]{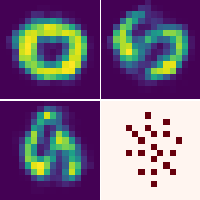}
      \subcaption{Ours}
  \end{minipage}
  \begin{minipage}{0.19\hsize}
      \centering
      \includegraphics[width=2.5cm]{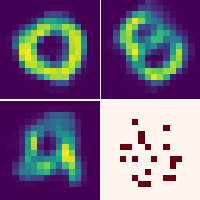}
      \subcaption{CAE-S}
  \end{minipage}
  \begin{minipage}{0.19\hsize}
      \centering
      \includegraphics[width=2.5cm]{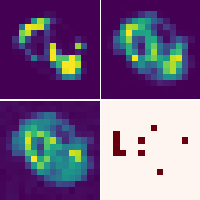}
      \subcaption{CAE-T}
  \end{minipage}
  \begin{minipage}{0.19\hsize}
      \centering
      \includegraphics[width=2.5cm]{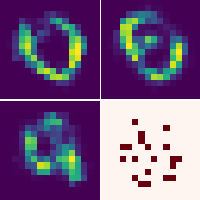}
      \subcaption{CAE-ST}
  \end{minipage}
  \caption{Test reconstructed images when 4 target support instances and $20$ selected features (pixels) were used on MNIST-r.
  The lower right picture for each method represents the features selected (red pixels) by the method.}
  \label{fig_images}
\end{figure*}

\begin{table*}[t]
\caption{Averages and standard deviations of test ARIs [\%] over different numbers of target support instances and selected features.
}
\label{table:ari}
\centering
\scalebox{0.85}{
\begin{tabular}{lrrrrrrrr}
\hline
Data & \multicolumn{1}{c}{Ours} &  \multicolumn{1}{c}{CAE-S} & \multicolumn{1}{c}{CAE-T} & \multicolumn{1}{c}{CAE-ST} & 
\multicolumn{1}{c}{LS-T} & \multicolumn{1}{c}{LS-ST} & \multicolumn{1}{c}{SPEC-T} & \multicolumn{1}{c}{SPEC-ST} \\
\hline
MNIST-r & \textbf{30.2(6.3)} & 24.7(5.1) & 16.7(5.6) & 24.7(5.1) & 21.8(5.3) & 20.0(5.9)  & 3.4(3.7) & 3.7(5.1)  \\
Isolet & \textbf{33.9(5.9)} & \textbf{33.8(5.6)} & 20.1(7.4) & \textbf{33.7(5.4)} & 21.0(7.4) & 29.3(7.1) & 14.7(5.7) & 19.3(6.9)  \\
Amazon & \textbf{0.5(0.9)} & 0.2(0.4) & 0.2(0.5) & 0.2(0.4) & 0.2(0.4) & 0.2(0.1) & 0.1(0.1) & 0.2(0.2)  \\
VLCS & \textbf{11.2(9.4)} & 3.5(4.8) & 7.8(10.0) & 3.7(4.5) & \textbf{11.0(14.8)} & 6.7(7.1) & \textbf{10.3(14.7)} & 2.0(4.6)  \\
\hline
\end{tabular}
}
\end{table*}
\begin{table*}[!]
\caption{Averages and standard deviations of test NMIs [\%] over different numbers of target support instances and selected features.
}
\label{table:nmi}
\centering
\scalebox{0.83}{
\begin{tabular}{lrrrrrrrr}
\hline
Data & \multicolumn{1}{c}{Ours} &  \multicolumn{1}{c}{CAE-S} & \multicolumn{1}{c}{CAE-T} & \multicolumn{1}{c}{CAE-ST} & 
\multicolumn{1}{c}{LS-T} & \multicolumn{1}{c}{LS-ST} & \multicolumn{1}{c}{SPEC-T} & \multicolumn{1}{c}{SPEC-ST} \\
\hline
MNIST-r & \textbf{42.6(6.1)} & 38.0(5.3) & 28.6(6.2) & 38.0(5.3) & 34.5(5.9) & 33.0(6.2)  & 15.5(4.6) & 12.3(8.6)  \\
Isolet & \textbf{61.4(4.9)} & \textbf{61.5(4.9)} & 46.2(8.8) & \textbf{61.4(4.8)} & 47.9(8.7) & 58.7(6.7) & 40.3(6.6) & 43.3(7.7)  \\
Amazon & \textbf{1.4(2.6)} & 0.6(0.9) & 0.5(1.0) & 0.6(0.9) & 0.3(0.9) & 0.1(0.1) & 0.6(0.8) & \textbf{1.3(1.1)}  \\
VLCS & \textbf{23.3(17.2)} & 15.4(14.3) & 21.0(19.0) & 15.5(14.2) & 21.9(19.9) & \textbf{23.2(19.7)} & 15.0(14.2) & 10.0(9.2)  \\
\hline
\end{tabular}
}
\end{table*}

\begin{figure*}[t]
  \begin{minipage}{0.245\hsize}
      \centering
      \includegraphics[width=3.85cm]{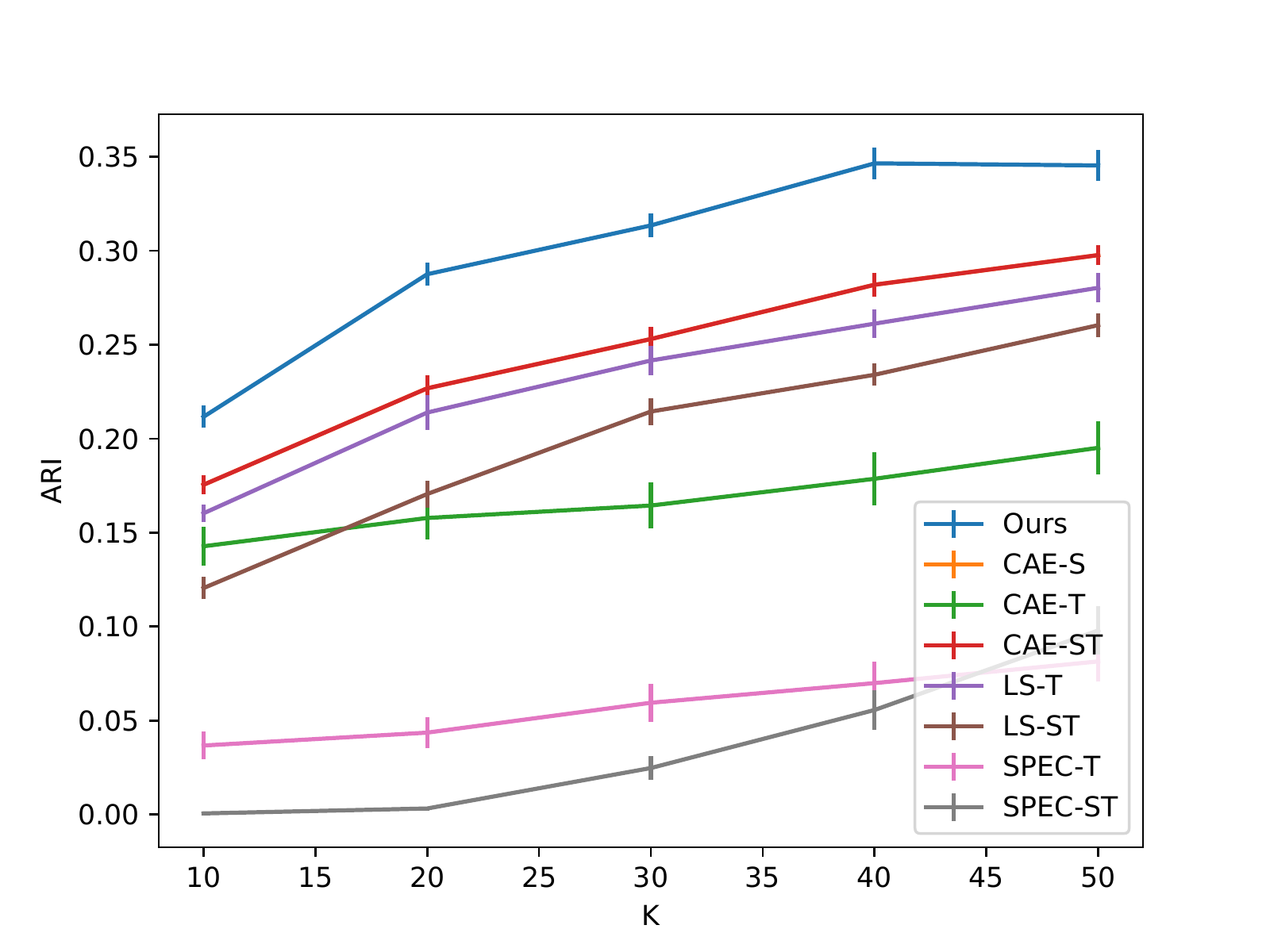}
      \subcaption{MNIST-r}
  \end{minipage}
  \begin{minipage}{0.245\hsize}
      \centering
      \includegraphics[width=3.85cm]{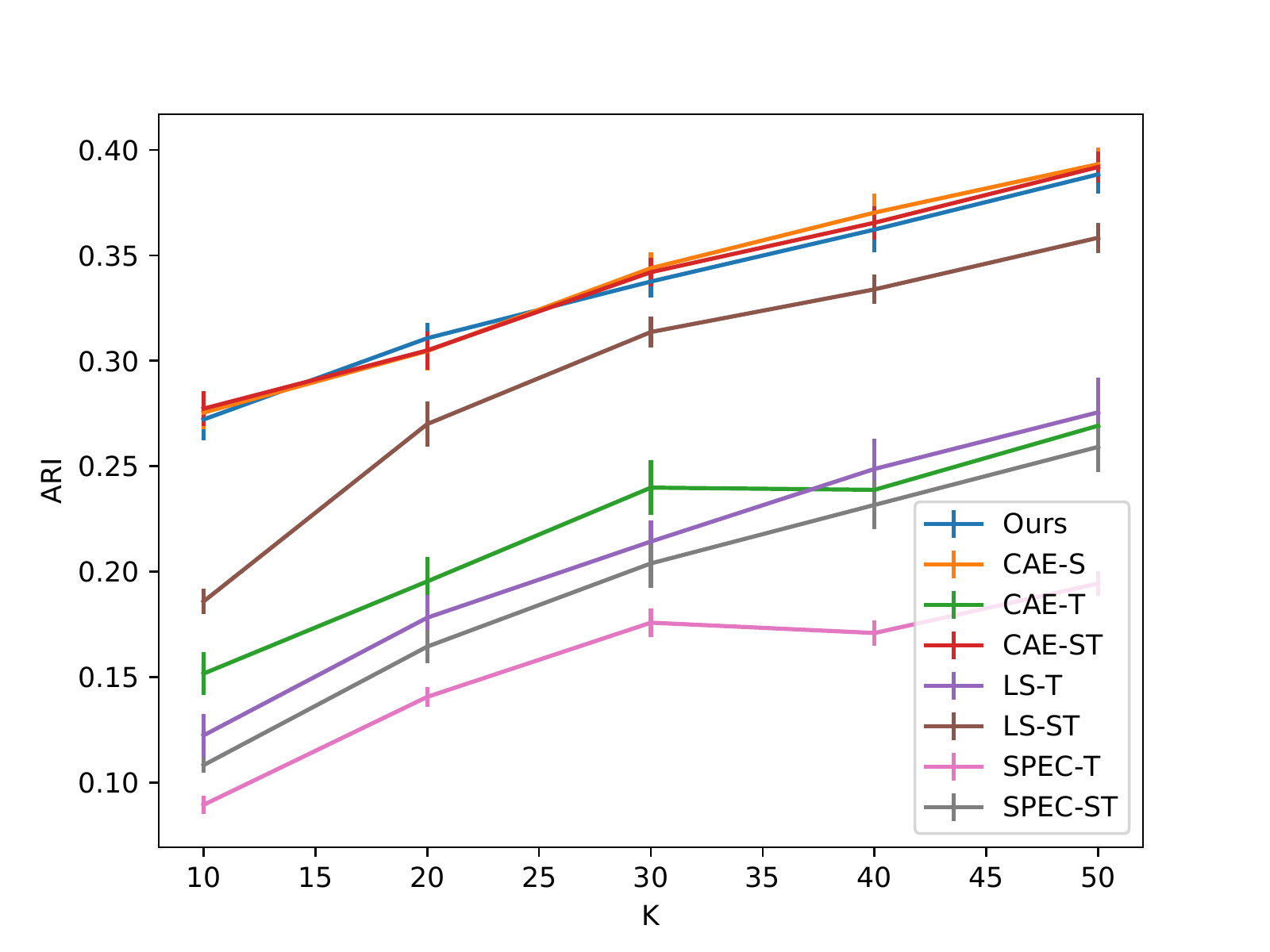}
      \subcaption{Isolet}
  \end{minipage}
  \begin{minipage}{0.245\hsize}
      \centering
      \includegraphics[width=3.85cm]{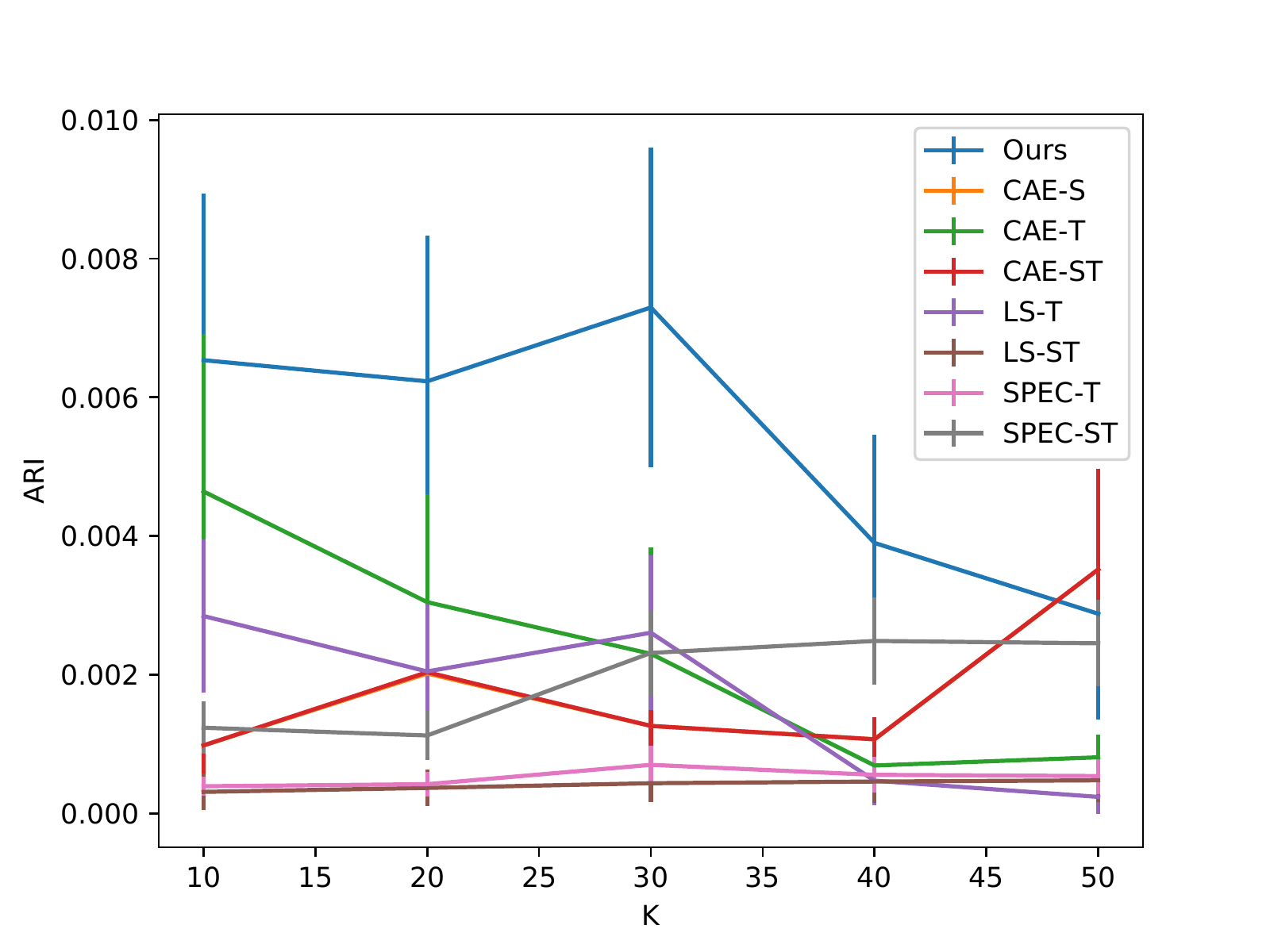}
      \subcaption{Amazon}
  \end{minipage}
   \begin{minipage}{0.245\hsize}
      \centering
      \includegraphics[width=3.85cm]{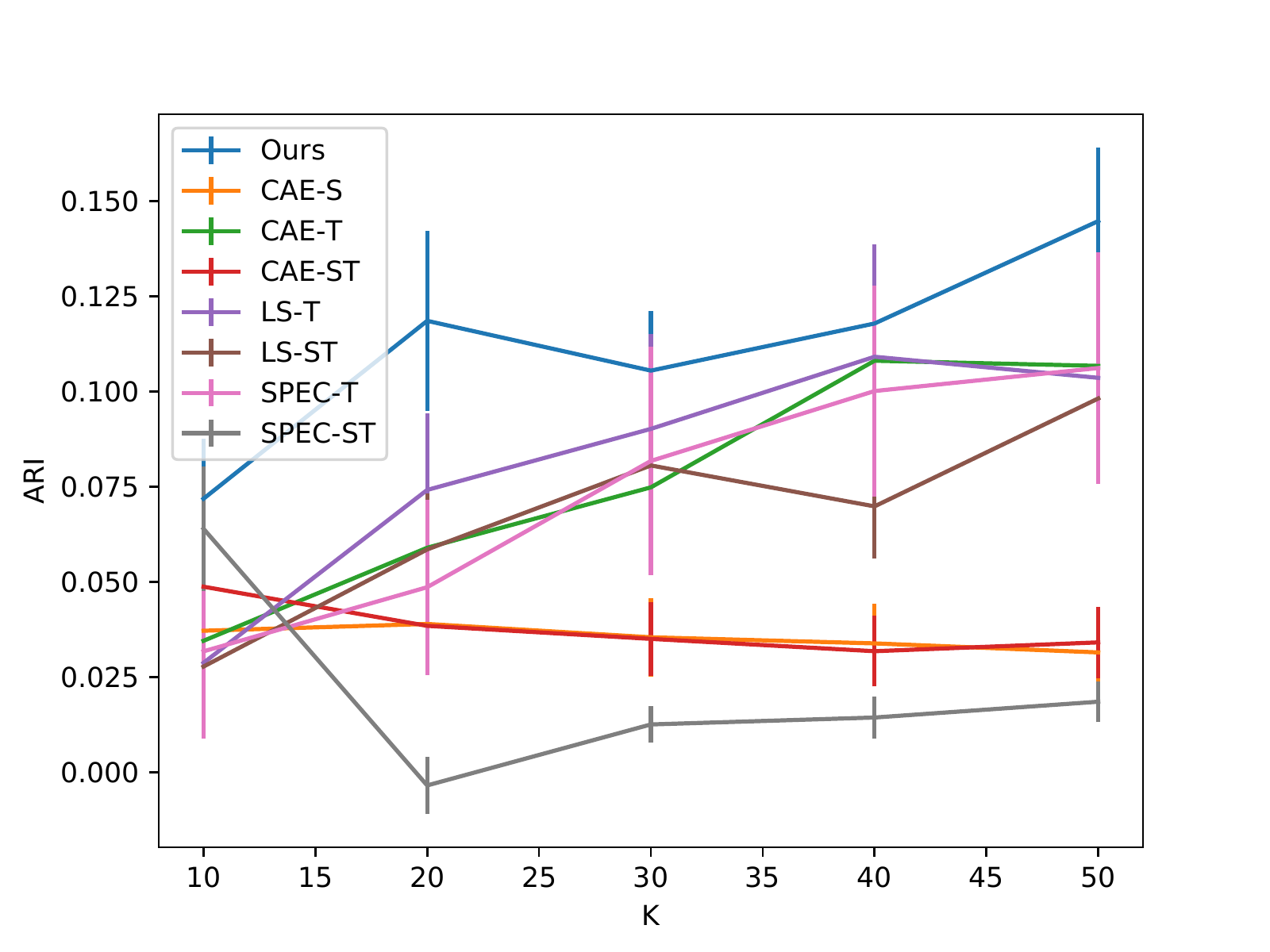}
      \subcaption{VLCS}
  \end{minipage}
  \caption{Average and standard errors of test ARIs with 2 target support instances when changing $K$.}
  \label{fig_ari}
\end{figure*}

\begin{figure*}[t]
  \begin{minipage}{0.245\hsize}
      \centering
      \includegraphics[width=3.85cm]{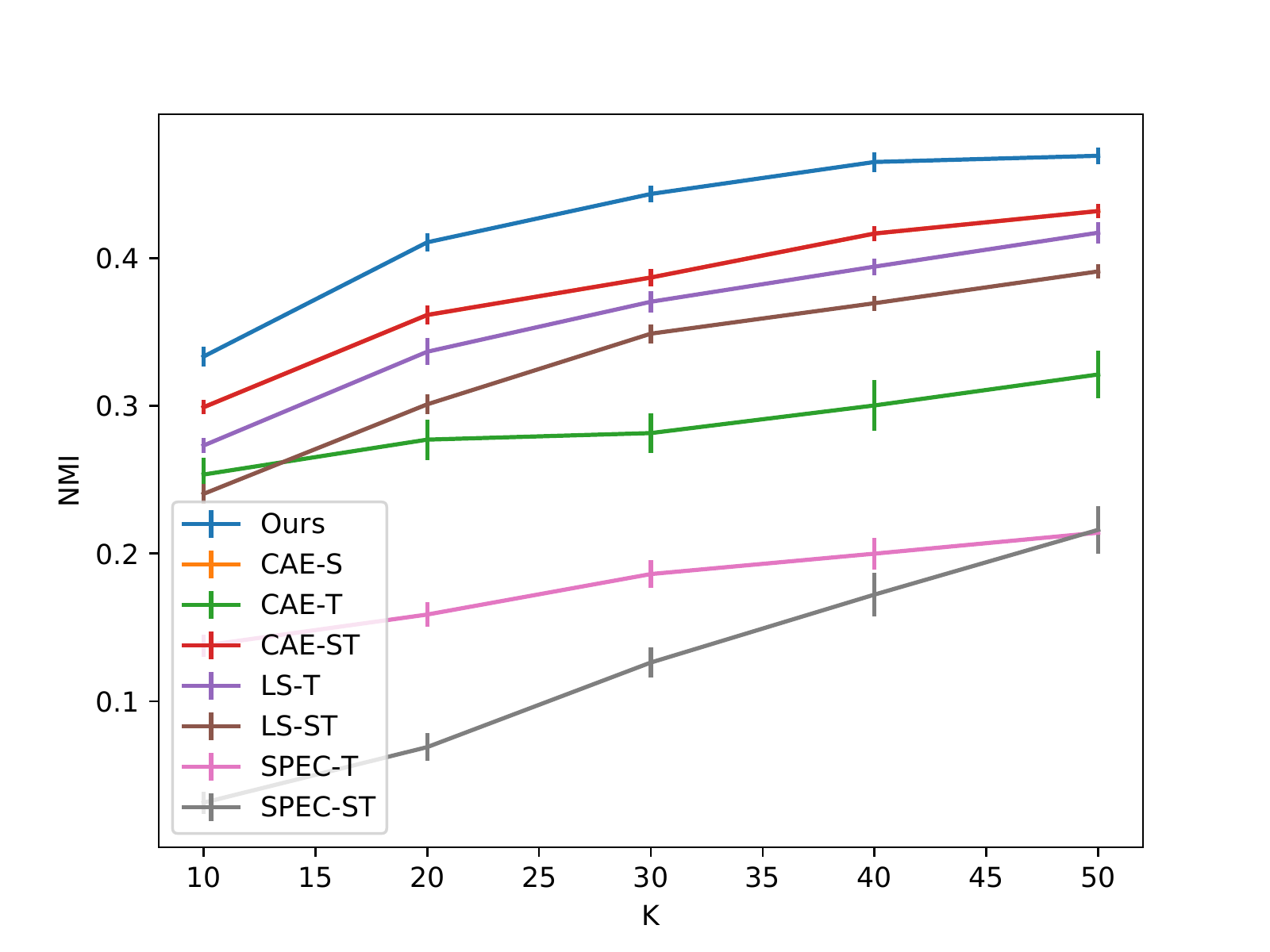}
      \subcaption{MNIST-r}
  \end{minipage}
  \begin{minipage}{0.245\hsize}
      \centering
      \includegraphics[width=3.85cm]{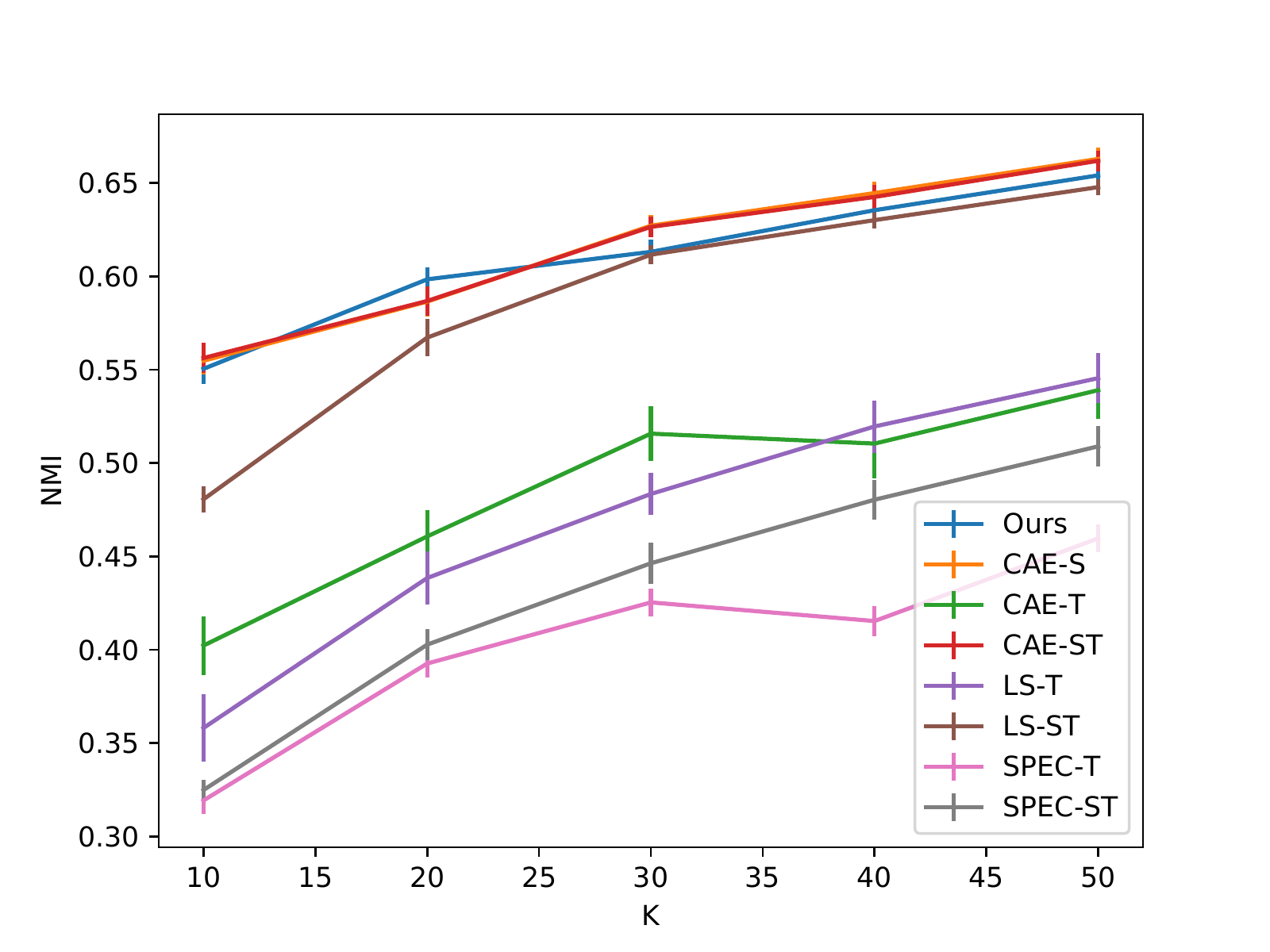}
      \subcaption{Isolet}
  \end{minipage}
  \begin{minipage}{0.245\hsize}
      \centering
      \includegraphics[width=3.85cm]{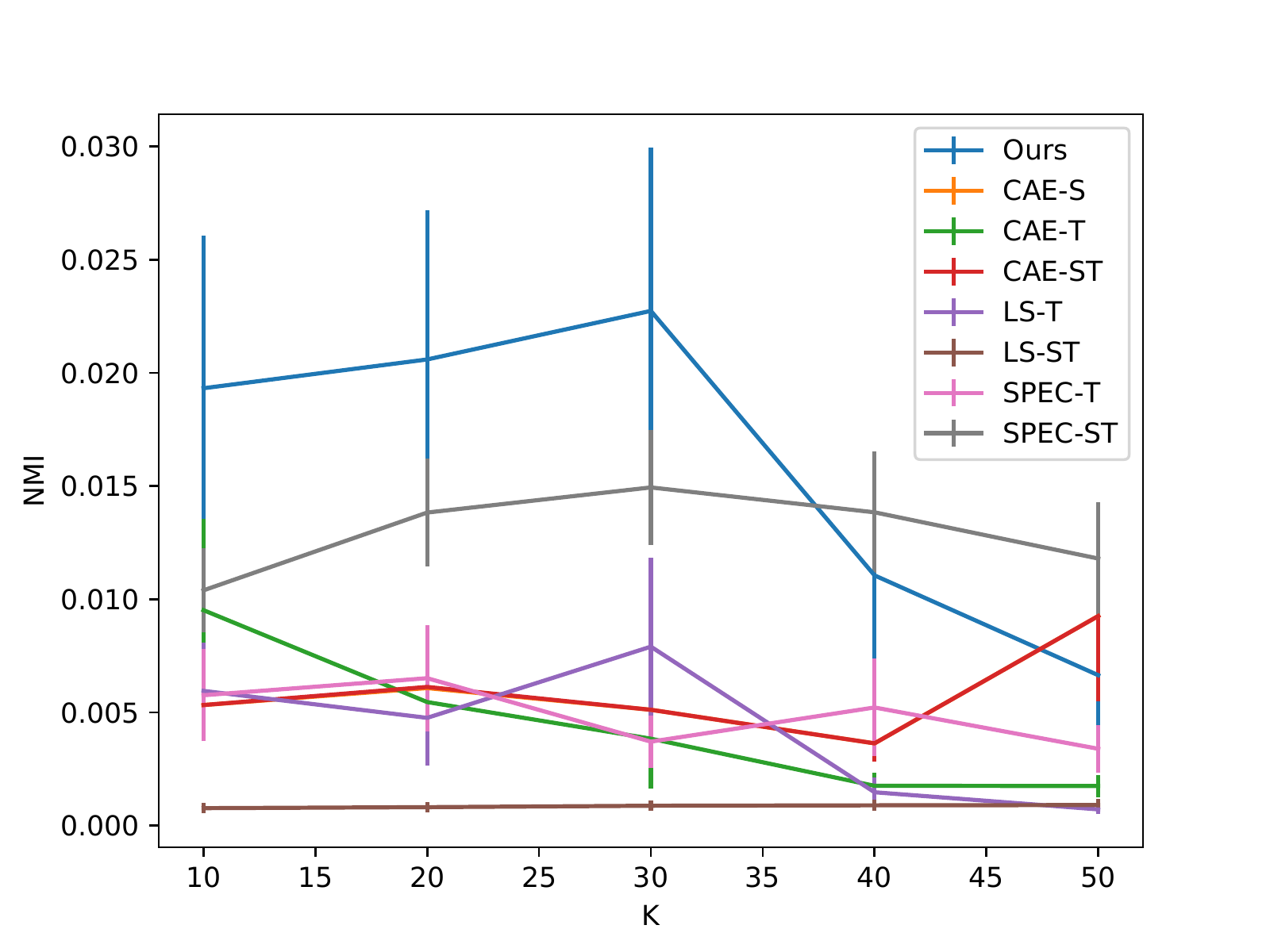}
      \subcaption{Amazon}
  \end{minipage}
   \begin{minipage}{0.245\hsize}
      \centering
      \includegraphics[width=3.85cm]{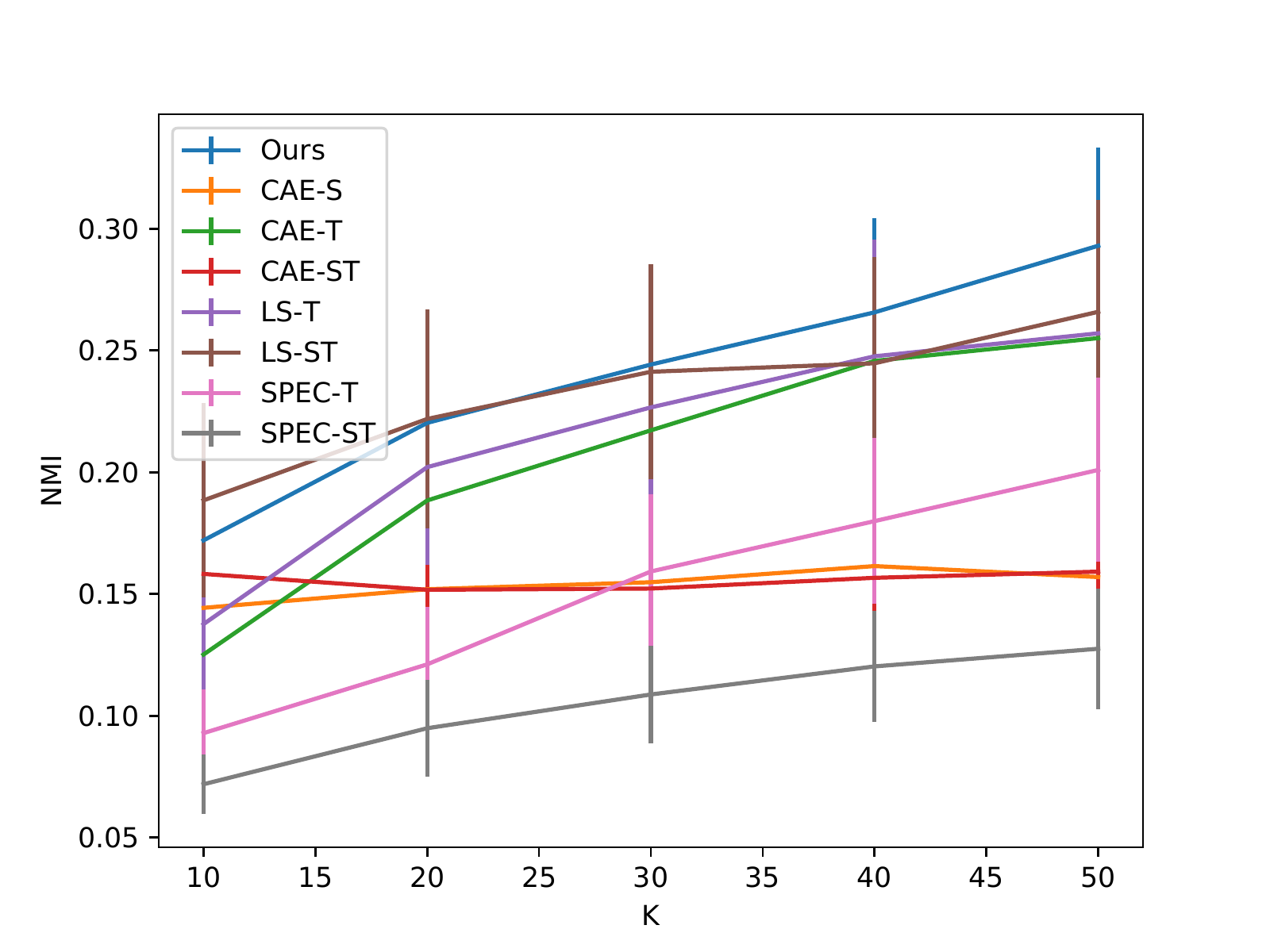}
      \subcaption{VLCS}
  \end{minipage}
  \caption{Average and standard errors of test NMIs with 2 target support instances when changing $K$.}
  \label{fig_nmi}
\end{figure*}

We qualitatively evaluated the proposed method by visualizing reconstructed images on MNIST-r.
We compared the proposed method with reconstruction-based methods (CAE).
Figure \ref{fig_images} shows three instances of test reconstructed images on a target task and the selected features.
The proposed method was able to reconstruct images from selected features more accurately than others.
The features selected by the proposed method were evenly distributed near the center.
Since all digits in this dataset are placed near the center, the proposed method captured the important information of digits well.
CAE-T was not able to reconstruct all images because it was trained with only four target images.
The features selected by CAE-T seem to be over-fitted to a few target images.
CAE-ST destroyed the reconstructed images obtained by CAE-S since fine-tuning with few target instances is difficult for CAE.
From these results, we found that the proposed method can select features from a few target instances that accurately reconstruct test images
by using useful knowledge on source tasks.

\paragraph{Clustering Ability}
We investigated the test ARIs and NMIs to evaluate the quality of selected features on clustering problems, which are commonly used to evaluate unsupervised feature selection abilities~\citep{li2017reconstruction,yang2011l2,zhao2007spectral,belkin2002laplacian}.
Tables \ref{table:ari} and \ref{table:nmi} show the averages and standard deviations of test ARIs and test NMIs over different numbers of target support instances and selected features,
respectively.
The proposed method showed the best ARIs and NMIs with all datasets.
Especially, it outperformed methods trained with the target support set only (CAE-T, LS-T, and SPEC-T) by a large margin in almost all cases,
which indicates that the difficulty of feature selection with a few instances and the effectiveness of using instances in source tasks.
Although non-neural network based methods (LS and SPEC) are generally suitable for small instance problems,
they performed worse than the proposed method.
Although CAE-ST performed worse than CAE-S in terms of reconstruction errors, it did not change test ARIs well.
This result suggests that the parameters of Concrete random variables are difficult to change drastically with a few instances by fine-training.
Note that similar results were obtained
even if only parameters of the feature selector were learned during fine-tuning.
In contrast, the proposed method can use information of the target support set effectively 
since our model is designed for few-shot feature selection.
Therefore, it performed well.
Figures \ref{fig_ari} and \ref{fig_nmi} show test ARIs and NMIs with $N_{\rm S}=2$ when varying the value of $K$ for each dataset.
We found that the proposed method performed well in all cases.
It also performed well when $N_{\rm S}=4$ and $6$, which is described in the supplemental material.

We conducted an ablation study to investigate the effects of our feature selector and decoder.
We compared our model with three models. 
w/o ${\bf r} ({\cal S})$ is our model without ${\bf r} ({\cal S})$ in Eq. \eqref{recon_model}, which uses the task-invariant decoder for all tasks.
w/o ${\boldsymbol \alpha} ({\cal S})$ is our model without ${\boldsymbol \alpha} ({\cal S})$ in Eq. \eqref{perm_net}, which 
uses the task-invariant (support set-independent) Concrete random variables ${\boldsymbol \alpha}$.
CAE-S is equivalent to our model without both ${\bf r} ({\cal S})$ and ${\boldsymbol \alpha} ({\cal S})$.
The upper half of Table \ref{table:t6} shows the average of ARIs and NMIs over all datasets, target support instances, and selected features.
w/o ${\boldsymbol \alpha} ({\cal S})$ and CAE-S did not perform well because both methods do not have any mechanism to perform task-specific feature selection.
In contrast, Ours and w/o ${\bf r} ({\cal S})$ performed well and showed similar results. 
This result indicates that ${\boldsymbol \alpha} ({\cal S})$ is particularly important for our model to select relevant features.
Note that w/o ${\bf r} ({\cal S})$ is also included in our proposal.
For further investigation, we additionally evaluated performance with few selected features $(K = 2$). In this case, Ours outperformed w/o ${\bf r} ({\cal S})$.
This is because instances are difficult to reconstruct from few features by a single decoder. By using task-specific decoders,
Ours performed well.
More detailed analysis of our model such as effects of temperature annealing and Concrete random variables are described in the supplemental material. These results show that both temperature annealing and randomness of Concrete random variables are useful for the proposed method and CAE.
\begin{table}[t!]
\begin{minipage}{0.55\hsize}
\centering
\caption{Ablation study. ARI (NMI) means the average of test ARIs (NMIs) [\%] over all datasets, target support instances within $\{2,4,6\}$, and selected features within $\{10,20,30,40,50\}$. ($K=2$) means the results with two selected features.}
\label{table:t6}
\scalebox{0.85}{
\begin{tabular}{lrrrr}
\hline
& Ours &  \multicolumn{1}{c}{w/o ${\bf r} ({\cal S})$} & \multicolumn{1}{c}{w/o ${\boldsymbol \alpha} ({\cal S})$} & CAE-S \\
\hline
ARI & 19.0 & 19.2 & 15.7 & 15.6 \\
NMI & 32.2 & 32.3 & 28.5 & 28.9  \\
\hline
ARI ($K=2$) & 6.5 & 4.4 & 5.1 & 4.8 \\
NMI ($K=2$) & 16.6 & 15.8 & 14.4 & 14.4  \\
\hline\end{tabular}}
\end{minipage}
~~
\begin{minipage}{0.44\hsize}
\centering
\caption{The average number of features selected by the proposed method after deduplication.}
\label{table:t5ded}
\scalebox{0.85}{
\begin{tabular}{lrrrrr}
\hline
Data \textbackslash \ $K$ & \multicolumn{1}{c}{10} &  \multicolumn{1}{c}{20} & \multicolumn{1}{c}{30} & \multicolumn{1}{c}{40} &  \multicolumn{1}{c}{50} \\
\hline
MNIST-r & 10.0 & 20.0 & 29.7 & 39.1 & 47.0 \\
Isolet & 10.0 & 20.0 & 29.8 & 40.0 & 49.9 \\
Amazon & 6.8 & 12.9 & 17.9 & 20.1 & 30.1 \\
VLCS & 10.0 & 19.9 & 30.0 & 40.0 & 50.0 \\
\hline\end{tabular}}
\end{minipage}
\end{table}

\paragraph{Deduplicated Features Selected by the Proposed Method}
We investigated the average number of deduplicated features selected by the proposed method.
Although the proposed method uses $K$ Concrete random variables to select features
in the feature selector, 
it possibly selects fewer features than $K$.
Table~\ref{table:t5ded} shows the average number of features selected by the proposed method after deduplication when $N_{{\cal S}}=2$.
For MNIST-r, Isolet, and VLCS, the proposed method selected almost $K$ features.
For Amazon, it selected fewer features than $K$.
This would be because Amazon has few important features.
Indeed, the proposed method shows good test MSRE, ARI, and NMI even if $K=10$ in Figures \ref{fig_recon}, \ref{fig_ari}, and \ref{fig_nmi}.

\section{Conclusions}
In this paper, we proposed a few-shot learning method for unsupervised feature selection.
Our model can perform target task-specific feature selection given a few target instances by using useful knowledge in source tasks.
Experimental results showed that the proposed method outperformed other feature selection methods.
For future work, we plan to extend our framework to supervised feature selection problems.
Finally, we describe a potential negative social impact of this work.
Since the proposed method can use various datasets for training, 
there is a risk that users might include biased datasets without careful thought in source tasks, 
which might result in biased feature selection. Therefore. we encourage research to automatically detect biased datasets.


\appendix

\section{Data Details}
We used four real-world datasets: MNIST-r\footnote{https://github.com/ghif/mtae}, Isolet\footnote{http://archive.ics.uci.edu/ml/datasets/ISOLET}, Amazon\footnote{http://multilevel.ioe.ac.uk/intro/datasets.html}, and VLCS\footnote{http://www.cs.dartmouth.edu/chenfang/projpage/FXR iccv13/index.php}.
MNIST-r is commonly used in domain generalization studies
\citep{ghifary2015domain}. 
This dataset, which
was derived from the hand-written digit dataset MNIST, was introduced by Ghifary et al
\citep{ghifary2015domain}.
Each task is created by rotating the images in multiples of 15 degrees: 0, 15, 30, 45, 60, and 75.
Therefore, this dataset has six tasks.
Each task has 1,000 images, which are represented by 256-dimensional vectors, of 10 classes (digits).

Isolet is a widely used real-world dataset in multi-task learning studies
\citep{parameswaran2010large,jebara2004multi}. 
This dataset is collected from 150 speakers who say each letter of the Roman alphabet twice.
Thus, there are 52 instances from each individual.
The individuals are grouped into five groups on the basis of speaking similarity, thus, this dataset has five tasks.
instances are represented as 617-dimensional vectors of 26 classes (letters).

Amazon is a widely used real-world dataset for cross-domain sentiment analysis
\citep{gong2013connecting}. 
This dataset consists of product reviews in four tasks: kitchen appliances, DVDs, books, and electronics.
We used the processed data from Gong et al.
\citep{gong2013connecting},
in which the dimensionality of the bag-of-words features was 400.
Each task has 1000 positive and 1000 negative reviews (two classes).

VLCS is also commonly used in domain adaptation studies
\citep{motiian2017unified}. 
This dataset consists of real images
on four tasks: VOC2007, LabelMe, Caltech-101,
and SUN09. Each task shares five object classes:
bird, car, chair, dog, and person. This dataset has
10,729 images in total. Instead of using the raw features, we used the $\ell ^2$-normalized DeCaF-6 features
\citep{donahue2014decaf}
as input features following previous studies
\citep{ghifary2015domain,motiian2017unified}.
These features have 4096 dimensions.

\section{Experimental Settings Details}
We used mean squared reconstruction error (MSRE) for evaluating the reconstruction ability.
For clustering, we used two metrics for evaluating each method in clustering tasks: Adjusted Rand Index (ARI) and Normalized Mutual Information (NMI).
Both metrics are widely used to evaluate clustering results.
In both metrics, values close to one indicate high quality clustering.
For the K-means, we set the number of clusters to that of classes for each dataset.

For the proposed method and concrete autoencoder (CAE),
we used two-hidden-layer neural networks with $32$ hidden units and rectified linear unit (ReLU) activation for the decoders $h_{\theta}$
with MNIST-r, Isolet, and Amazon.
For VLCS, we used two-hidden layer neural networks with $256$ hidden units and ReLU activation for $h_{\theta}$ because 
this dataset has higher dimensionality of features than other datasets.
For the activation of output layers, we used the sigmoid function for MNIST-r, Amazon, and VLCS 
since features are normalized to $[0,1]$ in these datasets.
For Isolet, the tanh function was used since this dataset takes feature values within $[-1,1]$.
For the permutation-invariant neural networks for Concrete random variables and decoders in the proposed method,
we used different one-layer neural networks with $64$ output units and ReLU activation for $f_{\phi_1}$ and $f_{\psi_1}$.
For VLCS, we used different one-layer neural networks with $512$ output units and ReLU activation for $f_{\phi_1}$ and $f_{\psi_1}$.
We used different one-layer neural networks for $g_{\phi_2}$ and $g_{\psi_2}$ with all datasets.
The dimension of parameters ${\boldsymbol \pi} ^{(k)}$ was set to $300$.
The output size of the permutation-invariant neural network for the decoder $g_{\psi_2}$ was set to one.
For the proposed method and CAE, we set the minibatch size to $64$.
(For the proposed method, we set $N_{\rm S} + N_{\rm Q} =64$.)
The initial temperature parameter $T_0$ was set to $10$ and the final temperature $T_1$ to $0.01$.
We used the Adam optimizer
\citep{kingma2014adam}
with a learning rate of 0.001, 
and the maximum number of iterations was 50,000.
For the proposed method, CAE-S, and CAE-T, we used early-stopping on the basis of the validation reconstruction errors to avoid overfitting.
For CAE-ST, the number of iterations for fine-turning was set to a large value (1000) 
because few iterations were not able to change the selected features obtained by CAE-S.
\begin{table*}[t]
\caption{Effects of annealing schedule for temperature parameter $\tau$. Average of test ARIs [\%] and NMIs [\%] over all datasets, target support instances within $\{ 2,4,6 \}$, and selected features within $\{ 10,20,30,40,50 \}$, respectively. 
}
\label{table:anneal}
\centering
\scalebox{1.0}{
\begin{tabular}{lrrrrrrrr}
\hline 
\multicolumn{1}{c}{} &  &  \multicolumn{1}{c}{Ours} & \multicolumn{1}{c}{Ours} & \multicolumn{1}{c}{Ours} & 
 & \multicolumn{1}{c}{CAE-S} & \multicolumn{1}{c}{CAE-S} & \multicolumn{1}{c}{CAE-S} \\

\multicolumn{1}{c}{} & \multicolumn{1}{c}{Ours}  &  \multicolumn{1}{c}{($\tau = 0.01$)} & \multicolumn{1}{c}{($\tau=1$)} & \multicolumn{1}{c}{($\tau=10$)} & 
\multicolumn{1}{c}{CAE-S} & \multicolumn{1}{c}{($\tau=0.01$)} & \multicolumn{1}{c}{($\tau=1$)} & \multicolumn{1}{c}{$(\tau=10)$} \\
\hline
ARI & \textbf{19.0} & 7.4 & 17.0 & 15.1 & 15.6  & 13.2 & 15.3 & 12.6  \\
NMI & \textbf{32.2} & 17.8 & 29.6 & 26.8 & 28.9  & 25.9 & 28.3 & 24.4  \\
\hline
\end{tabular}
}
\end{table*}

For LS-T and LS-ST, the candidates of heat kernel parameters were $\{0.1,1,10,100\}$, and the best value for testing data was selected.
For LS-ST, the number of nearest neighbors was set to five following previous studies
\citep{he2006laplacian,li2017reconstruction,yang2011l2}.
For LS-T, the best value was selected within $\{1,3,5\}$ because five nearest neighbors are not defined for a few target support instances. 
For SPEC-T and SPEC-ST, we used the second feature ranking function that has been reported to robustly perform well~\citep{zhao2007spectral}.

For MNIST-r and Isolet, 
we randomly chose one task for a target task and the rest for source tasks.
From the source tasks, we randomly chose one task for validation and the rest for training.
Then, we randomly chose 80\% of instances from each training/validation task.  
For Amazon and VLCS, we used 80\% of instances from each source task for training and the rest for validation 
since these datasets have only four tasks. 
For each target task, we randomly chose a few instances for the target support set and the remainder for testing.
Specifically, we evaluated the performance by changing the number of target support instances within $\{2,4,6\}$.
For each case, we randomly created 20 different datasets and calculated an average test MSRE, ARI, and NMI.

\section{Additional Experimental Results}
\subsection{Effects of Annealing Schedule for Temperature Parameter $\tau$}
We investigated the effect of the annealing schedule for temperature parameter $\tau$, which is used for the proposed method and CAE in the main paper.
Table \ref{table:anneal} shows the average of test ARIs and NMIs over all datasets, target support instances, and selected features.
Ours and CAE-S used the annealing of temperature $\tau$ from $10$ to $0.01$ during training.
Other methods did not use the annealing and instead used a fixed value of $\tau$ during training.
First, Ours, which uses the annealing, performed the best.
For the proposed method, Ours ($\tau=0.01$) and Ours ($\tau=10$) were not able to perform as well as Ours.
This is because although small $\tau$ can select individual features in each Concrete random variable, it cannot explore various combinations of features and converge 
to a poor local minimum, and large $\tau$ cannot select individual features in each Concrete random variable during training.
Ours ($\tau=1$) outperformed both methods with the suitable fixed value of $\tau$ but was outperformed by Ours with the annealing.
By using the annealing of $\tau$, the proposed method can explore various combinations of features in the initial phase of training and converge to a good local minimum in the last phase of training.
For CAE, the results were similar to the proposed method.
Overall, these results indicate that the proposed method can improve feature selection performance by using the annealing schedule of temperature parameters.
\begin{table}[t]
\caption{Effects of randomness in the feature selector of our model. Average of test ARIs [\%] and NMIs [\%] over all datasets, target support instances within $\{ 2,4,6 \}$, and selected features within $\{ 10,20,30,40,50 \}$, respectively. 
}
\label{table:rand}
\centering
\scalebox{0.95}{
\begin{tabular}{lrrrr}
\hline 
\multicolumn{1}{c}{} &  &  \multicolumn{1}{c}{Ours} &  & \multicolumn{1}{c}{CAE-S}  \\
\multicolumn{1}{c}{} & \multicolumn{1}{c}{Ours}  &  \multicolumn{1}{c}{w/o rand.} & CAE-S & \multicolumn{1}{c}{w/o rand.}  \\
\hline
ARI & \textbf{19.0} & 16.9 & 15.6 & 14.5  \\
NMI & \textbf{32.2} & 29.1 & 28.9 & 26.6  \\
\hline
\end{tabular}
}
\end{table}

\subsection{Effects of Randomness in the Feature Selector}
We investigated the effect of randomness in the feature selector (Concrete random variables), which is used for the proposed method and CAE in the main paper.
The feature selector has random variables $g_m$.
Therefore, we investigate the role of this random variable.
Table \ref{table:rand} shows the average of test ARIs and NMIs over all datasets, target support instances, and selected features.
Ours and CAE-S used the random variables $g_m$ although Ours w/o rand. and CAE-S w/o rand. did not.
Ours performed the best.
Ours (CAE-S) performed better than Ours w/o rand. (CAE-S w/o rand.).
By using the random variables $g_m$, the proposed method would explored informative feature combinations during training while avoiding to being trapped at a poor local minimum.
These results indicate the effectiveness of randomness in the feature selector in our framework.

\begin{figure*}[t]
  \begin{minipage}{0.24\hsize}
      \centering
      \includegraphics[width=3.8cm]{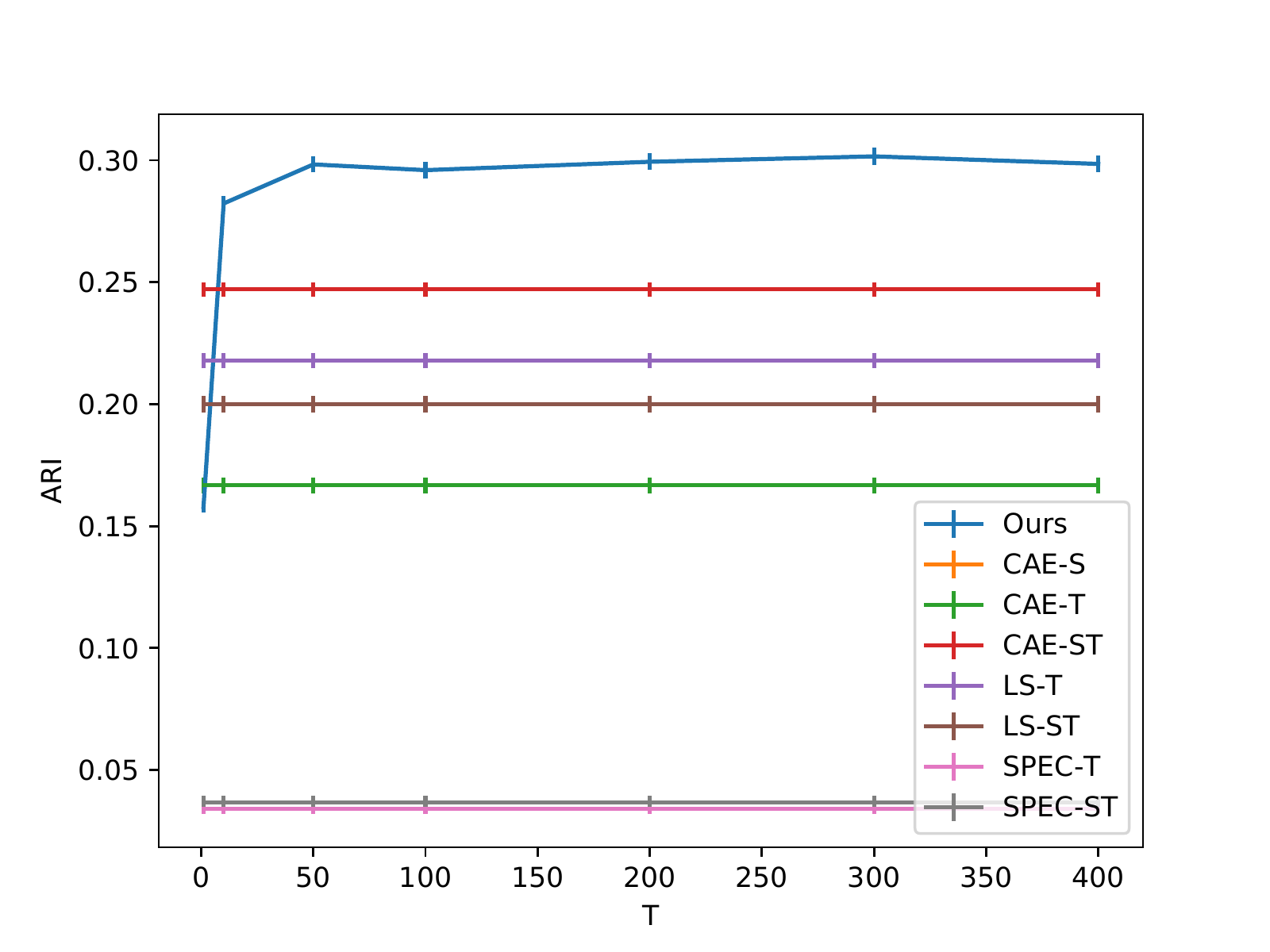}
      \subcaption{MNIST-r}
  \end{minipage}
  \begin{minipage}{0.24\hsize}
      \centering
      \includegraphics[width=3.8cm]{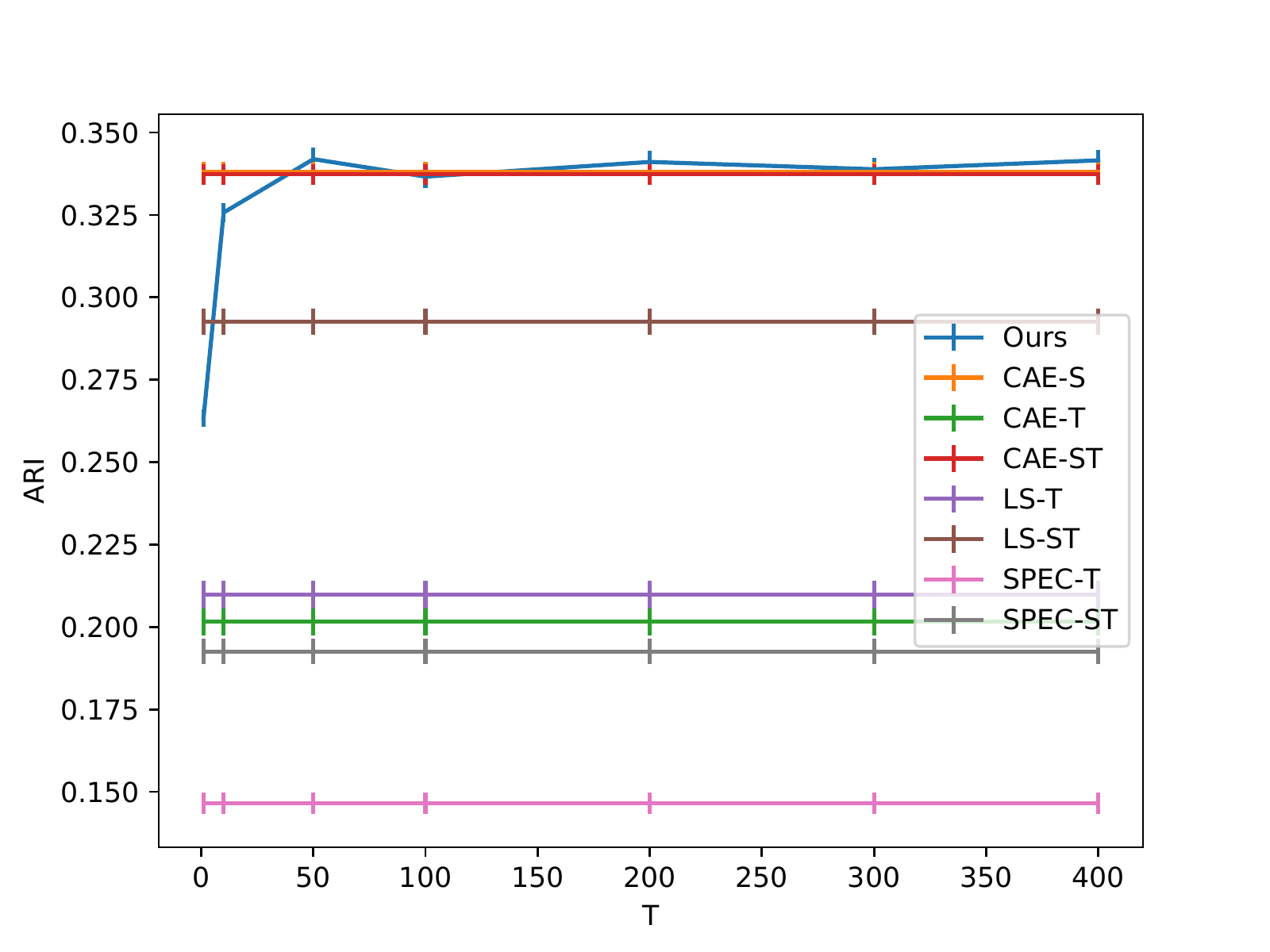}
      \subcaption{Isolet}
  \end{minipage}
  \begin{minipage}{0.24\hsize}
      \centering
      \includegraphics[width=3.8cm]{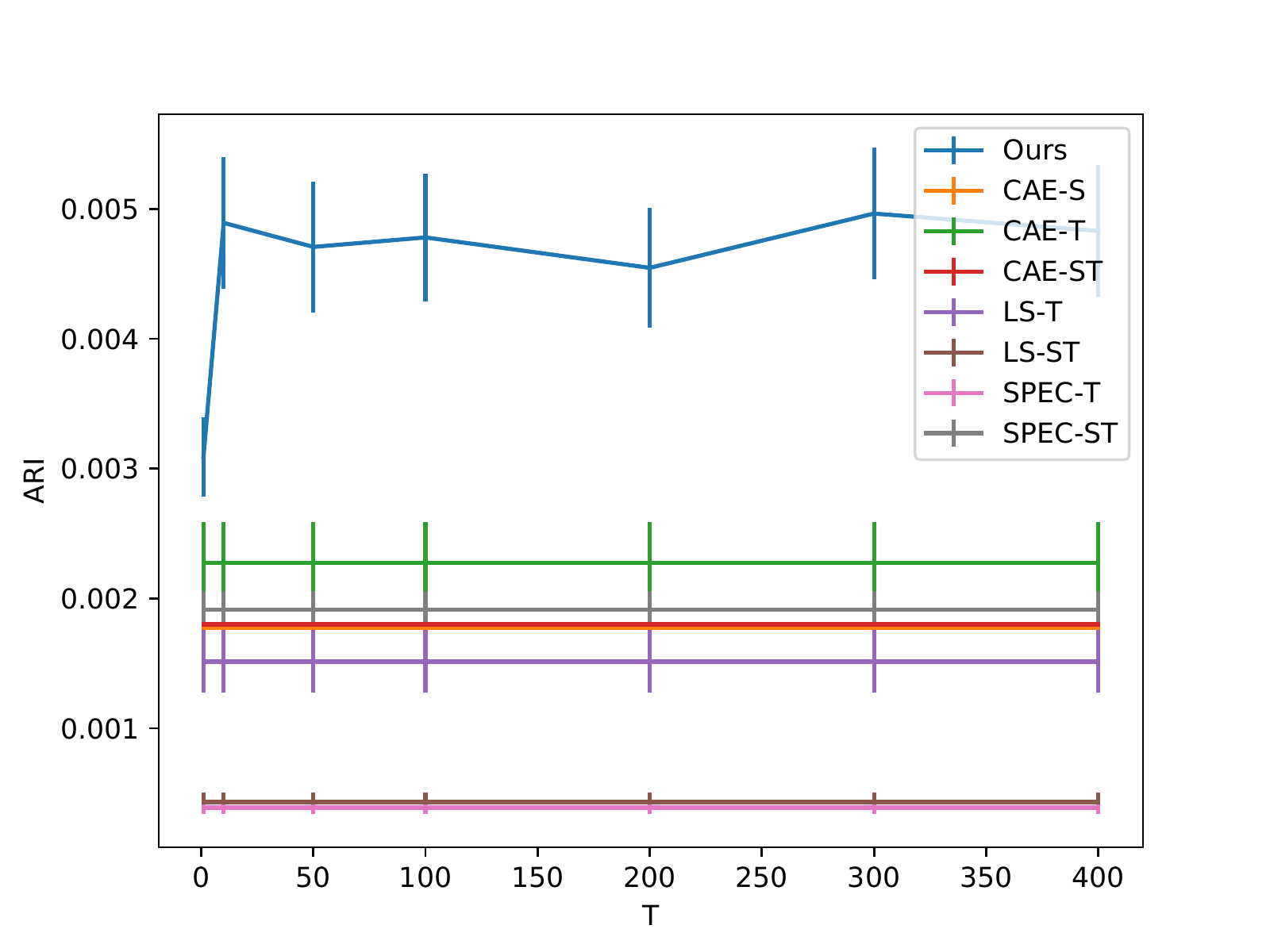}
      \subcaption{Amazon}
  \end{minipage}
   \begin{minipage}{0.24\hsize}
      \centering
      \includegraphics[width=3.8cm]{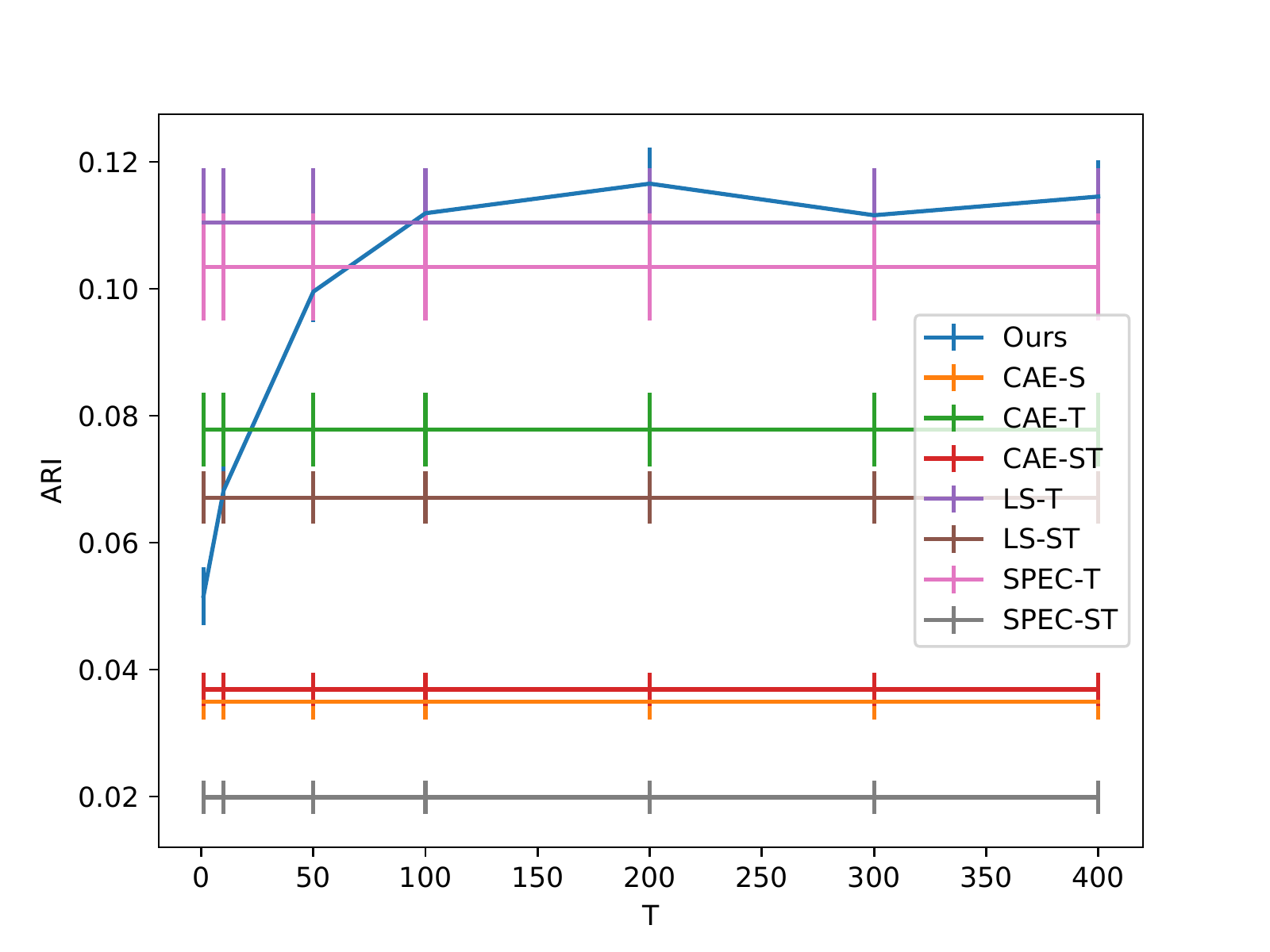}
      \subcaption{VLCS}
  \end{minipage}
  \caption{Effect of dimension of parameter ${\boldsymbol \pi} ^{(k)}$ for test ARIs. Average and standard errors of test ARIs over different $N_{\rm S}$ and $K$ when $T$ was changed.}
  \label{fig_T_change_ari}
\end{figure*}

\begin{figure*}[!]
  \begin{minipage}{0.24\hsize}
      \centering
      \includegraphics[width=3.8cm]{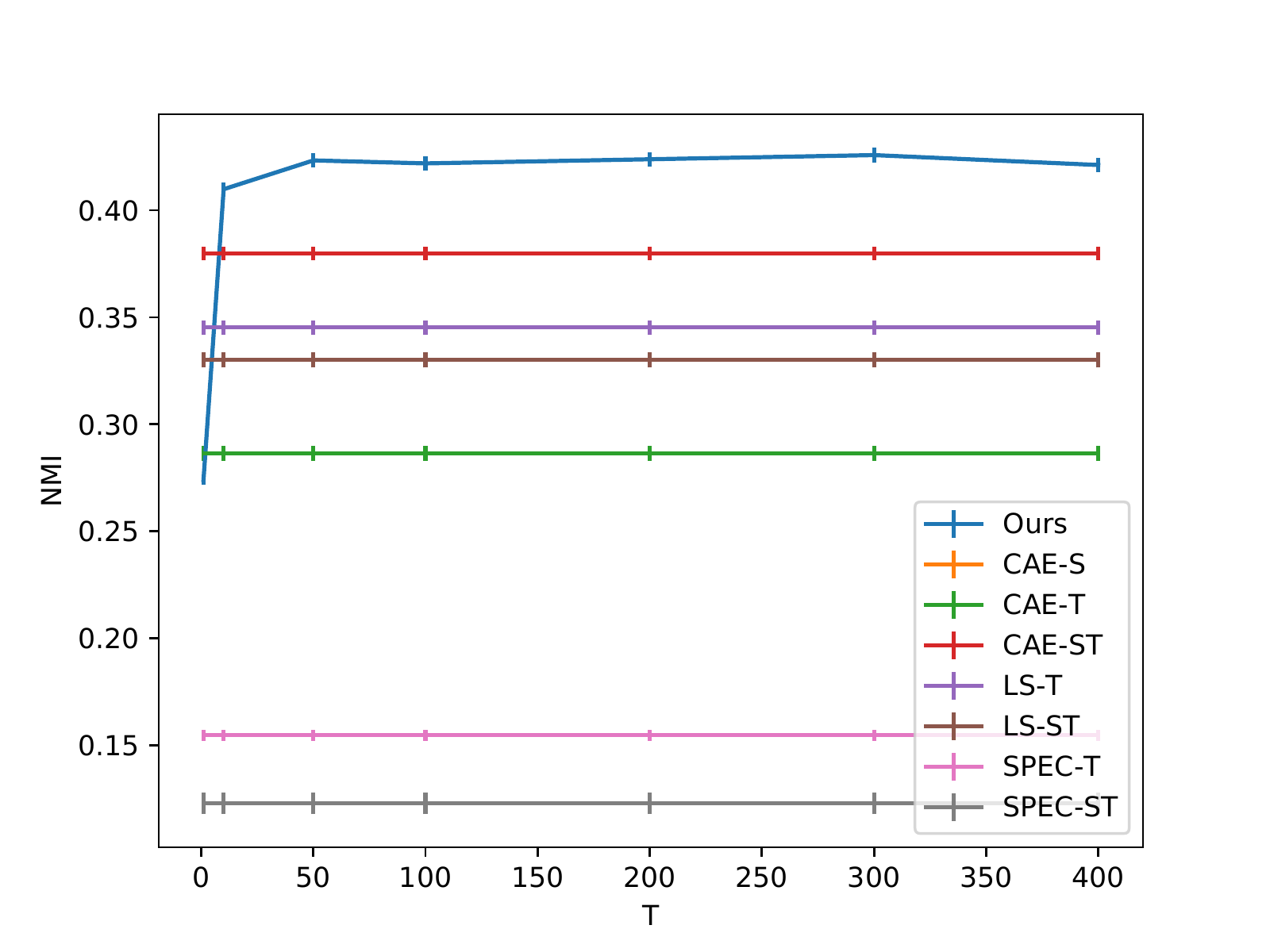}
      \subcaption{MNIST-r}
  \end{minipage}
  \begin{minipage}{0.24\hsize}
      \centering
      \includegraphics[width=3.8cm]{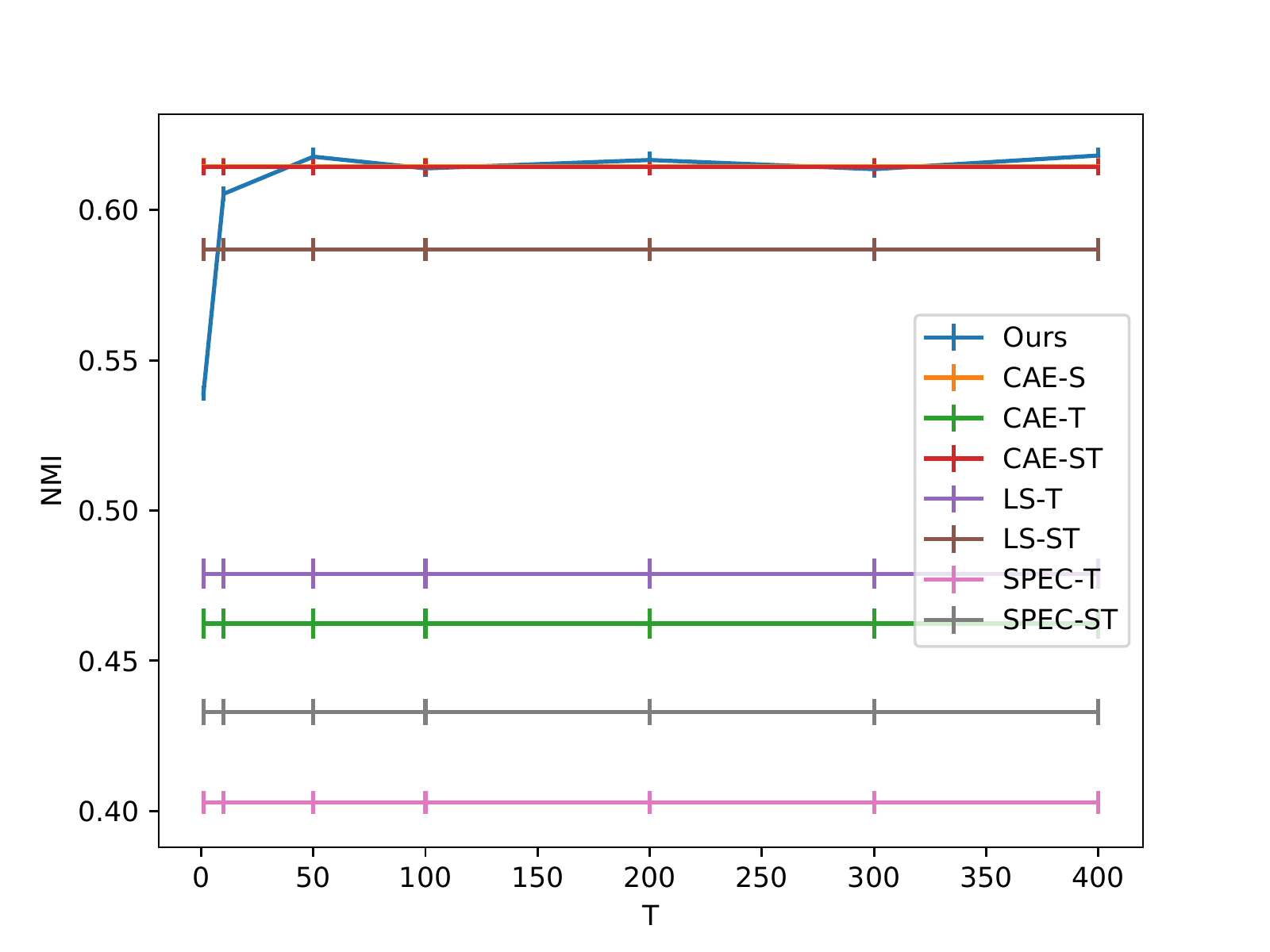}
      \subcaption{Isolet}
  \end{minipage}
  \begin{minipage}{0.24\hsize}
      \centering
      \includegraphics[width=3.8cm]{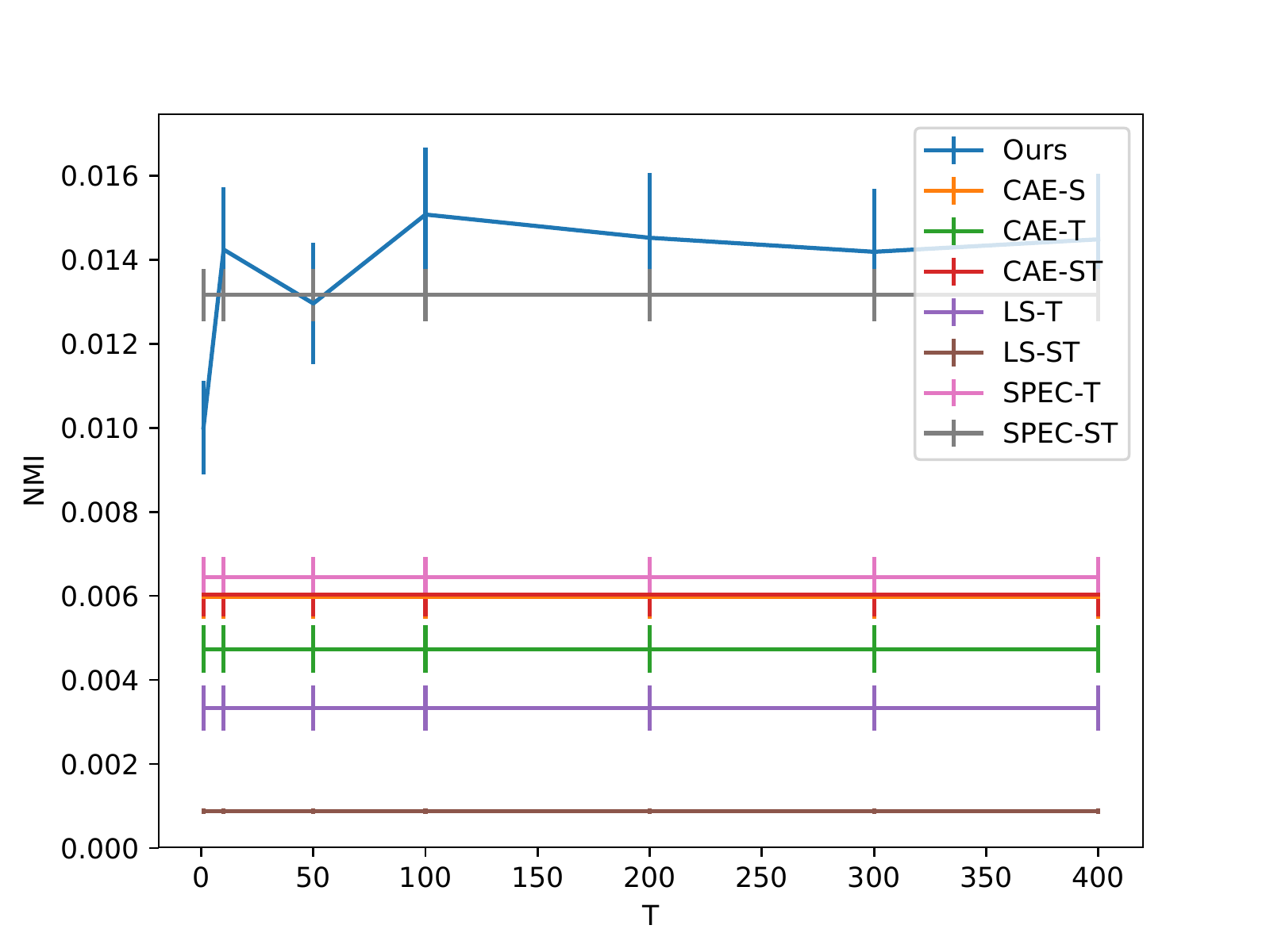}
      \subcaption{Amazon}
  \end{minipage}
   \begin{minipage}{0.24\hsize}
      \centering
      \includegraphics[width=3.8cm]{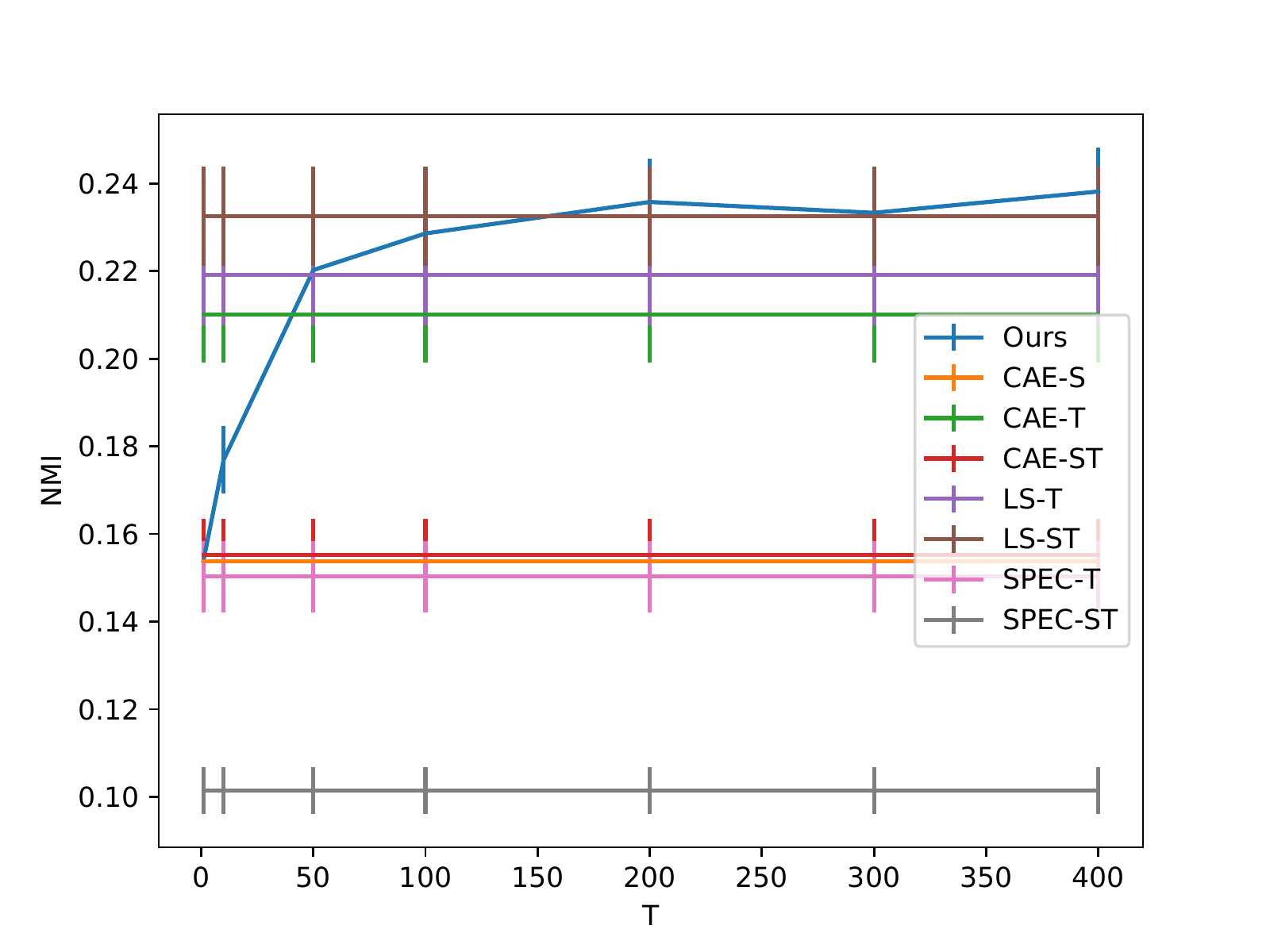}
      \subcaption{VLCS}
  \end{minipage}
  \caption{Effect of dimension of parameter ${\boldsymbol \pi} ^{(k)}$ for test NMIs. Average and standard errors of test NMIs over different $N_{\rm S}$ and $K$ when $T$ was changed.}
  \label{fig_T_change_nmi}
\end{figure*}

\subsection{Dependency of the Dimension of Parameters ${\boldsymbol \pi} ^{(k)}$}
We investigated the dependency of the dimension of parameters ${\boldsymbol \pi} ^{(k)} \in \mathbb{R}^{T}$ in the proposed method.
These parameters are used for varying the class probability of each Concrete random variable, which is necessary for selecting different features via
the feature selector.
Figures \ref{fig_T_change_ari}--\ref{fig_T_change_nmi} 
show the average and standard errors of test ARIs and NMIs over different numbers of target support instances and selected features 
when varying $T$ within $\{1,10,50,100,200,300,400\}$, respectively.
For all evaluation measures and datasets, the proposed method consistently performed well when the value of $T$ was larger than $100$.
When the value of $T$ was small, the proposed method was not able to perform well in our experiments.
This would be because the small dimension is not enough to change the class probabilities of each Concrete variable.
These results show that the proposed method can perform well when $T$ is set to a relatively large value.

\subsection{Additional Comparison Method}
We compared the proposed method with an additional variant of CAE-ST.
Although CAE-ST was trained with target instances by fine-tuning after pre-training with source instances, which is a standard technique for transfer learning,

CAE can be also trained with source and target instances, simultaneously (CAE-ST-sim).
Table \ref{table:ad_comp} shows the average of test ARIs and NMIs over all datasets, target support instances, and selected features.
The proposed method clearly outperformed CAE-ST-sim.
Since CAE-ST-sim does not have explicit mechanisms for few-shot learning, it did not work well.
\begin{table}
\caption{Comparison with CAE-ST-sim. Average of test ARIs [\%] and NMIs [\%] over all datasets, target support instances within $\{ 2,4,6 \}$, and selected features within $\{ 10,20,30,40,50 \}$, respectively. 
}
\label{table:ad_comp}
\centering
\scalebox{0.95}{
\begin{tabular}{lrrrr}
\hline 
\multicolumn{1}{c}{} & \multicolumn{1}{c}{Ours}  &  \multicolumn{1}{c}{CAE-S} & CAE-ST & \multicolumn{1}{c}{CAE-ST-sim}  \\
\hline
ARI & \textbf{19.0} & 15.6 & 15.6 & 16.9  \\
NMI & \textbf{32.2} & 28.9 & 28.9 & 27.7  \\
\hline
\end{tabular}
}
\end{table}

\begin{table}[t]
\caption{Computation time in seconds for training by each method. Ours-target represents the feature selection time for a target task of the proposed method, i.e., the total time of estimation of parameters of Concrete random variables from 
target support instances and applying the ${\argmax}$ operator to the estimated parameters.
}
\label{table:comp}
\centering
\scalebox{0.95}{
\begin{tabular}{rrrrrrrrr}
\hline 
\multicolumn{1}{c}{Ours} & \multicolumn{1}{c}{Ours-target}  &  \multicolumn{1}{c}{CAE-S} & \multicolumn{1}{c}{CAE-T} & \multicolumn{1}{c}{CAE-ST} & 
\multicolumn{1}{c}{LS-T} & \multicolumn{1}{c}{LS-ST} & \multicolumn{1}{c}{SPEC-T} & \multicolumn{1}{c}{SPEC-ST} \\
\hline
 303.725 & 0.003 & 125.065 & 119.699 & 4.673 & 0.003  & 1.493 & 0.011 & 14.627  \\
\hline
\end{tabular}
}
\end{table}

\subsection{Computation Cost}
We investigated the computation time of the proposed method on MNIST-r.
Table~\ref{table:comp} shows the training time of each method and target task-specific feature selection time of the proposed method on a computer with the 2.20GHz CPU.
CAE-ST represents the training time for fine-tuning.
Although our model took longer to train than the others,
it can perform fast and accurate target task-specific feature selection without re-training.

\subsection{Full Results of Test MSREs, ARIs, and NMIs}
Figures \ref{fig_recon}--\ref{fig_nmi} show the average and standard error of test MSREs, ARIs, and NMIs
when changing the number of selected features $K$ with different numbers of target support instances, respectively.
The proposed method performed well in almost all cases.

\subsection{Visualization of Test Reconstructed Images on MNIST-r}
\begin{figure*}[t]
  \begin{minipage}{1.00\hsize}
      \centering
      \includegraphics[width=14.0cm]{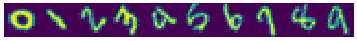}
  \subcaption{Original}
  \end{minipage}
    \begin{minipage}{1.00\hsize}
      \centering
      \includegraphics[width=14.0cm]{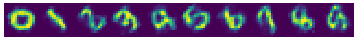}
  \subcaption{Ours}
  \end{minipage}
    \begin{minipage}{1.00\hsize}
      \centering
      \includegraphics[width=14.0cm]{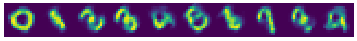}
  \subcaption{CAE-S}
  \end{minipage}
    \begin{minipage}{1.00\hsize}
      \centering
      \includegraphics[width=14.0cm]{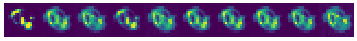}
  \subcaption{CAE-T}
  \end{minipage}
    \begin{minipage}{1.00\hsize}
      \centering
      \includegraphics[width=14.0cm]{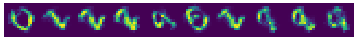}
  \subcaption{CAE-ST}
  \end{minipage}
  \caption{Ten instances of test reconstructed images when 4 target support instances and $20$ selected features (pixels) were used on MNIST-r.}
  \label{fig_supp_mnist}
\end{figure*}

Figure \ref{fig_supp_mnist} shows the 10 instances (digits from 0 to 9) of test reconstructed images on a target task when 
4 target support instances and 20 selected features were used on MNIST-r.
The proposed method was able to reconstruct images from selected features more accurately than the others.

\color{black}

\begin{figure*}[t]
  \begin{minipage}{1.00\hsize}
  \begin{minipage}{0.33\hsize}
      \centering
      \includegraphics[width=5.2cm]{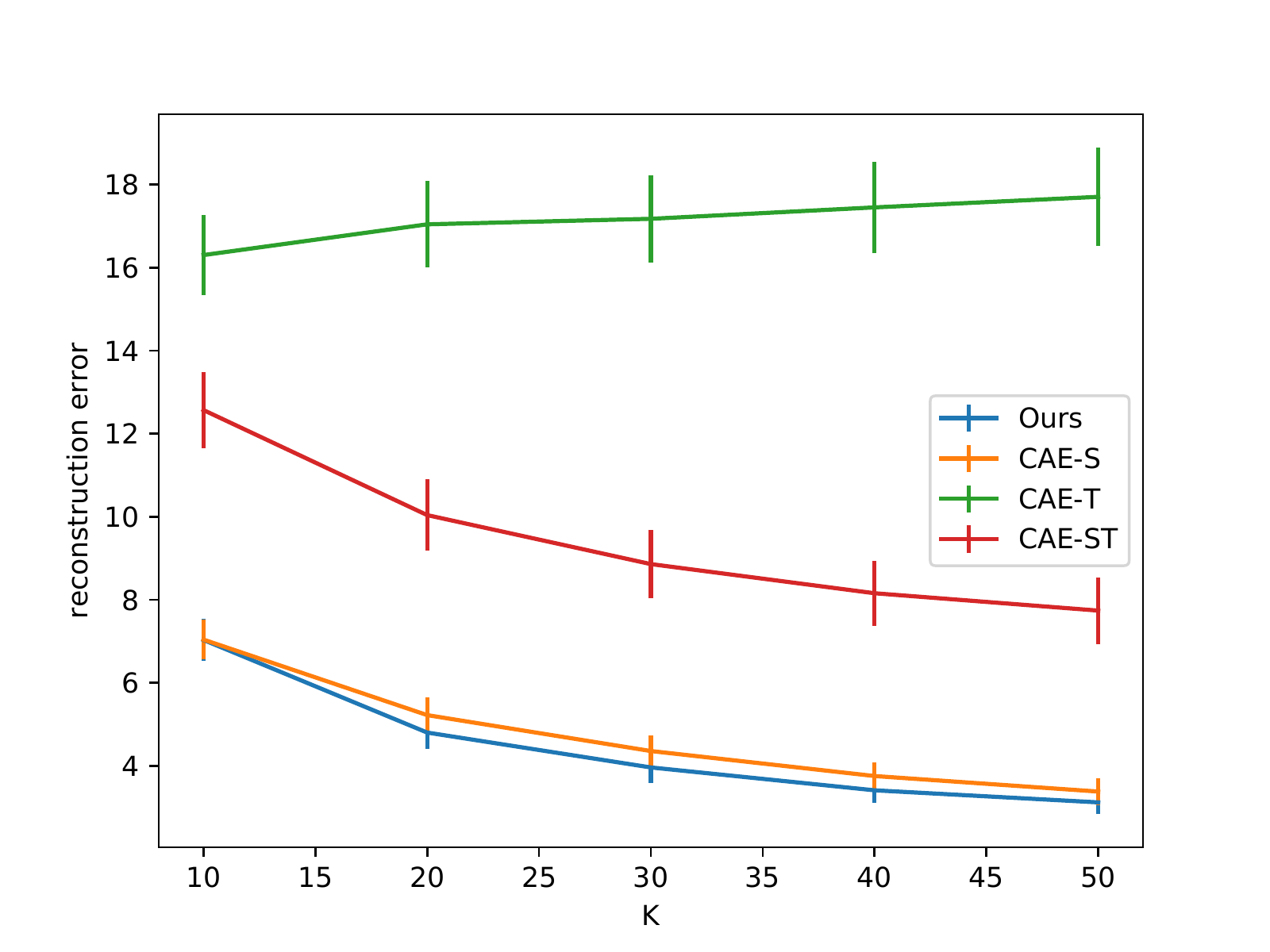}
  \end{minipage}
  \begin{minipage}{0.33\hsize}
      \centering
      \includegraphics[width=5.2cm]{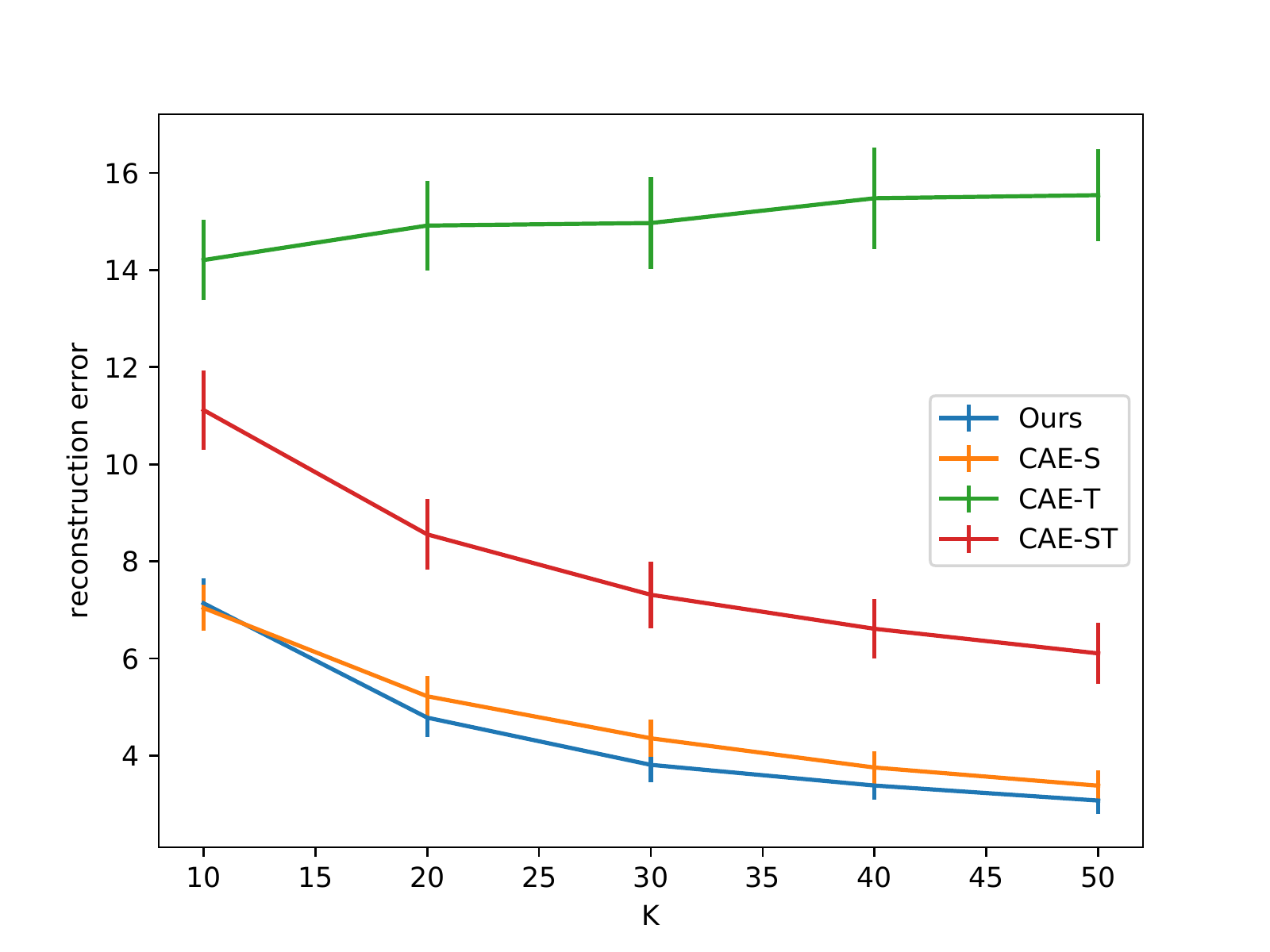}
  \end{minipage}
  \begin{minipage}{0.33\hsize}
      \centering
      \includegraphics[width=5.2cm]{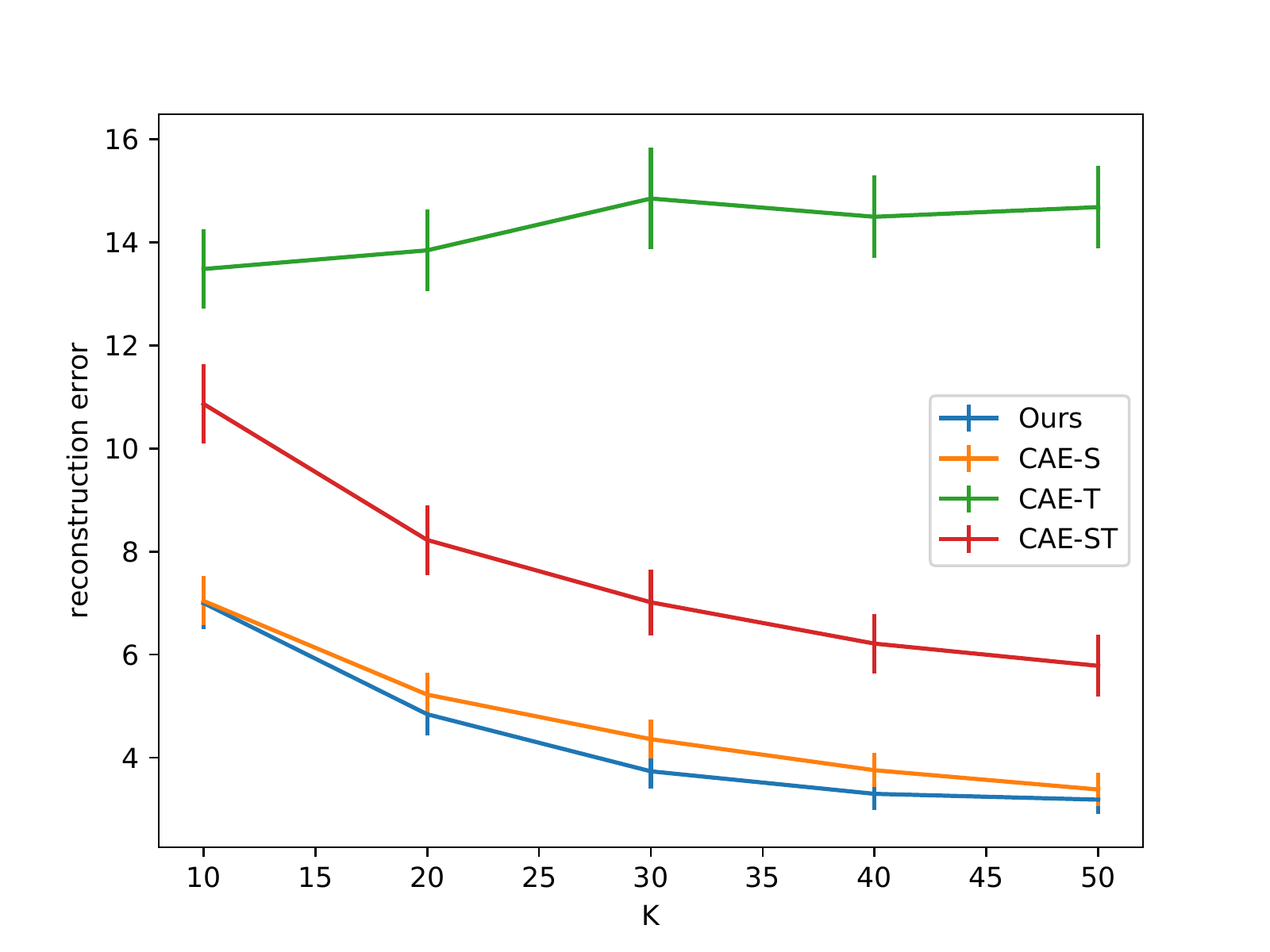}
  \end{minipage} 
  \subcaption{MNIST-r}
  \end{minipage}
  \begin{minipage}{1.00\hsize}
  \begin{minipage}{0.33\hsize}
      \centering
      \includegraphics[width=5.2cm]{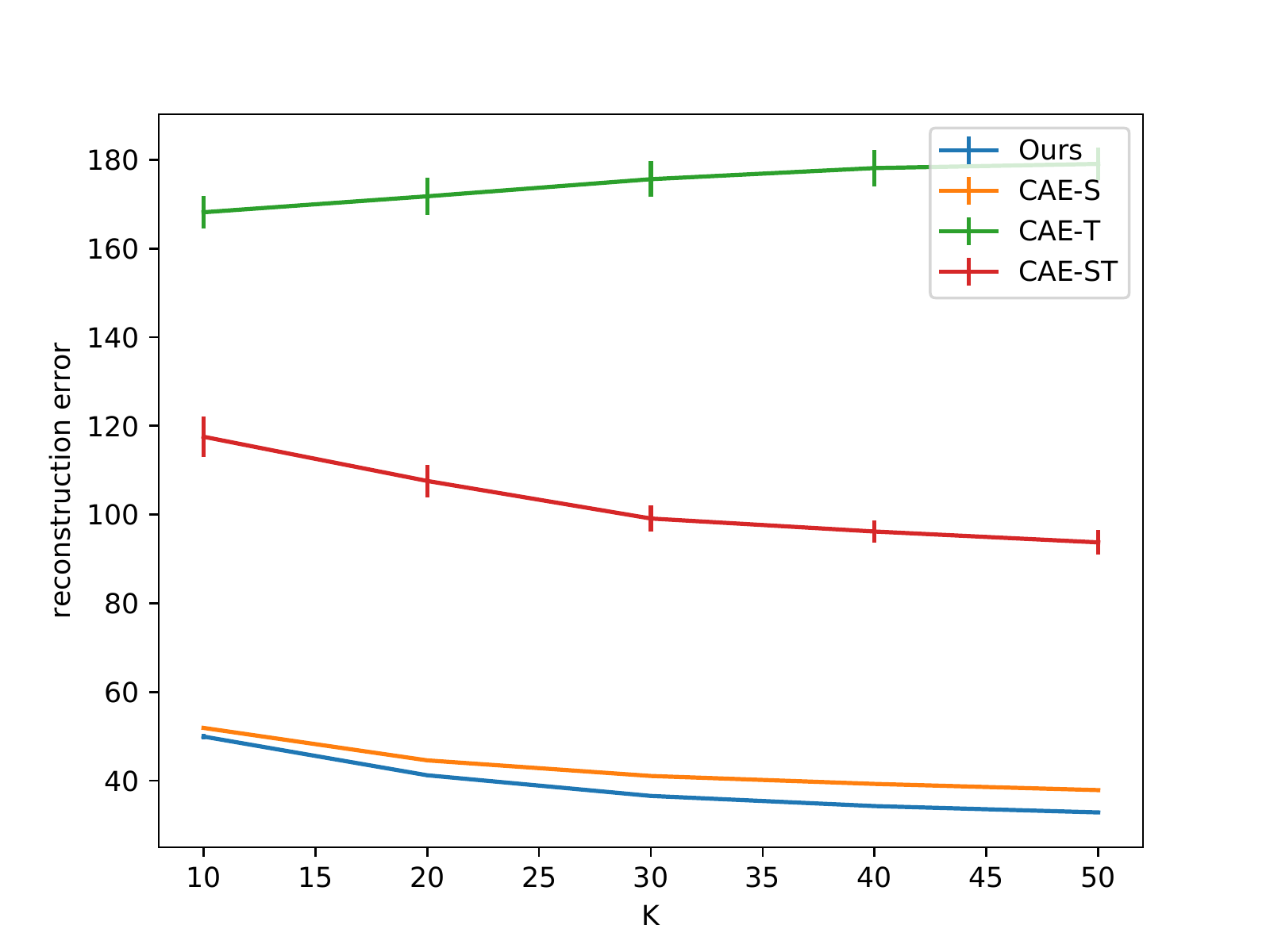}
  \end{minipage}
  \begin{minipage}{0.33\hsize}
      \centering
      \includegraphics[width=5.2cm]{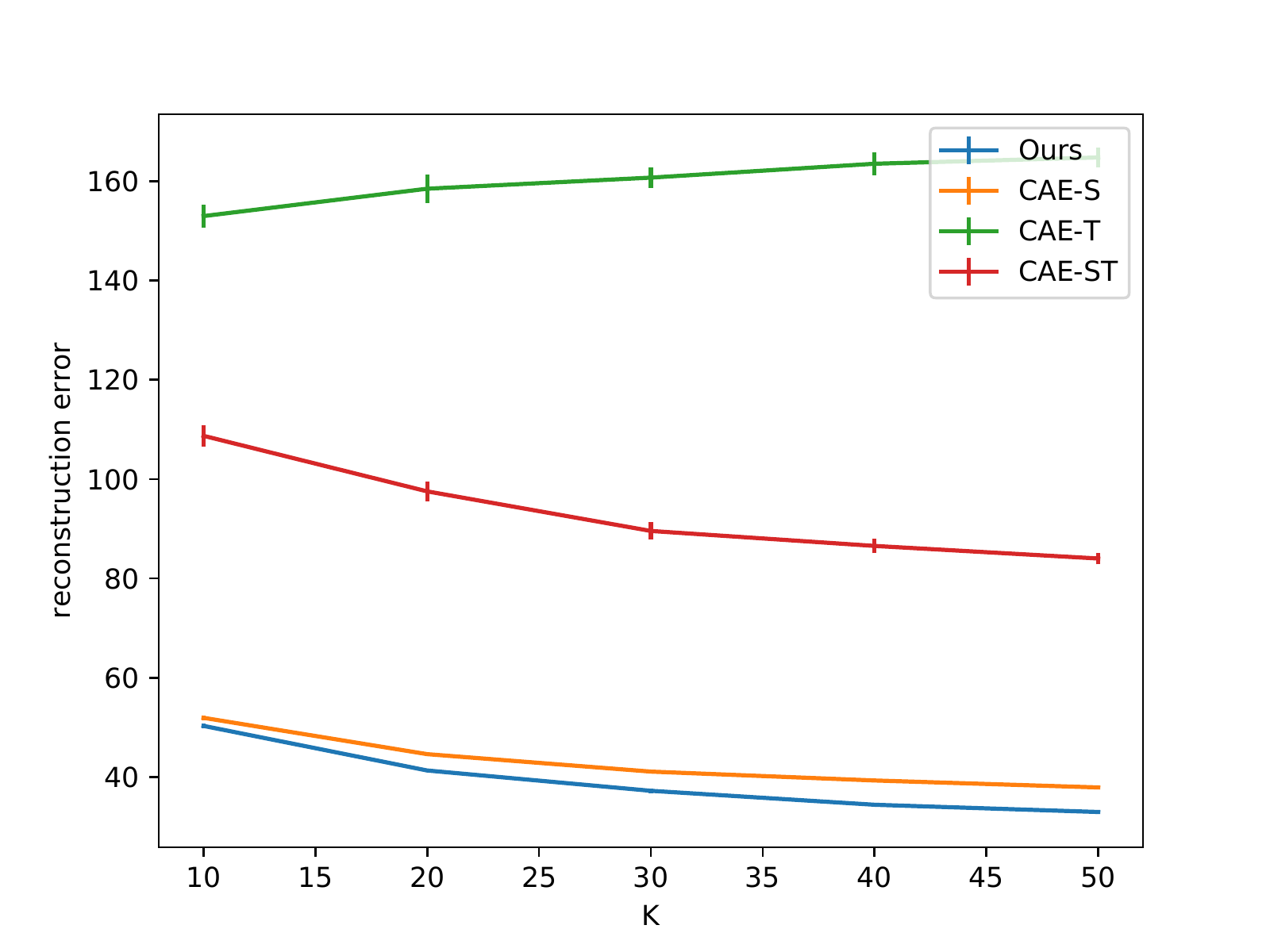}
  \end{minipage} 
  \begin{minipage}{0.33\hsize}
      \centering
      \includegraphics[width=5.2cm]{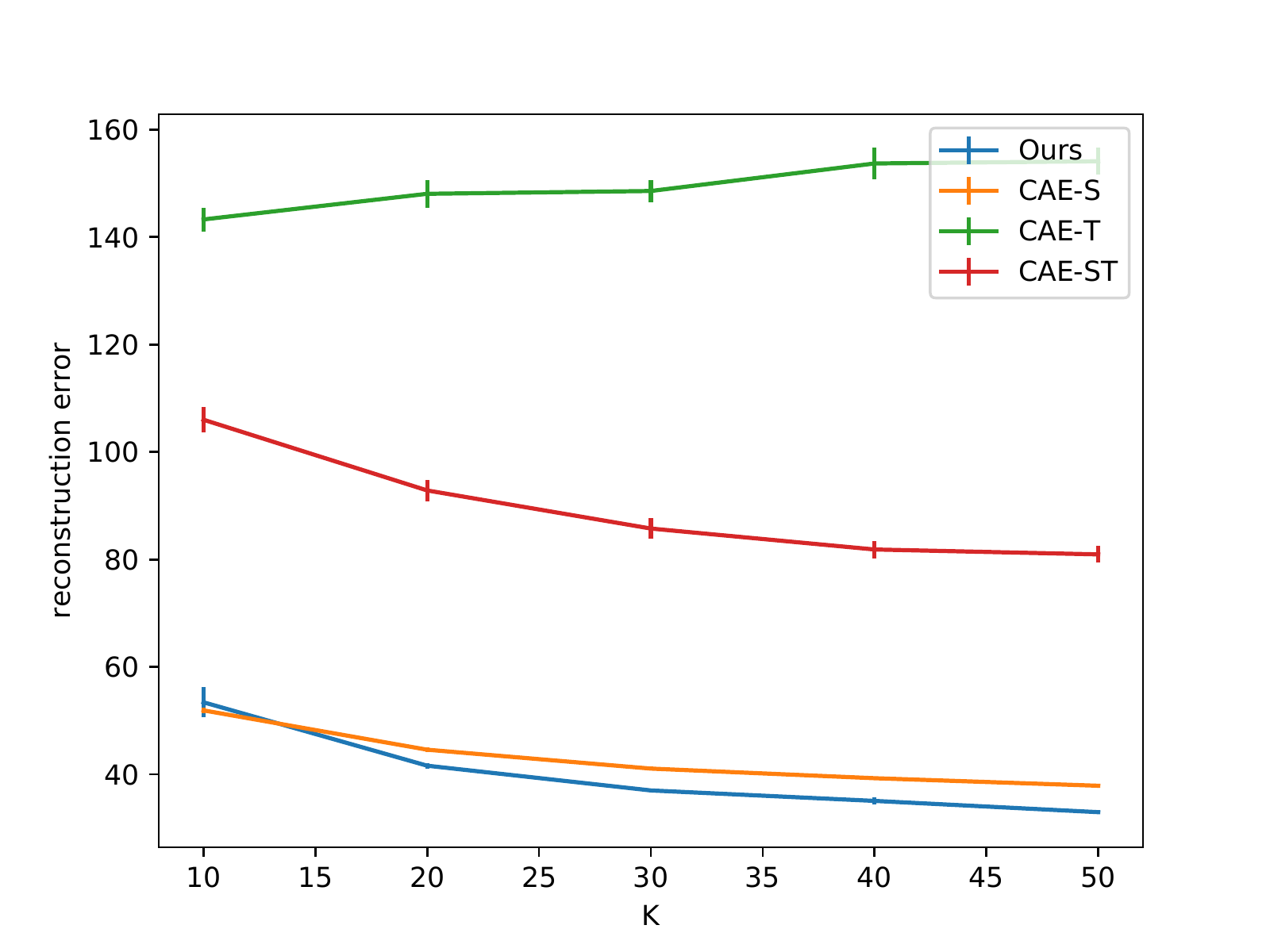}
  \end{minipage}
  \subcaption{Isolet}
  \end{minipage}
  \begin{minipage}{1.00\hsize}
  \begin{minipage}{0.33\hsize}
      \centering
      \includegraphics[width=5.2cm]{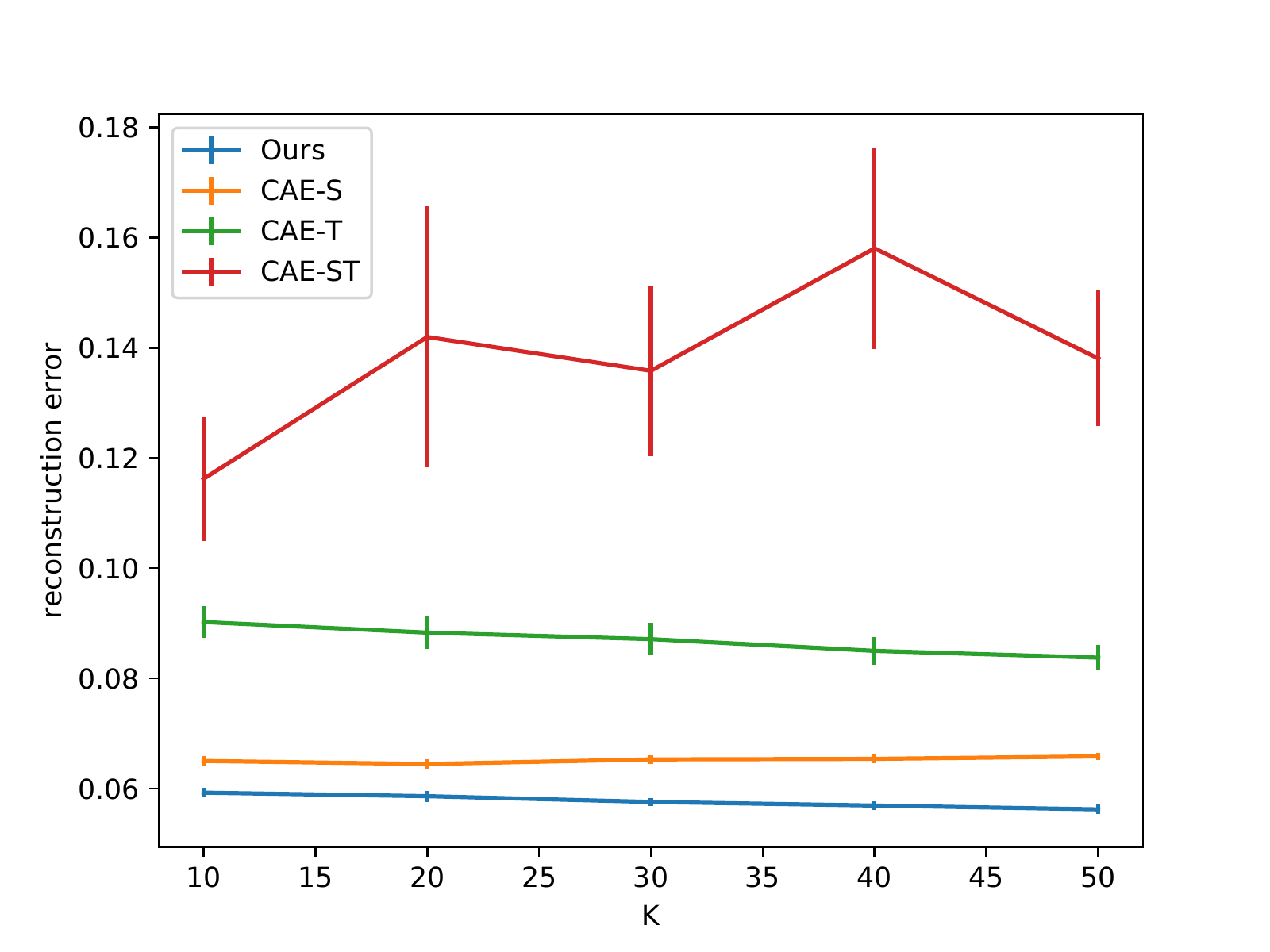}
  \end{minipage} 
  \begin{minipage}{0.33\hsize}
      \centering
      \includegraphics[width=5.2cm]{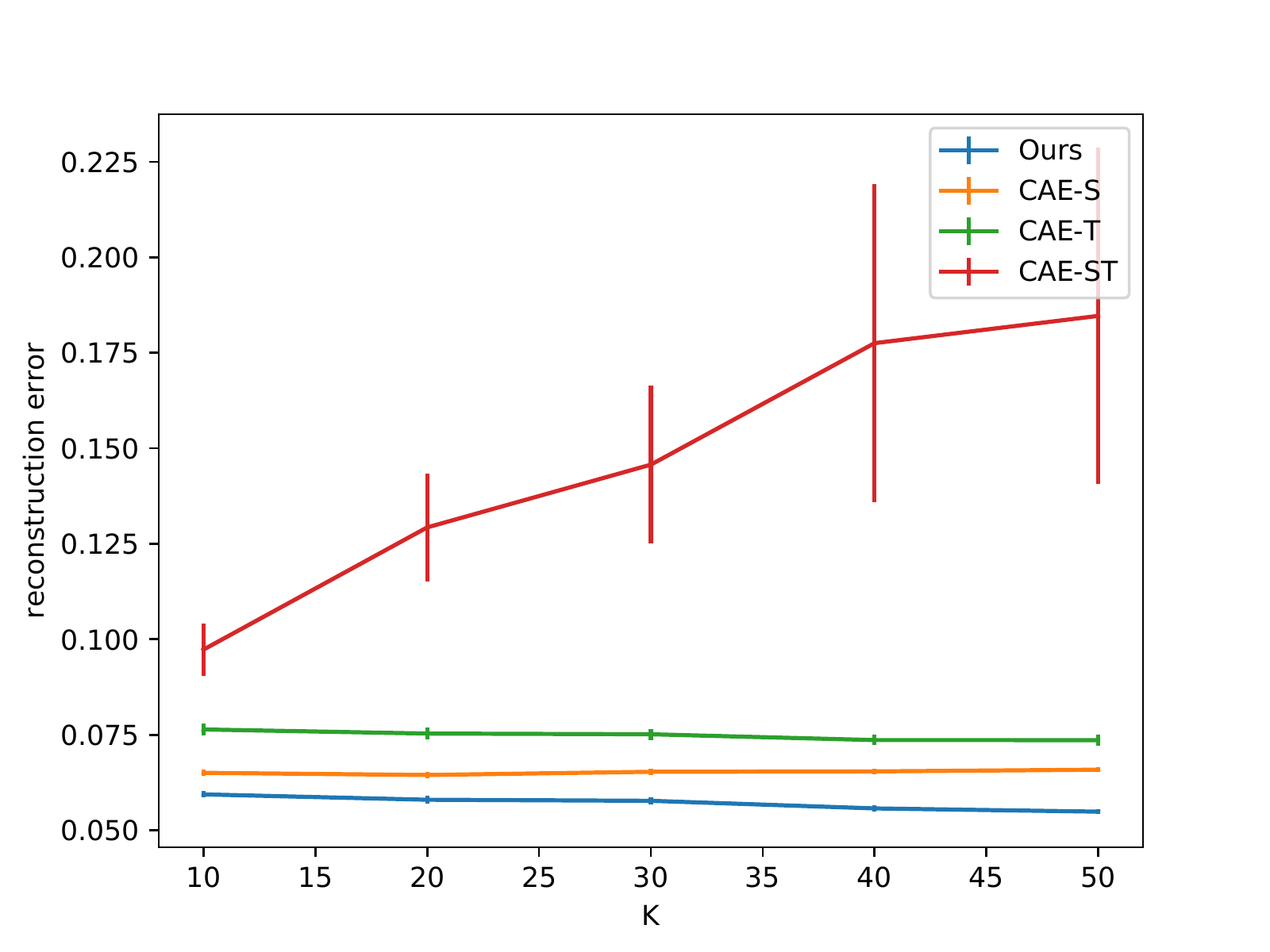}
  \end{minipage}
  \begin{minipage}{0.33\hsize}
      \centering
      \includegraphics[width=5.2cm]{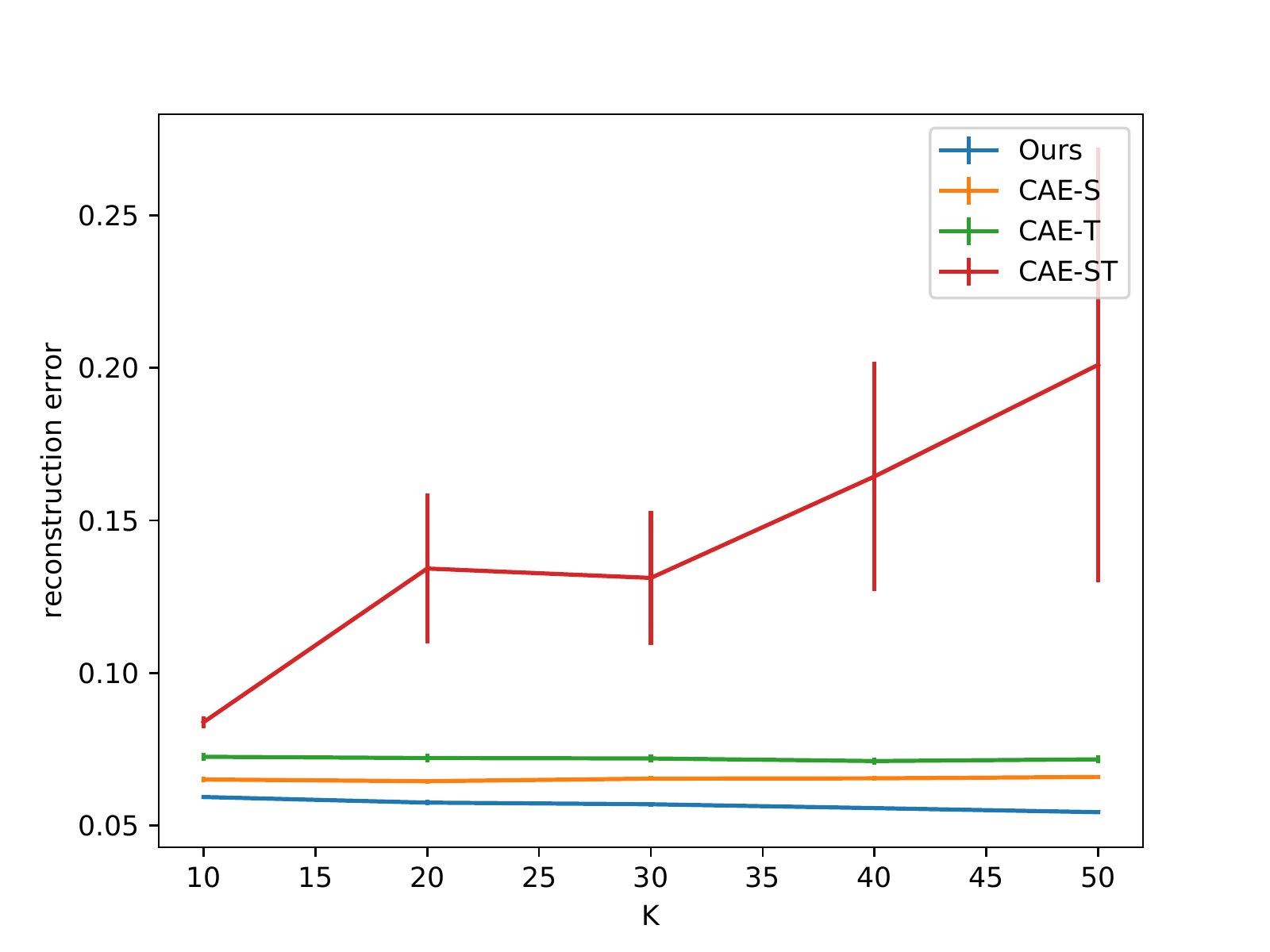}
  \end{minipage}
  \subcaption{Amazon}
  \begin{minipage}{1.00\hsize}
  \begin{minipage}{0.33\hsize}
      \centering
      \includegraphics[width=5.2cm]{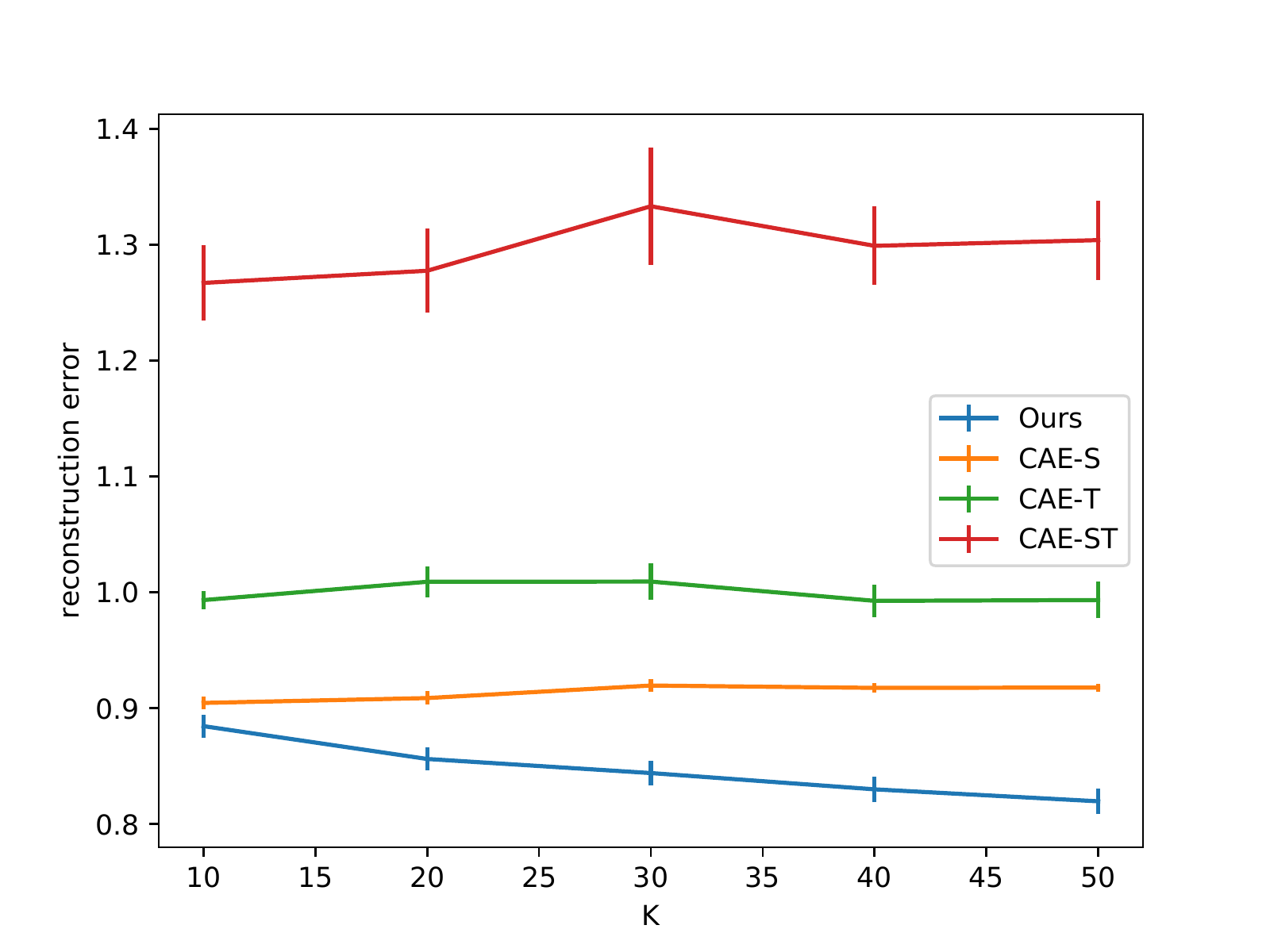}
  \end{minipage}
  \begin{minipage}{0.33\hsize}
      \centering
      \includegraphics[width=5.2cm]{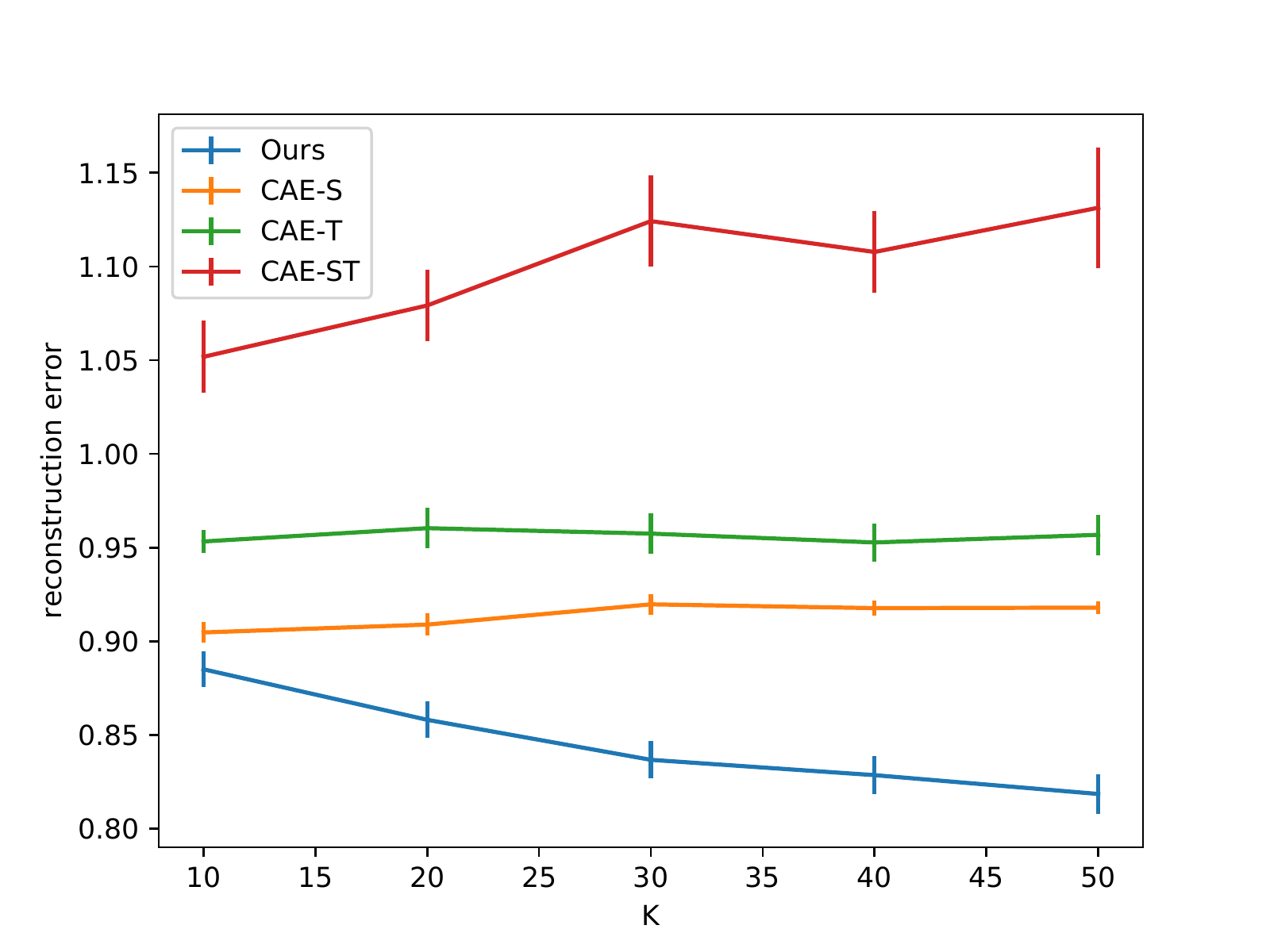}
  \end{minipage} 
  \begin{minipage}{0.33\hsize}
      \centering
      \includegraphics[width=5.2cm]{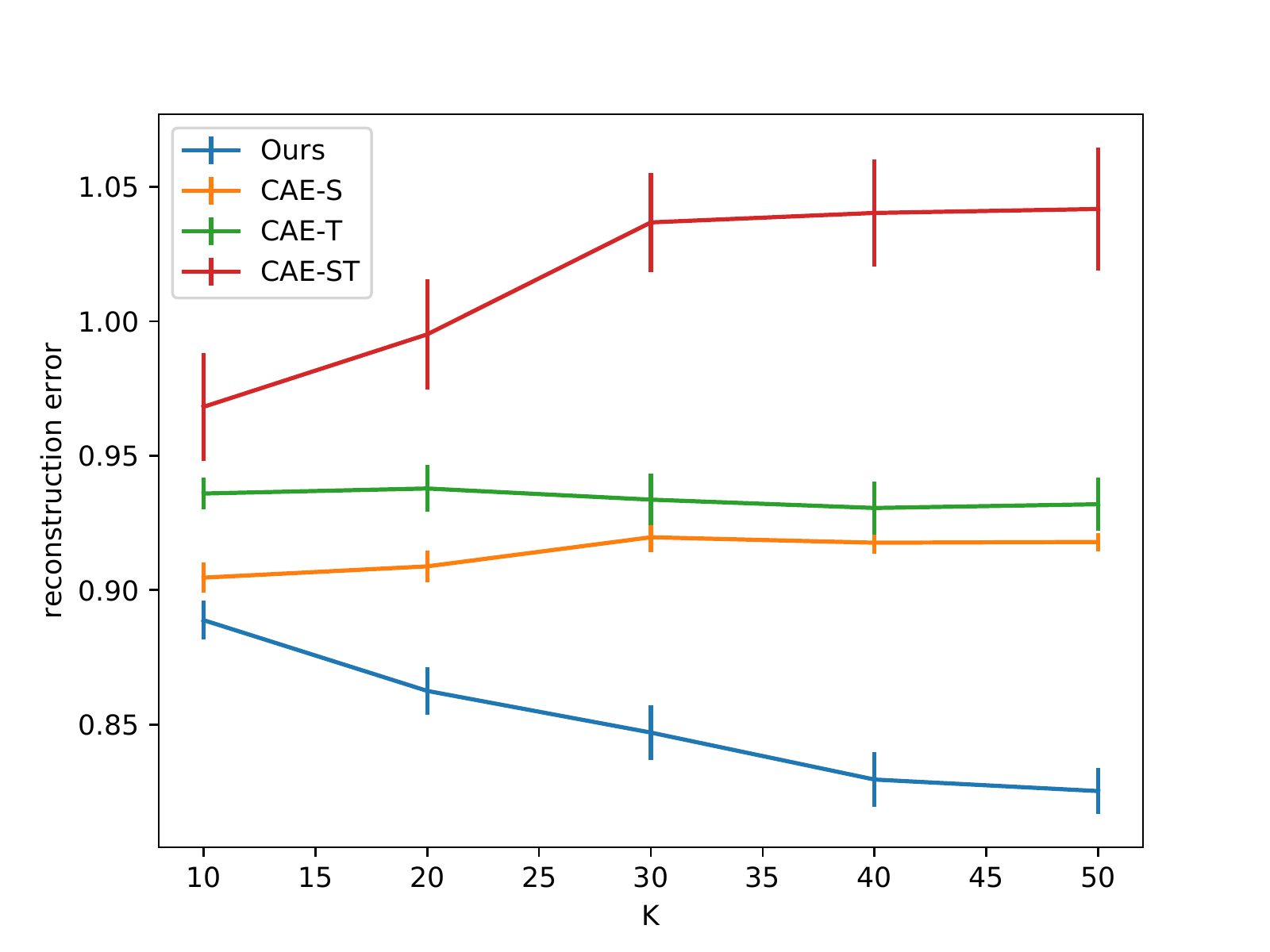}
  \end{minipage}
  \subcaption{VLCS}
  \end{minipage}
  \end{minipage}
  \caption{Average and standard error of test MSREs when changing the number of selected features $K$.
Each column represents the result with the number of target support instances $N_{\rm S}=2,4$, and $6$, respectively from left to right.}
  \label{fig_recon}
\end{figure*}

\begin{figure*}[t]
  \begin{minipage}{1.00\hsize}
  \begin{minipage}{0.33\hsize}
      \centering
      \includegraphics[width=5.2cm]{mnist_ari_test_hw32_hz64_d300_e50000_s2}
  \end{minipage}
  \begin{minipage}{0.33\hsize}
      \centering
      \includegraphics[width=5.2cm]{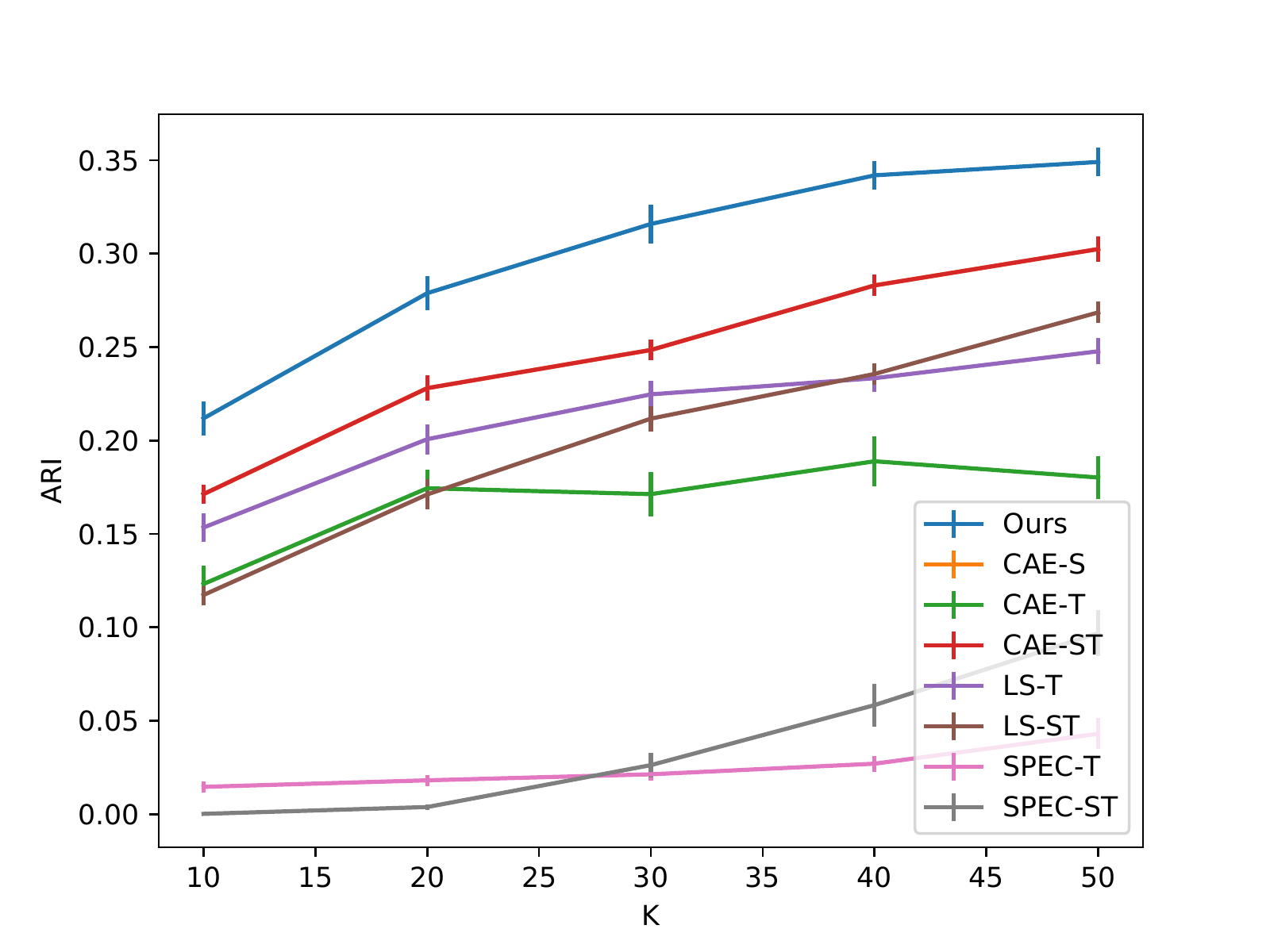}
  \end{minipage}
  \begin{minipage}{0.33\hsize}
      \centering
      \includegraphics[width=5.2cm]{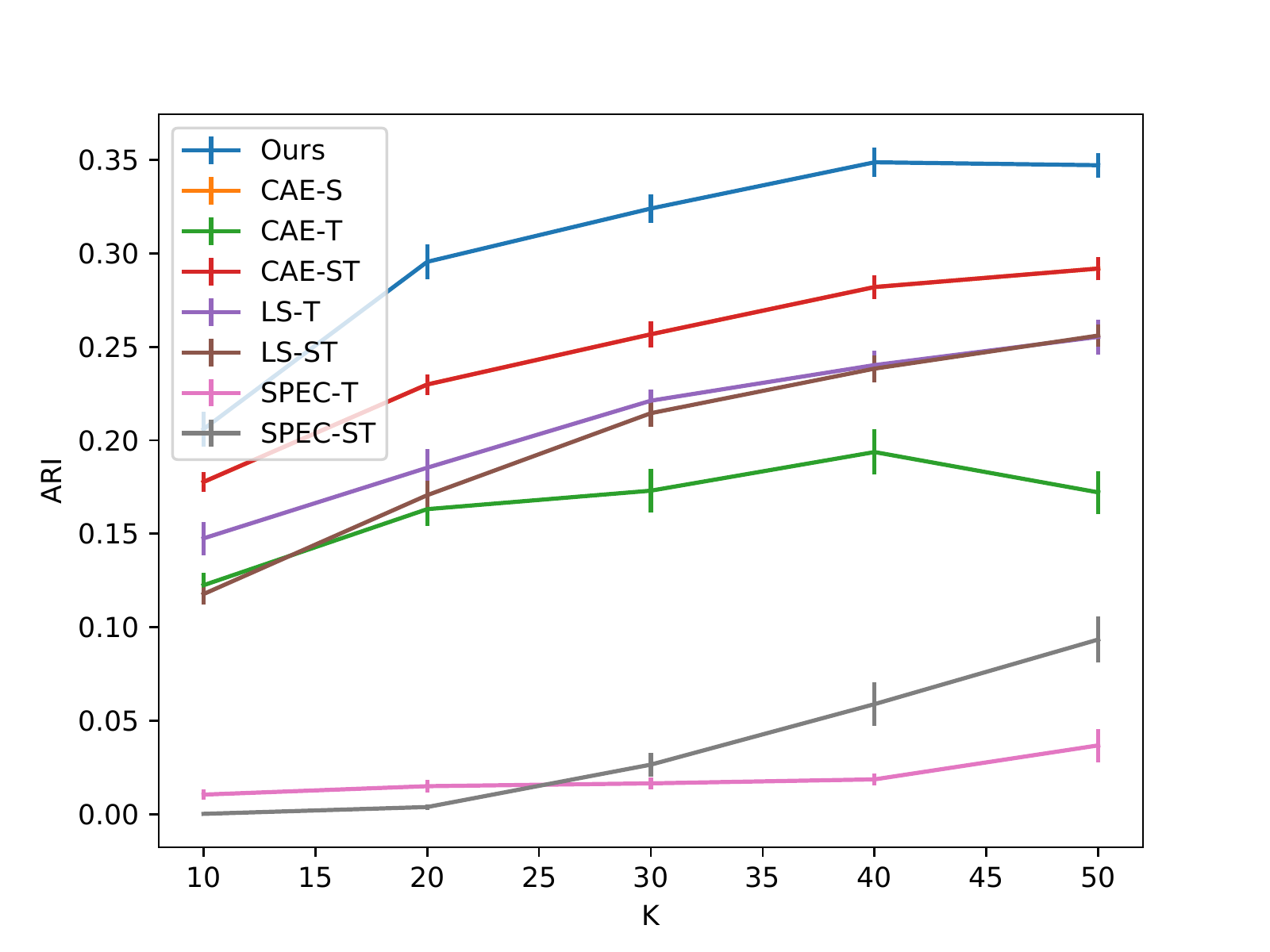}
  \end{minipage} 
  \subcaption{MNIST-r}
  \end{minipage}
  \begin{minipage}{1.00\hsize}
  \begin{minipage}{0.33\hsize}
      \centering
      \includegraphics[width=5.2cm]{isolet_ari_test_hw32_hz64_d300_e50000_s2}
  \end{minipage}
  \begin{minipage}{0.33\hsize}
      \centering
      \includegraphics[width=5.2cm]{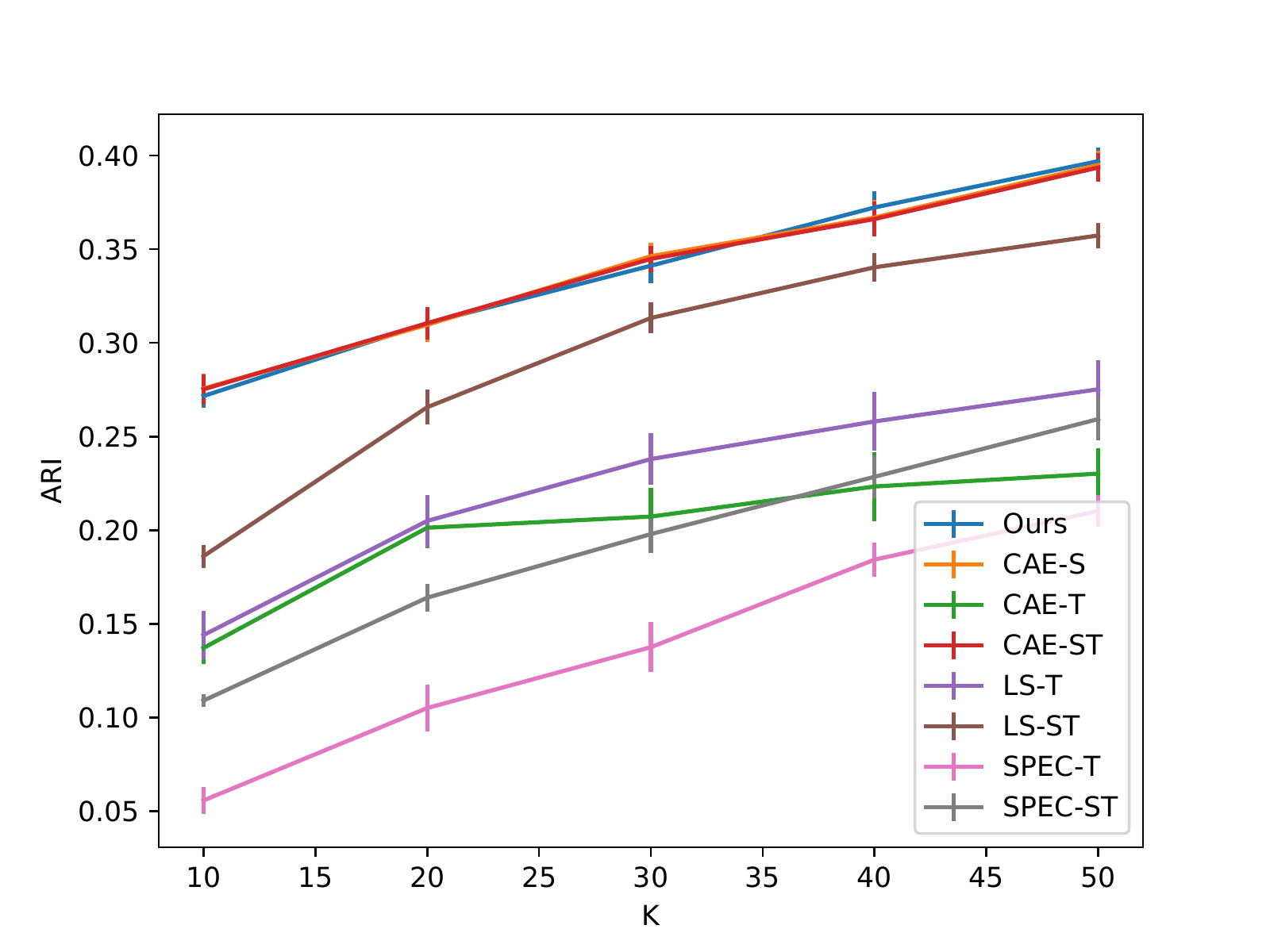}
  \end{minipage} 
  \begin{minipage}{0.33\hsize}
      \centering
      \includegraphics[width=5.2cm]{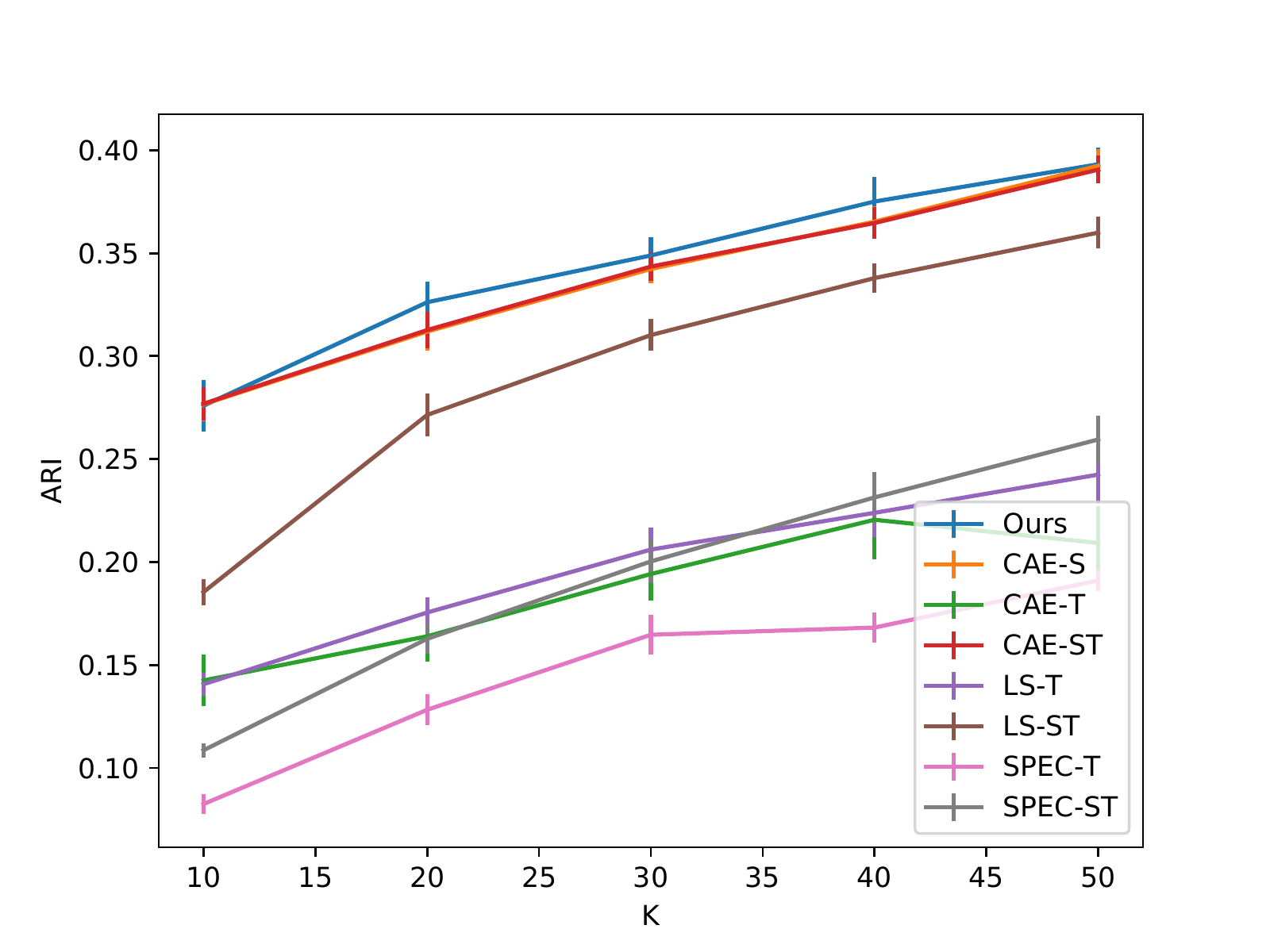}
  \end{minipage}
  \subcaption{Isolet}
  \end{minipage}
  \begin{minipage}{1.00\hsize}
  \begin{minipage}{0.33\hsize}
      \centering
      \includegraphics[width=5.2cm]{sent_ari_test_hw32_hz64_d300_e50000_s2}
  \end{minipage} 
  \begin{minipage}{0.33\hsize}
      \centering
      \includegraphics[width=5.2cm]{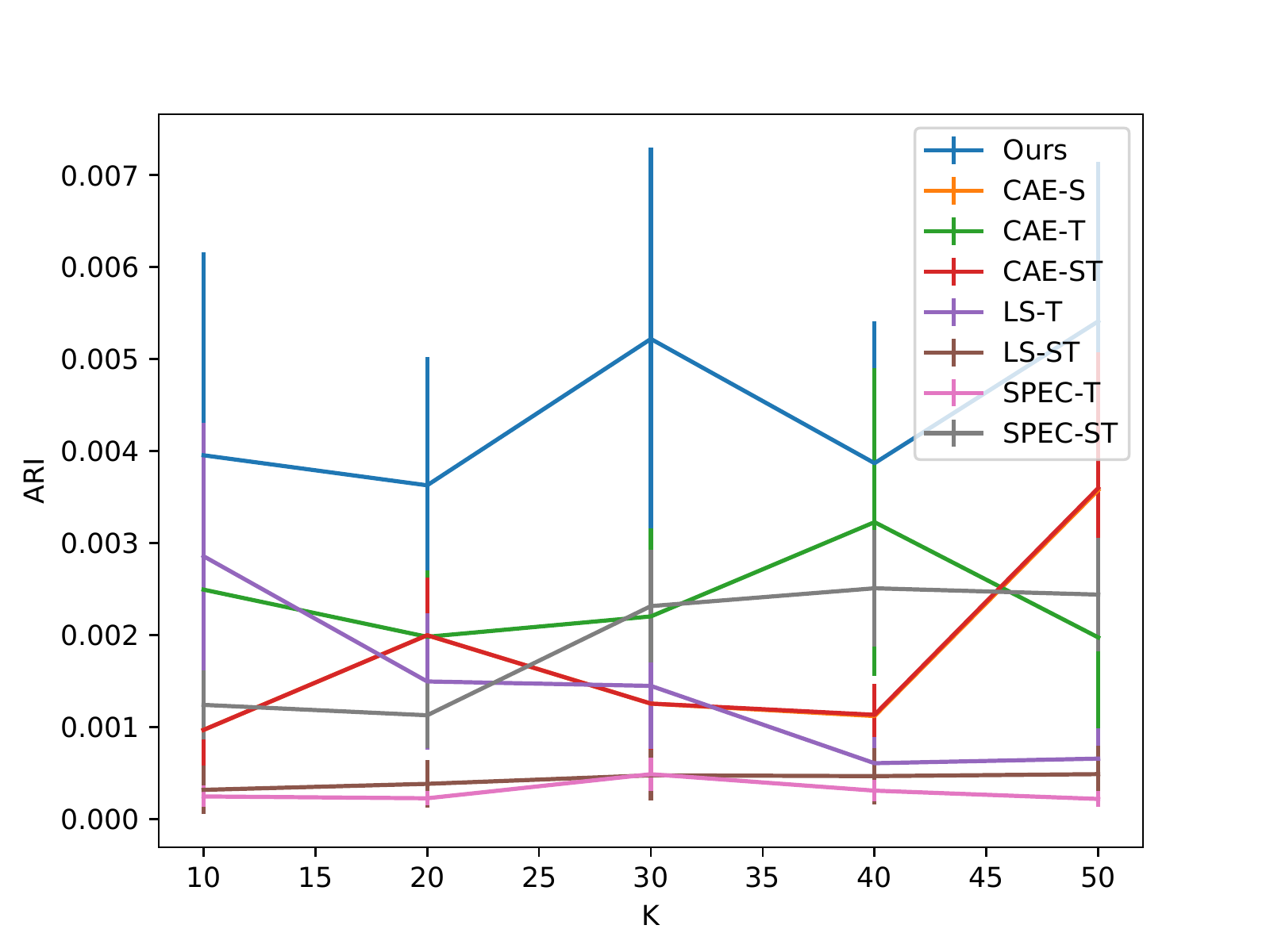}
  \end{minipage}
  \begin{minipage}{0.33\hsize}
      \centering
      \includegraphics[width=5.2cm]{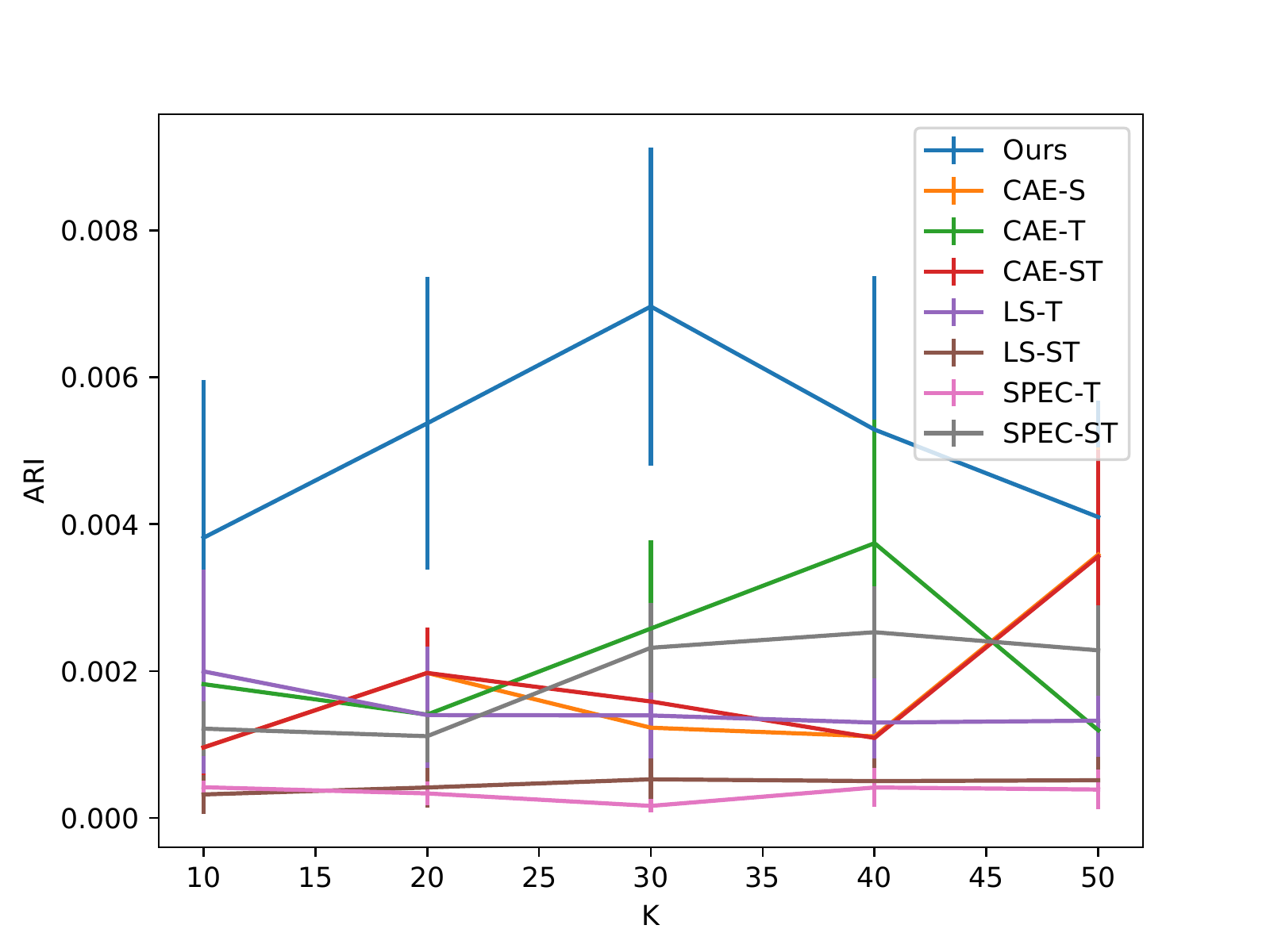}
  \end{minipage}
  \subcaption{Amazon}
  \end{minipage}
  \begin{minipage}{1.00\hsize}
  \begin{minipage}{0.33\hsize}
      \centering
      \includegraphics[width=5.2cm]{vlcs_ari_test_hw256_hz512_d300_e50000_s2}
  \end{minipage} 
  \begin{minipage}{0.33\hsize}
      \centering
      \includegraphics[width=5.2cm]{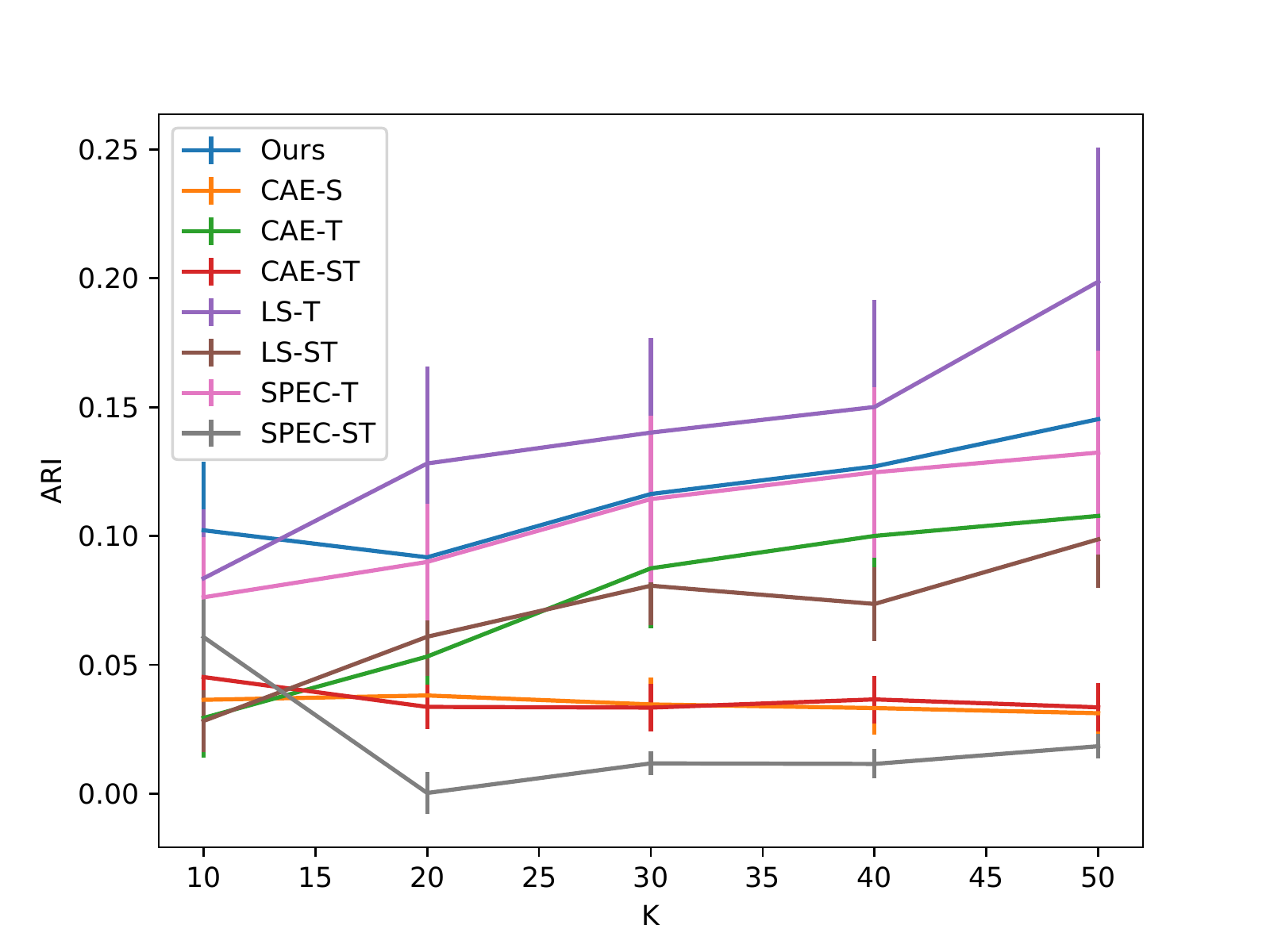}
  \end{minipage}
  \begin{minipage}{0.33\hsize}
      \centering
      \includegraphics[width=5.2cm]{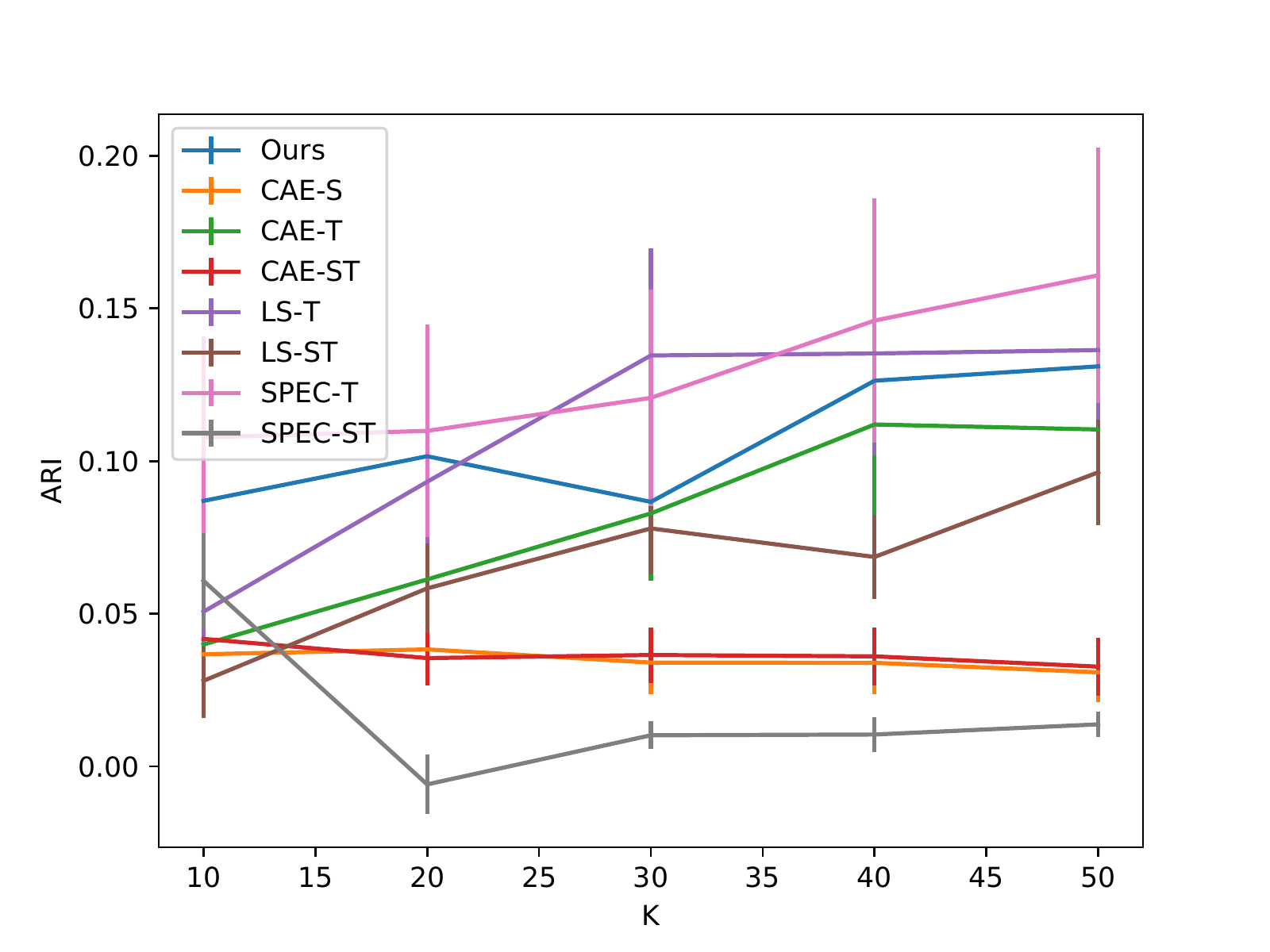}
  \end{minipage}
  \subcaption{VLCS}
  \end{minipage}
  \caption{Average and standard error of test ARIs when changing the number of selected features $K$.
Each column represents the result with the number of target support instances $N_{\rm S}=2,4$, and $6$, respectively from left to right}
  \label{fig_ari}
\end{figure*}

\begin{figure*}[t]
  \begin{minipage}{1.00\hsize}
  \begin{minipage}{0.33\hsize}
      \centering
      \includegraphics[width=5.2cm]{mnist_nmi_test_hw32_hz64_d300_e50000_s2}
  \end{minipage}
  \begin{minipage}{0.33\hsize}
      \centering
      \includegraphics[width=5.2cm]{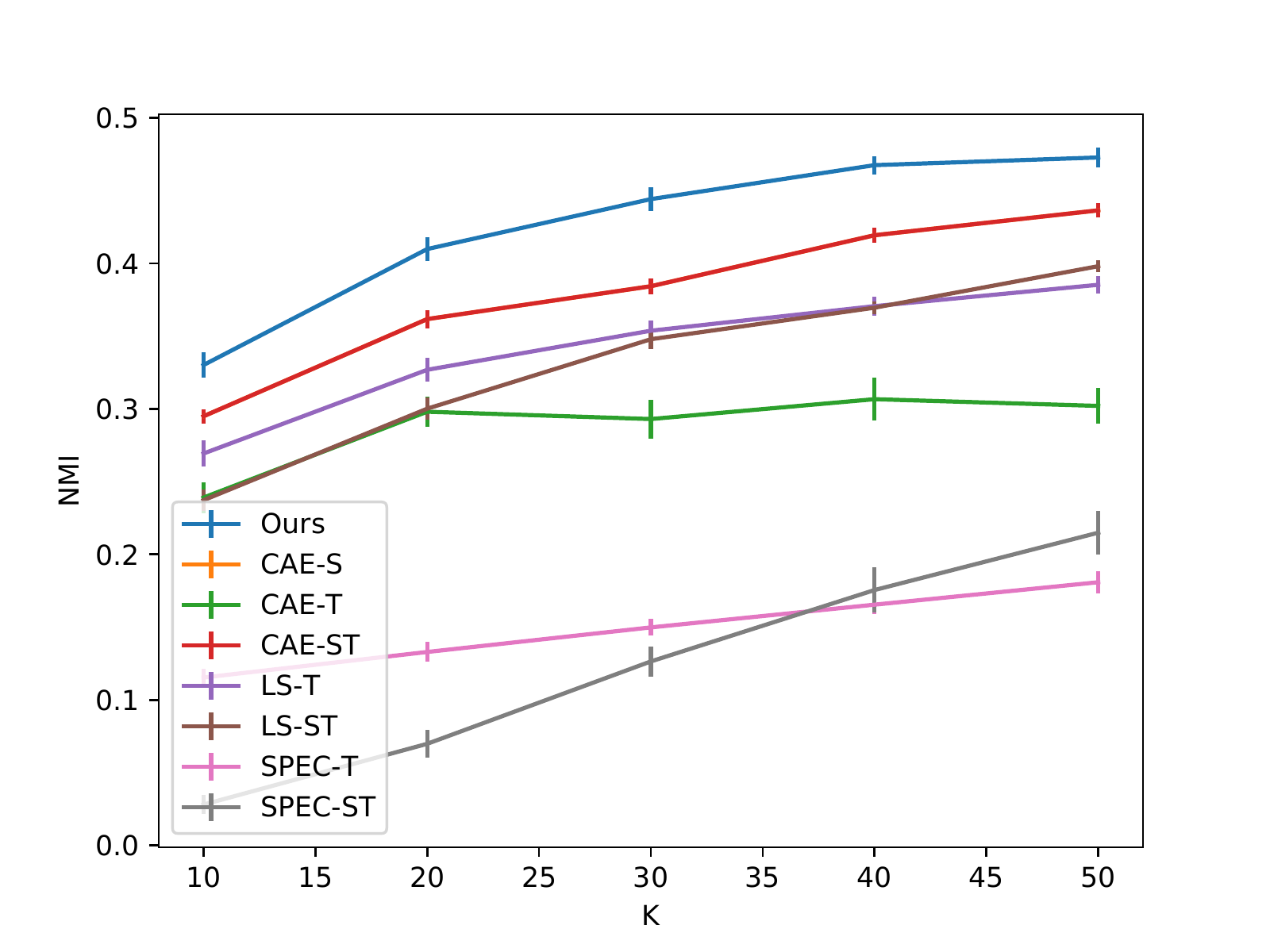}
  \end{minipage}
  \begin{minipage}{0.33\hsize}
      \centering
      \includegraphics[width=5.2cm]{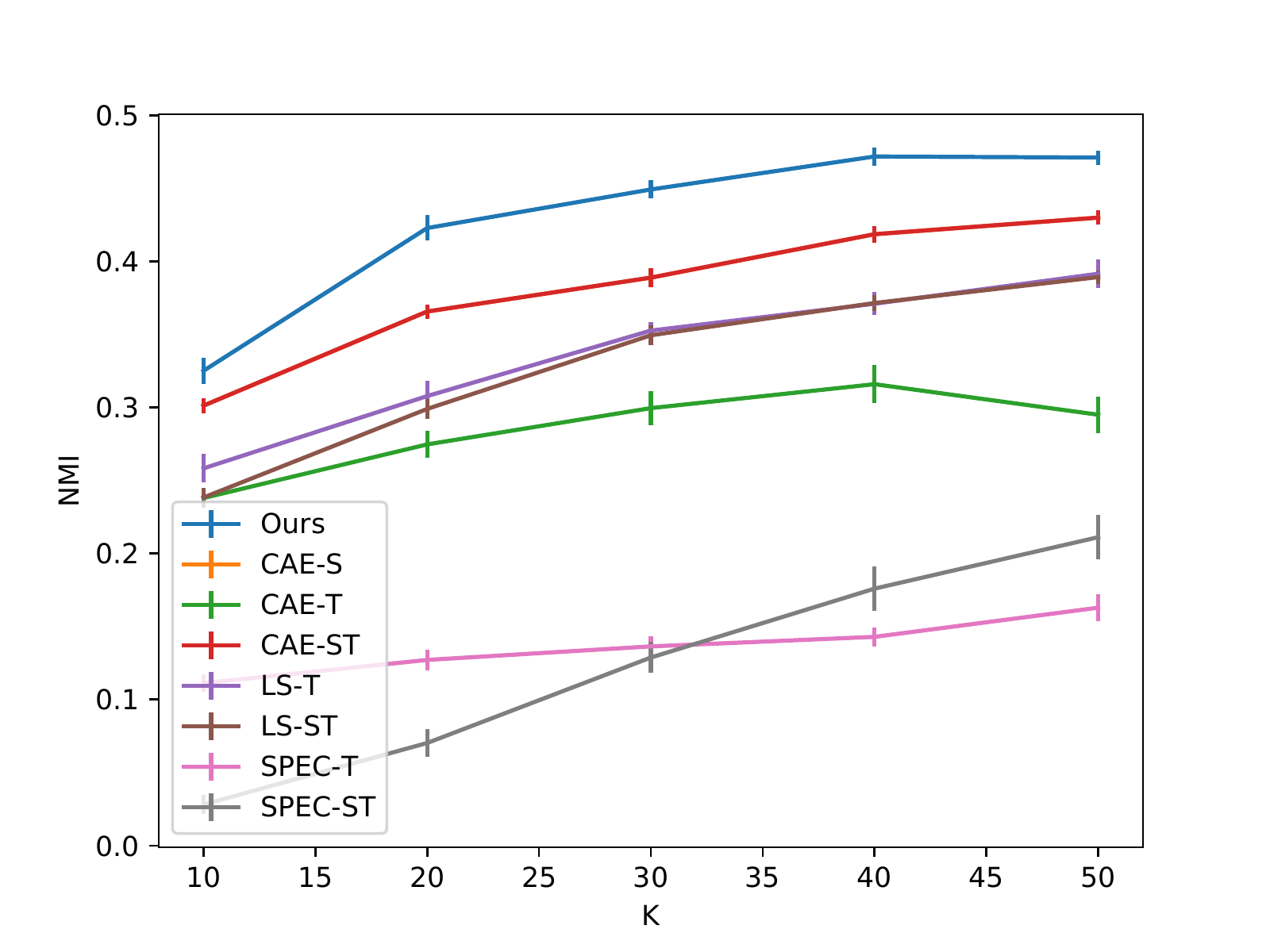}
  \end{minipage} 
  \subcaption{MNIST-r}
  \end{minipage}
  \begin{minipage}{1.00\hsize}
  \begin{minipage}{0.33\hsize}
      \centering
      \includegraphics[width=5.2cm]{isolet_nmi_test_hw32_hz64_d300_e50000_s2}
  \end{minipage}
  \begin{minipage}{0.33\hsize}
      \centering
      \includegraphics[width=5.2cm]{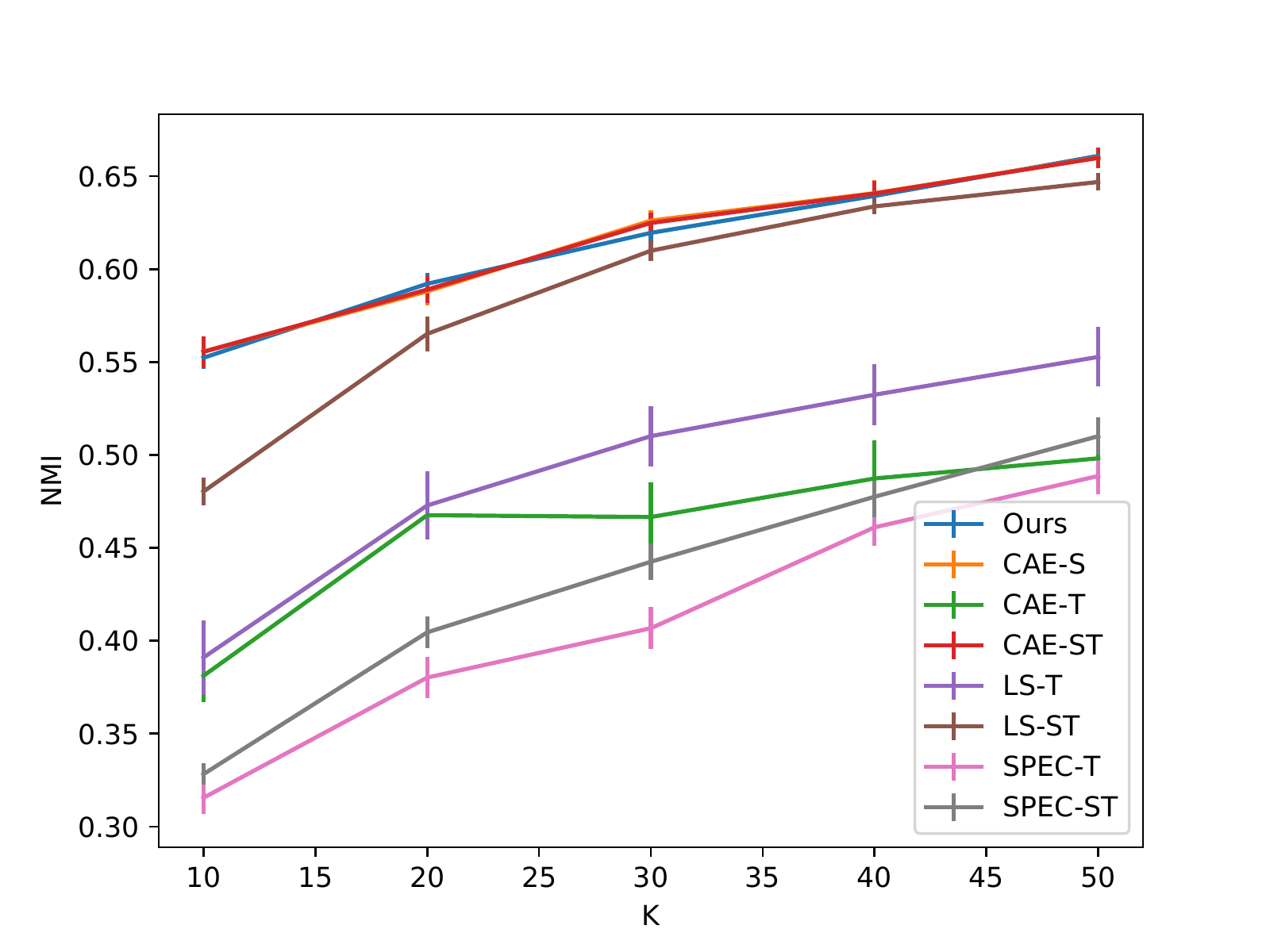}
  \end{minipage} 
  \begin{minipage}{0.33\hsize}
      \centering
      \includegraphics[width=5.2cm]{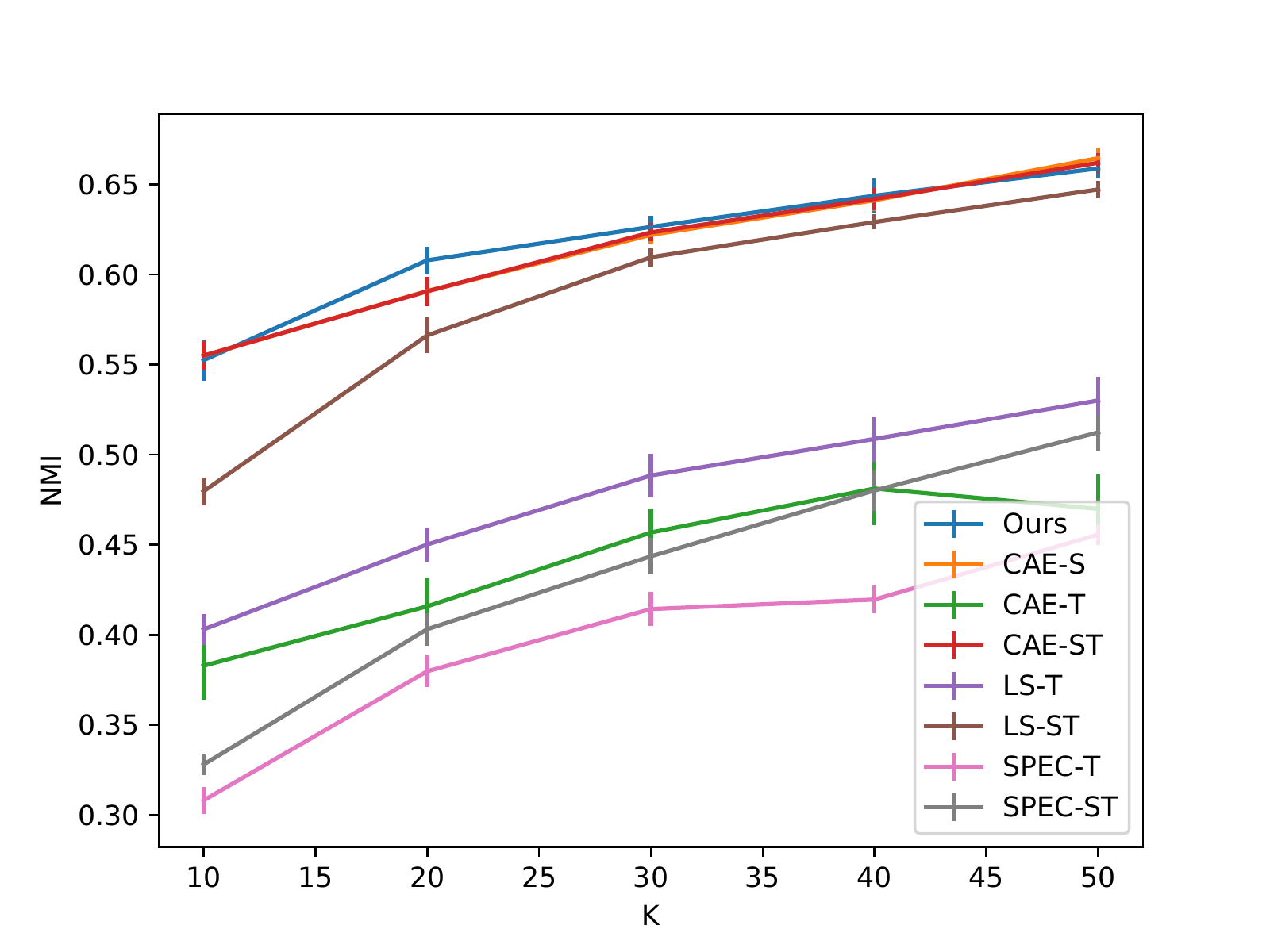}
  \end{minipage}
  \subcaption{Isolet}
  \end{minipage}
  \begin{minipage}{1.00\hsize}
  \begin{minipage}{0.33\hsize}
      \centering
      \includegraphics[width=5.2cm]{sent_nmi_test_hw32_hz64_d300_e50000_s2}
  \end{minipage} 
  \begin{minipage}{0.33\hsize}
      \centering
      \includegraphics[width=5.2cm]{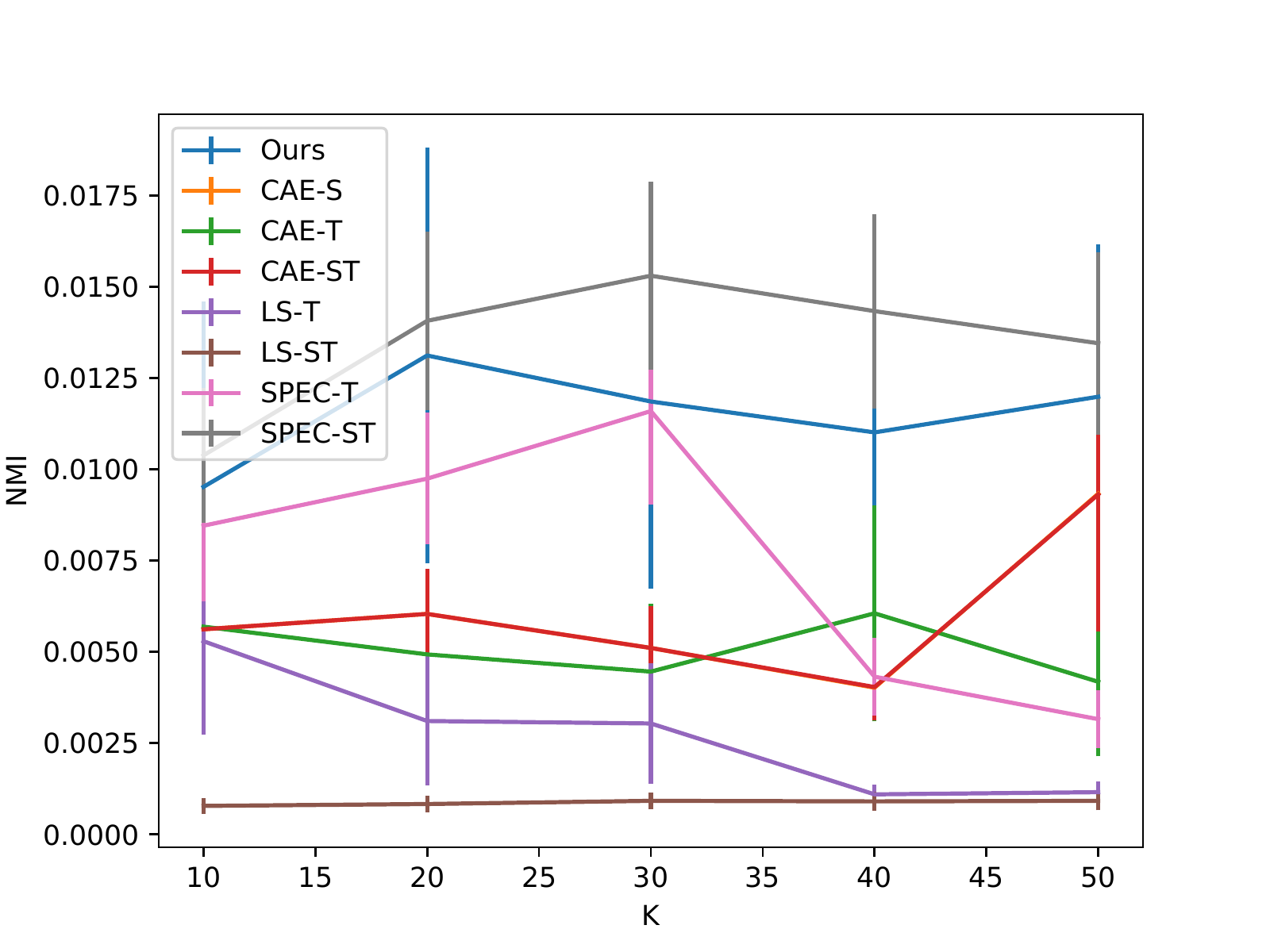}
  \end{minipage}
  \begin{minipage}{0.33\hsize}
      \centering
      \includegraphics[width=5.2cm]{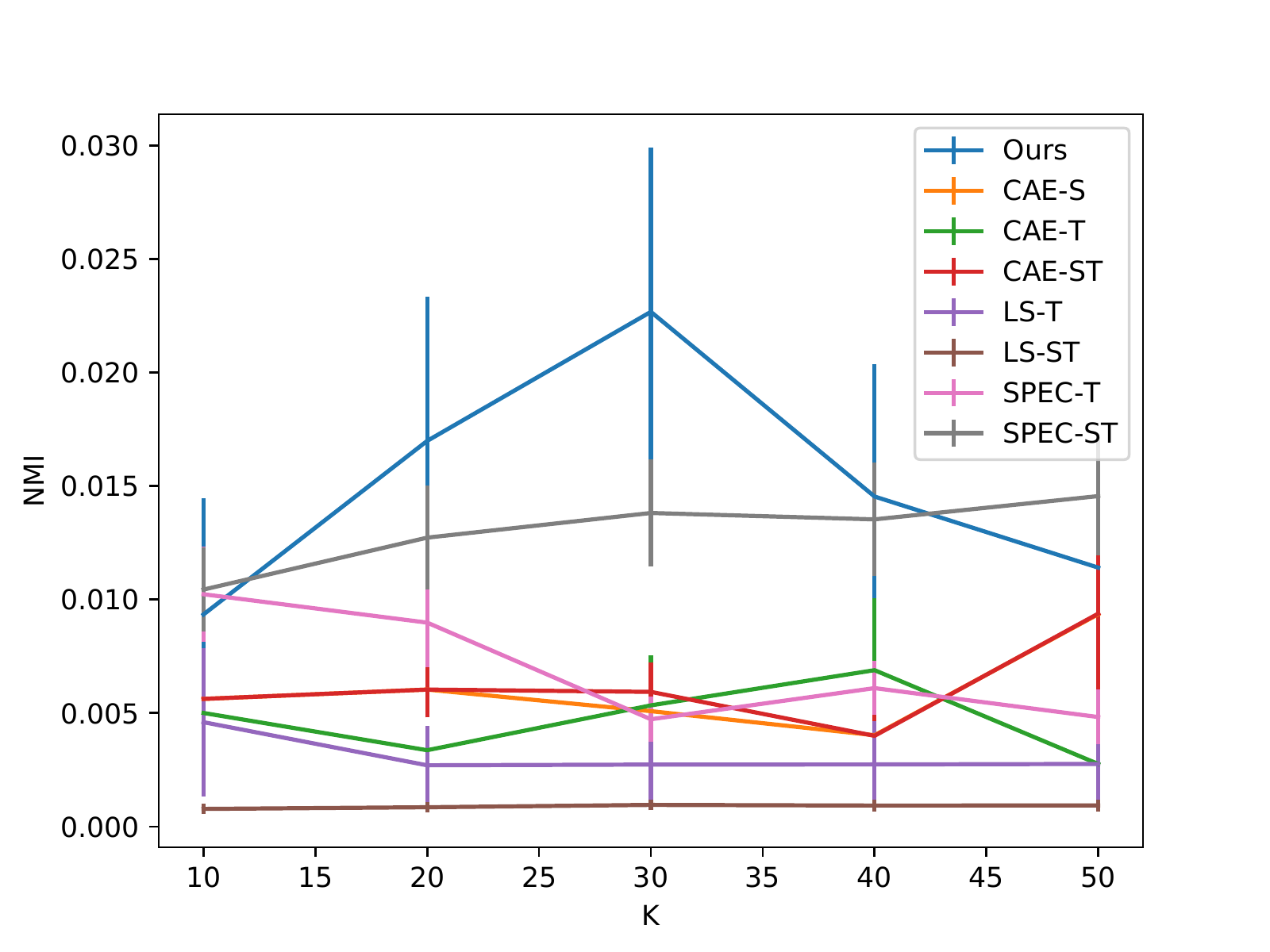}
  \end{minipage}
  \subcaption{Amazon}
  \end{minipage}
  \begin{minipage}{1.00\hsize}
  \begin{minipage}{0.33\hsize}
      \centering
      \includegraphics[width=5.2cm]{vlcs_nmi_test_hw256_hz512_d300_e50000_s2}
  \end{minipage} 
  \begin{minipage}{0.33\hsize}
      \centering
      \includegraphics[width=5.2cm]{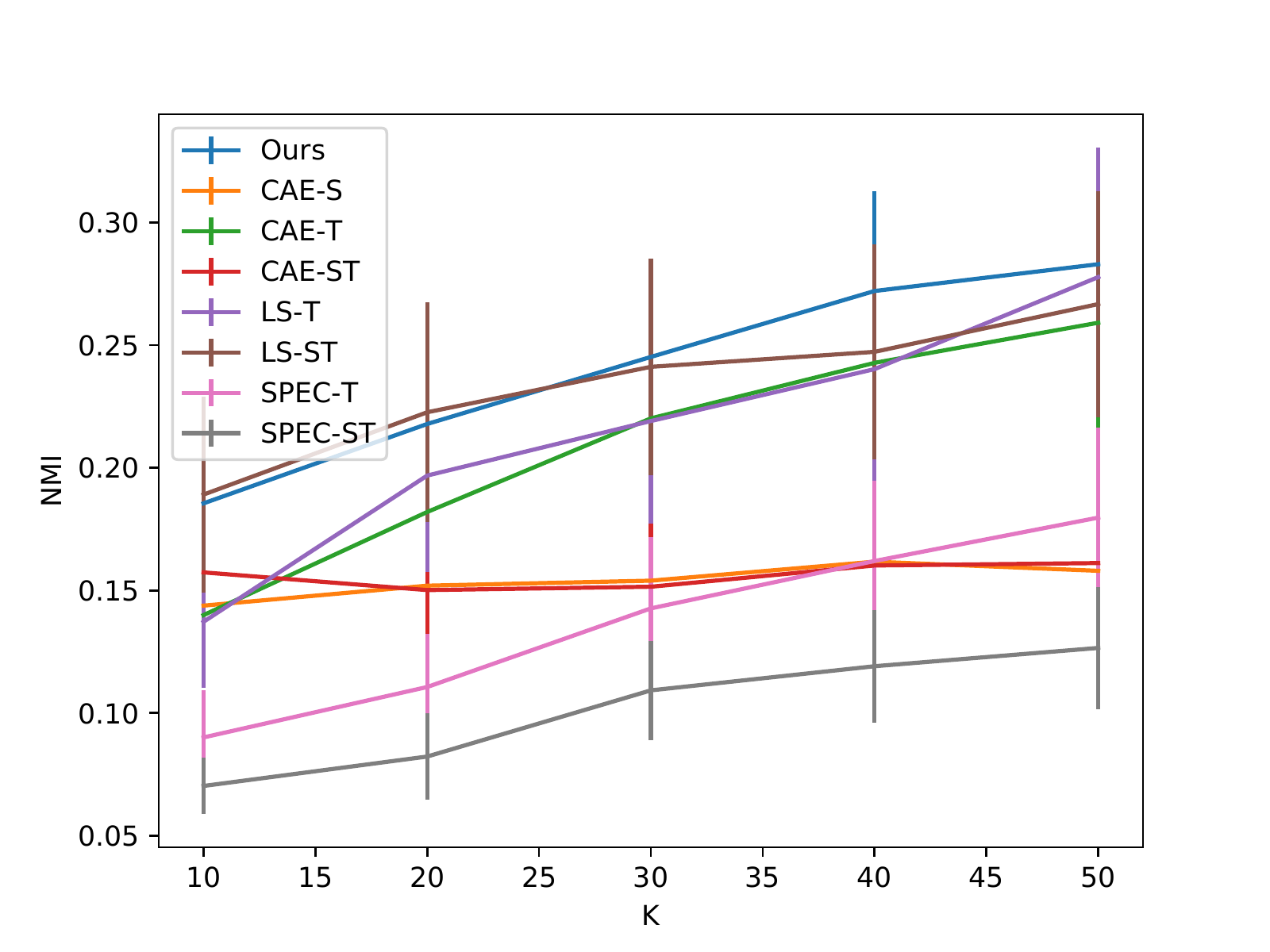}
  \end{minipage}
  \begin{minipage}{0.33\hsize}
      \centering
      \includegraphics[width=5.2cm]{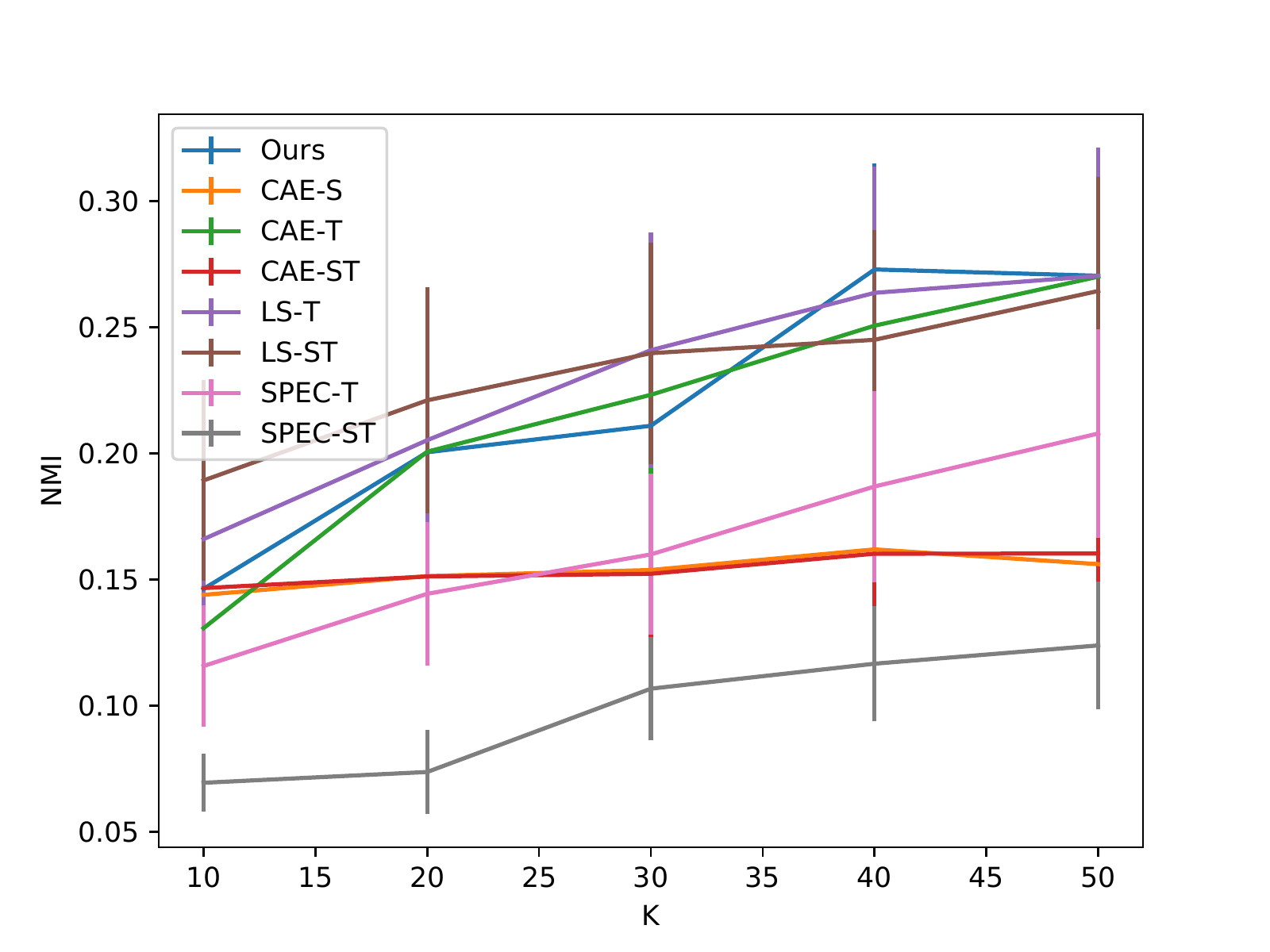}
  \end{minipage}
  \subcaption{VLCS}
  \end{minipage}
  \caption{Average and standard error of test NMIs when changing the number of selected features $K$.
Each column represents the result with the number of target support instances $N_{\rm S}=2,4$, and $6$, respectively from left to right}
  \label{fig_nmi}
\end{figure*}

\end{document}